\documentclass[11pt]{article}

\usepackage[preprint]{acl}

\usepackage{times}
\usepackage{latexsym}

\usepackage[T1]{fontenc}

\usepackage[utf8]{inputenc}

\usepackage{microtype}

\usepackage{inconsolata}

\usepackage{graphicx}
\usepackage{hyperref}       
\usepackage{url}            
\usepackage{booktabs}       
\usepackage{nicefrac}       
\usepackage{xcolor}         
\usepackage{enumitem}       
\usepackage{appendix}       
\usepackage{xspace}
\usepackage{listings}
\usepackage{tcolorbox}
\tcbuselibrary{breakable,listings}
\usepackage[ruled,vlined]{algorithm2e}
\usepackage{adjustbox}
\usepackage{colortbl}
\usepackage{multirow}
\usepackage{wrapfig}
\usepackage{makecell}
\usepackage{authblk}
\usepackage{tikz}
\usetikzlibrary{positioning,shapes,arrows}
\usepackage{amsmath, amsthm, amsfonts, amssymb}
\usepackage{cleveref}
\newtheorem{definition}{Definition}

\crefname{figure}{Figure}{Figures}
\crefname{table}{Table}{Tables}

\title{
\includegraphics[height=0.5cm]{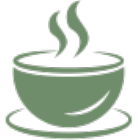}\hspace{0.08cm}AgentOrchestra: Orchestrating Multi-Agent Intelligence with the Tool-Environment-Agent(TEA) Protocol
}

\author{\textbf{Wentao Zhang}\textsuperscript{\rm 1,$\dagger$}, \textbf{Liang Zeng}\textsuperscript{\rm 1,$\dagger$}, \textbf{Yuzhen Xiao}\textsuperscript{\rm 1}, \textbf{Yongcong Li}\textsuperscript{\rm 1}, \textbf{Ce Cui}\textsuperscript{\rm 1}, \\
\textbf{Yilei Zhao}\textsuperscript{\rm 2},
\textbf{Rui Hu}\textsuperscript{\rm 1},
\textbf{Yang Liu}\textsuperscript{\rm 1}, \textbf{Yahui Zhou}\textsuperscript{\rm 1}, \textbf{Bo An}\textsuperscript{\rm 2,1,$\ddagger$}   \\
\textsuperscript{\rm 1} Skywork AI \textsuperscript{\rm 2} Nanyang Technological University\\
\texttt{zhangwent963@gmail.com}, \texttt{bo.an@ntu.edu.sg}
}

\newcommand{\projectname}{\textbf{\textsc{AgentOrchestra}}\xspace}

\definecolor{rowOpen}{HTML}{F0FDF4}
\definecolor{rowOurs}{HTML}{DBEAFE}

\begin{document}
\maketitle
\begin{abstract}

Recent advances in LLM-based agent systems have shown promise on complex, long-horizon tasks, but existing agent protocols (e.g., A2A and MCP) do not adequately support lifecycle-aware coordination across agents, tools, and environments. To address this limitation, we introduce the \textbf{Tool-Environment-Agent} (TEA) protocol, a unified abstraction that models these components as first-class, versioned resources with explicit lifecycles. TEA supports end-to-end context and version management, improving traceability and reproducibility, while also enabling continual self-evolution of agent-associated components\footnote{Unless otherwise specified, \emph{agent-associated components} include prompts, memory/tool/agent/environment code, and agent outputs (solutions).}. Building on TEA, we present \projectname, a hierarchical multi-agent framework in which a central planner coordinates specialized sub-agents and dynamically extends capabilities during execution. Experiments on four challenging benchmarks, spanning expert-level agent tasks and scientific/mathematical reasoning, show that \projectname consistently outperforms strong baselines; in particular, it achieves 89.04\% on the GAIA Test set, placing it among the leading methods to the best of our knowledge. These results highlight the value of explicit protocol design and hierarchical orchestration for building more robust and adaptive multi-agent systems.
\end{abstract}

\section{Introduction}

Recent advances in LLM-based agent systems have enabled strong performance on general-purpose tasks, including complex long-horizon settings, across diverse domains such as web navigation~\citep{openai2025operator,browser_use2024}, computer use~\citep{anthropic2024computeruse,qin2025ui}, code execution~\citep{wang2024executablecodeactionselicit}, game playing~\citep{wang2023voyager,tan2024cradle}, and research assistance~\citep{openai2024deepresearch,google2024deepresearch,xai2025grok3}. Yet cross-environment generalization remains limited. Context is scattered across prompts and logs, environment integration relies on brittle glue code, and agent-associated components are typically fixed rather than improved through execution feedback.

Current agent protocols also fall short as a general substrate for scalable agents. As summarized in Table~\ref{tab:protocol_comparison}, representative protocols such as Google’s A2A~\citep{google2024a2a} and Anthropic’s MCP~\citep{anthropic2024mcp} provide important building blocks, including task-level collaboration and messaging in A2A, and tool/resource schemas, discovery, and invocation in MCP. However, three protocol-level gaps remain. First, \textbf{lifecycle and context management are fragmented}, since neither protocol standardizes primitives for maintaining consistent, versioned execution context across agent-associated components. Second, \textbf{self-evolution is not supported at the protocol level}, since prompts and resources are largely treated as externally maintained assets rather than components refined from execution feedback under traceable versioning. Third, \textbf{environments are not first-class}, since they are delegated to application-specific runtimes rather than managed components with explicit boundaries and constraints. As a result, switching agents across environments, reusing environments, and isolating parallel runs often devolve into glue-code orchestration.

\begin{table}[t]
  \centering
  \scriptsize
  \renewcommand{\arraystretch}{1.35}
  \setlength{\tabcolsep}{5pt}
  \begin{tabular}{l>{\columncolor{rowOurs}}ccc}
    \toprule
    \rowcolor{gray!18}
    \textbf{Dimension}
      & \textbf{TEA (Ours)}
      & \textbf{A2A}
      & \textbf{MCP} \\
    \midrule
    Core Entities
      & \textbf{Tool, Env, Agent}
      & Agent, Tool
      & Model, Tool \\
    Lifecycle \& Versioning
      & \textcolor{green!60!black}{$\checkmark$}
      & \textcolor{red!80!black}{$\boldsymbol{\times}$}
      & \textcolor{red!80!black}{$\boldsymbol{\times}$} \\
    Entity Transformations
      & \textcolor{green!60!black}{$\checkmark$}
      & \textcolor{red!80!black}{$\boldsymbol{\times}$}
      & \textcolor{red!80!black}{$\boldsymbol{\times}$} \\
    Self-Evolution
      & \textcolor{green!60!black}{$\checkmark$}
      & \textcolor{red!80!black}{$\boldsymbol{\times}$}
      & \textcolor{red!80!black}{$\boldsymbol{\times}$} \\
    Open Ecosystem
      & \textcolor{green!60!black}{$\checkmark$}
      & \textcolor{orange!90!black}{$\blacktriangle$}
      & \textcolor{orange!90!black}{$\blacktriangle$} \\
    \bottomrule
  \end{tabular}
  
  \caption{Comparison of TEA Protocol with A2A and MCP.
    \textcolor{green!60!black}{$\checkmark$}~Supported;
    \textcolor{orange!90!black}{$\blacktriangle$}~Partial;
    \textcolor{red!80!black}{$\boldsymbol{\times}$}~Not supported.}
  \label{tab:protocol_comparison}
  \vspace{-0.5cm}
\end{table}

To address these limitations, we propose the \textbf{Tool–Environment–Agent} (TEA) protocol, a unified layer that treats environments, agents, and tools as explicitly managed components. TEA standardizes component identifiers, version semantics, and run-scoped context, allowing execution state and artifacts to remain traceable across iterations. It also introduces closed-loop evolution hooks driven by execution feedback and models environments as first-class components with explicit boundaries and constraints, such as web sandboxes, file systems, and code execution runtimes. Together, these abstractions make agent construction more composable, reusable, and reproducible in practice. Detailed motivations for the TEA protocol and in-depth comparisons with existing protocols are provided in Appendix~\ref{appx_sec:comprehensive_motivation}, ~\ref{appx_sec:comparison_with_other_protocols}.


Based on the TEA protocol, we develop \projectname, a hierarchical multi-agent framework for general-purpose task solving. \projectname uses a central planner to decompose a user objective and delegate sub-tasks to specialized agents for research, web navigation, analysis, tool synthesis, and reporting. Unlike flat coordination, where an orchestrator must select from a growing global pool of agents and tools and often accumulates irrelevant context, \projectname adopts hierarchical delegation with localized tool ownership. The planner routes each sub-task to a domain-specific sub-agent (or environment), which exposes only a curated toolset and local context. This turns global coordination into a sequence of localized routing decisions. As a result, the framework supports tree-structured expansion while keeping the orchestrator’s decision scope bounded. \projectname also incorporates a self-evolution module that leverages TEA’s lifecycle and versioning mechanisms to refine agent-associated components from execution feedback. Our contributions are threefold:
\begin{itemize}[leftmargin=*,itemsep=0.1em]
\item We introduce the TEA protocol, which unifies environments, agents, and tools as first-class, versioned components with lifecycles to support context management and execution.
\item We develop \projectname, a hierarchical multi-agent system built on TEA, demonstrating scalable orchestration through tree-structured routing and feedback-driven self-evolution.
\item We conduct extensive evaluations on four challenging benchmarks, including ablations to isolate the effects of key components. \projectname consistently outperforms strong baselines and achieves 89.04\% on GAIA, placing it among the leading methods to the best of our knowledge.
\end{itemize}

\section{Related Work}

\subsection{Tool and Agent Protocols}
Recent protocols standardize tool interfaces and agent communication. For instance, MCP~\citep{anthropic2024mcp} unifies tool integration for LLMs, while A2A~\citep{google2024a2a} enables agent-to-agent messaging and coordination. Other efforts, such as the Agent Network Protocol (ANP)~\citep{ehtesham2025survey} and frameworks like SAFEFLOW~\citep{li2025safeflow}, enhance interoperability and safety in multi-agent systems. While these protocols provide essential building blocks, they primarily treat agents and tools as isolated service endpoints, often overlooking environments as dynamic, first-class components. TEA extends these existing standards rather than replacing them. By integrating tools, environments, and agents into a unified context-aware framework, TEA resolves protocol fragmentation with integrated lifecycle and version management missing in MCP or A2A.

\subsection{General-Purpose Agents}
Integrating tools with LLMs represents a paradigm shift, enabling agents to exhibit enhanced flexibility, cross-domain reasoning, and natural language interaction~\citep{liang2025llm}. Such systems have demonstrated efficacy across diverse domains, including web browsing~\citep{openai2025operator, browser_use2024}, computer operation~\citep{anthropic2024computeruse, qin2025ui}, code execution~\citep{wang2024executablecodeactionselicit}, and game playing~\citep{wang2023voyager, tan2024cradle}. Standardized interfaces like OpenAI's Function Calling and Anthropic's MCP~\citep{openai2023functioncalling, anthropic2024mcp}, alongside frameworks such as ToolMaker~\citep{wolflein2025llm}, have further streamlined the synthesis of LLM-compatible tools. Building upon these foundations, multi-agent architectures like MetaGPT~\citep{hong2023metagpt} demonstrate the potential of specialized agent coordination for complex problem-solving. However, many current approaches still struggle with efficient communication, dynamic role allocation, and scalable teamwork. The emergence of generalist frameworks, including Manus~\citep{shen2025mindmachinerisemanus}, OpenHands~\citep{wang2024openhands}, and smolagents~\citep{smolagents2025}, has advanced unified perception and tool-augmented action. While recent efforts like Alita~\citep{qiu2025alita} explore minimal predefinition and maximal self-evolution, these systems often lack unified protocols for cross-layer resource management. This gap motivates our proposal of the TEA Protocol and \projectname.

\section{The TEA Protocol}
\label{sec:tea_protocol}

\begin{figure}[t]
  \centering
  \includegraphics[width=0.40\textwidth]{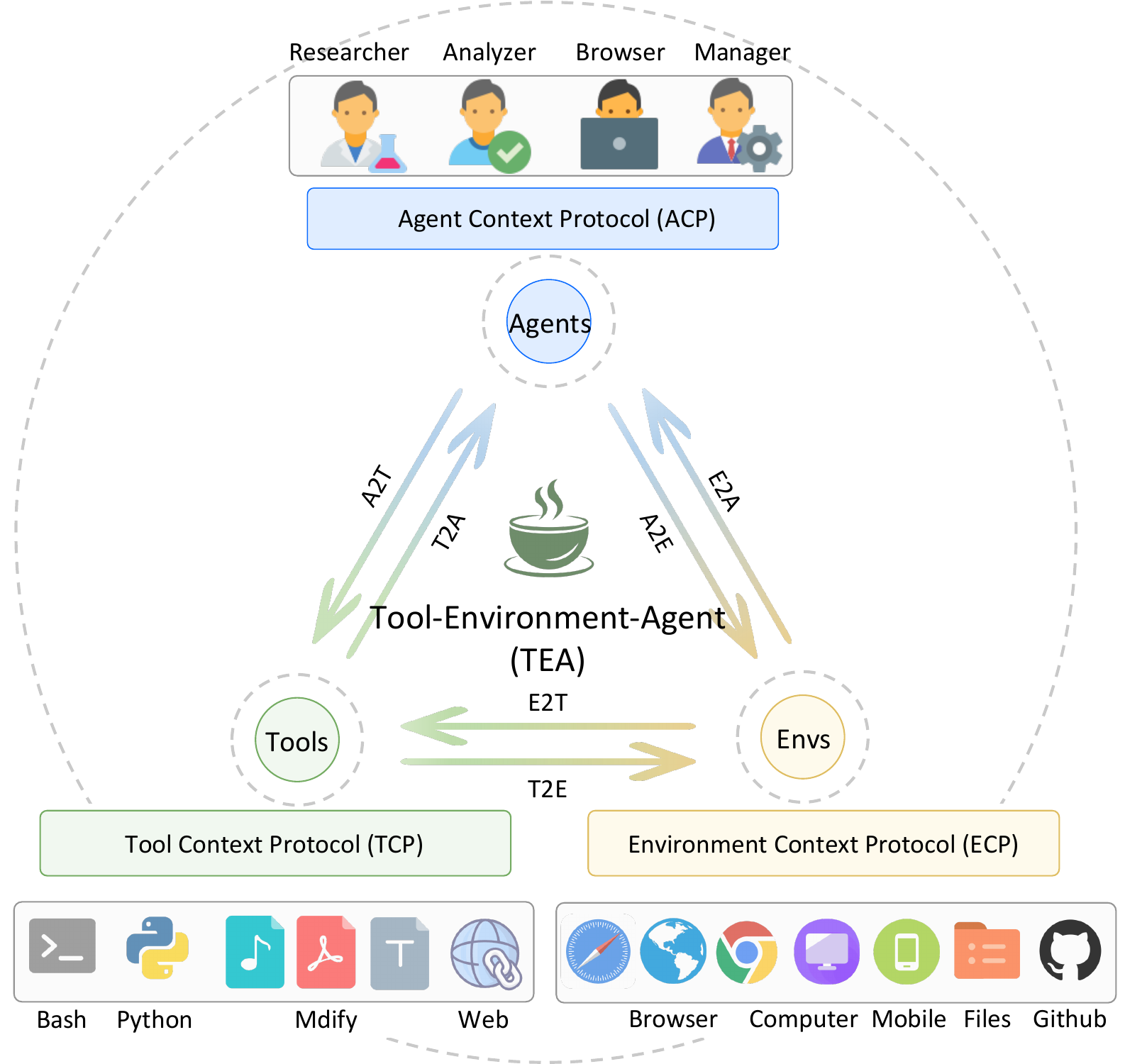}
  \caption{Architecture of the TEA Protocol.}
  \label{fig:tea_details}
\vspace{-0.6cm}
\end{figure}

The TEA Protocol is fundamentally designed around coroutine-based asynchronous execution, enabling concurrent task processing and parallel multi-agent coordination. As illustrated in Figure~\ref{fig:tea_details}, the protocol architecture comprises three primary layers: i) \textbf{Basic Managers} provide foundational services through six specialized components (model, prompt, memory, dynamic, version, and tracer); ii) \textbf{Core Protocols} define the Tool Context Protocol (TCP), Environment Context Protocol (ECP), and Agent Context Protocol (ACP), each implemented through a \textit{context manager} for lifecycle engineering and a \textit{server} for standardized orchestration; and iii) \textbf{Protocol Transformations} establish bidirectional conversion pathways (e.g., A2T, E2T, A2E) enabling dynamic role reconfiguration. Additionally, the protocol incorporates a \textbf{Self-Evolution Module} that wraps agent-associated components as evolvable variables for iterative optimization. Details and formalization can be found in Appendix~\ref{appx_sec:details_of_tea_protocol}.

\subsection{Basic Managers}
The Basic Managers constitute the foundation of the TEA Protocol, providing essential services through six specialized managers: i) the \textit{model manager} abstracts heterogeneous LLM backends through a unified interface; ii) the \textit{prompt manager} handles prompt lifecycle and versioning; iii) the \textit{memory manager} coordinates persistence via session-based concurrency control; iv) the \textit{dynamic manager} enables runtime code execution and serialization; v) the \textit{version manager} maintains evolution histories for all components; and vi) the \textit{tracer} records comprehensive execution trajectories and system-wide telemetry, serving as a data collection engine for audit, debugging, and the synthesis of high-quality datasets for agent training.

\subsection{Core Protocols}

The TEA Protocol defines three core context protocols: the \textbf{Tool Context Protocol} (TCP), the \textbf{Environment Context Protocol} (ECP), and the \textbf{Agent Context Protocol} (ACP). These protocols share a unified architectural design, each implemented through two core components: a \textit{context manager} for context engineering, lifecycle management, and semantic retrieval, and a \textit{server} that exposes standardized interfaces to other system modules. Each protocol generates a unified \textit{contract document} (analogous to Agent Skills~\citep{anthropic2025agentskills}) that aggregates all registered components' descriptions to facilitate resource discovery and usage.

\textbf{Tool Context Protocol.} TCP fundamentally extends MCP~\citep{anthropic2024mcp} by introducing integrated context engineering and comprehensive lifecycle management. Implemented through a \textit{ToolContextManager} and a \textit{TCPServer}, it supports seamless tool loading from both local registries and persistent configurations. During registration, TCP automatically synthesizes multiple representation formats, including function-calling schemas for LLM interfaces, natural language descriptions for documentation, and type-safe argument schemas for validation, providing LLMs with rich semantic information for accurate parameter inference. Furthermore, TCP incorporates a robust versioning system and a semantic retrieval mechanism based on vector embeddings, ensuring that tools can evolve over time while remaining easily discoverable through similarity-based queries.

\textbf{Environment Context Protocol.} ECP addresses the lack of unified interfaces in current agent systems by formalizing computational environments as first-class components with distinct \textit{observation} and \textit{action} spaces. Following an architectural pattern similar to TCP, it employs an \textit{EnvironmentContextManager} to maintain state coherence and manage the contextual execution environments required by tools. ECP automatically discovers and registers environment-specific actions, converting them into standardized interfaces that agents can invoke via action toolkits. This design enables agents to operate across heterogeneous domains, such as browsers or file systems, without bespoke adaptations, while leveraging versioning and semantic retrieval to manage environment-level capabilities.

\textbf{Agent Context Protocol.} ACP establishes a unified framework for the registration, representation, and orchestration of autonomous agents, overcoming the poor interoperability and fragmented attribute definitions in existing multi-agent systems. It utilizes an \textit{AgentContextManager} to maintain agent states and execution contexts, providing a foundation for persistent coordination across tasks and sessions. ACP captures semantically enriched metadata regarding agents' roles, competencies, and objectives, and formalizes the modeling of complex inter-agent dynamics, including cooperative, competitive, and hierarchical configurations. By embedding structured contextual descriptions and maintaining relationship representations, ACP facilitates adaptive collaboration and systematic integration within the broader TEA ecosystem.

\subsection{Protocol Transformations}

While TCP, ECP, and ACP provide independent specifications for tools, environments, and agents, practical deployment requires seamless interoperability across these protocols. Well-defined transformation pathways are essential for enabling computational components to assume alternative roles and exchange contextual information in a principled manner. These transformations constitute the foundation for dynamic role reconfiguration, allowing components to flexibly adapt their functional scope in response to evolving task requirements and system constraints. We identify six fundamental categories of protocol transformations:
\begin{itemize}[leftmargin=*,itemsep=0.1em]
  \item \textbf{Agent-to-Tool} (A2T). Encapsulates an agent's capabilities and reasoning into a standardized tool interface while preserving awareness. For example, a deep researcher workflow can be packaged as a general-purpose search tool.
  
  \item \textbf{Tool-to-Agent} (T2A). Treats tools as operational actuators by mapping an agent's goals into parameterized tool invocations, aligning reasoning with tool constraints. For example, a data analysis agent may invoke SQL tools to query structured databases.
  
  \item \textbf{Environment-to-Tool} (E2T). Converts actions of environments into standardized tool interfaces, enabling agents to interact with environments through consistent tool calls. For example, browser actions such as Navigate and Click can be consolidated into a context-aware toolkit.
  
  \item \textbf{Tool-to-Environment} (T2E). Elevates a collection of tools into an environment abstraction where functions become actions within a coherent action space governed by shared state. For example, a development toolkit can be encapsulated as a programming environment for sequential code-edit-compile-debug workflows.
  
  \item \textbf{Agent-to-Environment} (A2E). Encapsulates an agent as an interactive environment by exposing its decision rules and state dynamics as an operational context for other agents. For example, a market agent can be represented as an environment that provides trading rules and dynamic responses for training.
  
  \item \textbf{Environment-to-Agent} (E2A). Embeds reasoning and adaptive decision-making into an environment's dynamics, transforming it into an autonomous agent that can initiate behaviors and enforce constraints. For example, a game environment can be elevated into an opponent agent that adapts its strategy to the player's actions.
\end{itemize}

\subsection{Self-Evolution Module}
The Self-Evolution Module enables agents to continuously improve performance by optimizing system components during task execution. It wraps evolvable components, including prompts, tool/agent/environment/memory code, and successful execution solutions, as variables for iterative optimization. The module employs two primary methods: \textit{textgrad}~\citep{yuksekgonul2025optimizing} for gradient-based refinement and \textit{self-reflection} for strategic analysis. Optimized components are automatically registered as new versions via the version manager, ensuring that subsequent tasks leverage improved capabilities while maintaining access to historical records for analysis and rollback.

\begin{figure*}[tb]
  \vspace{-0.5cm}
  \centering
  \includegraphics[width=0.9\linewidth]{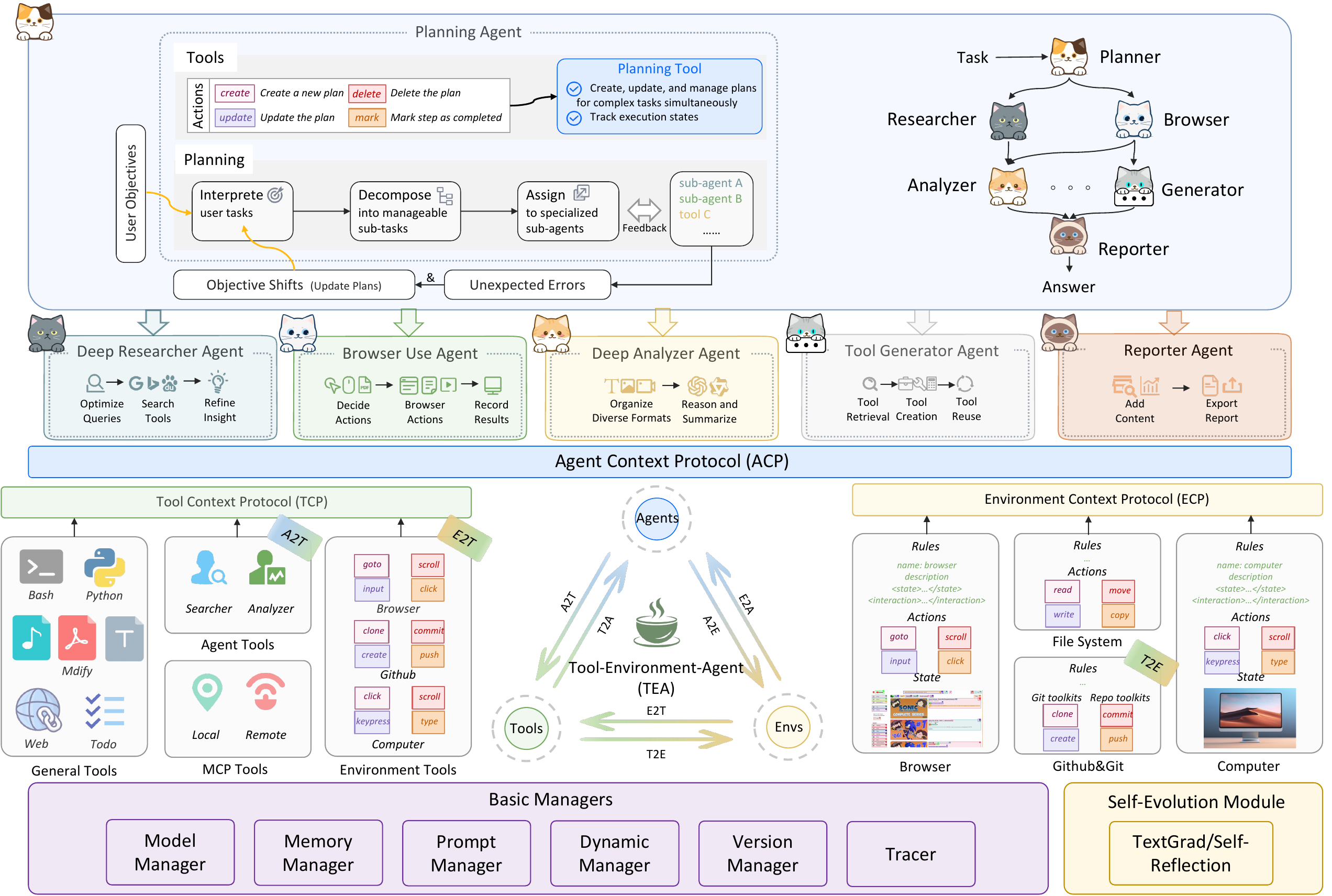}
  \caption{Architecture of \projectname implemented based on TEA protocol.}
  \label{fig:architecture}
  \vspace{-0.5cm}
\end{figure*}

\section{AgentOrchestra}
\label{sec:agentorchestra}

\projectname is a concrete instantiation of the TEA Protocol, designed as a hierarchical multi-agent framework that integrates high-level planning with modular agent collaboration. As illustrated in Figure~\ref{fig:architecture}, \projectname features a central planning agent that decomposes complex objectives and delegates sub-tasks to a team of specialized sub-agents. This section outlines our agent design principles and the architecture of both planning and specialized sub-agents. Details can be found in Appendix~\ref{app_sec:the_agentorchestra_framework_based_on_tea_protocol}.

\subsection{Agent Design Principles}
\label{sec:agent_design}
Within the TEA Protocol framework, agents are autonomous components that follow a structured interaction model with six core components. i) \textbf{Agent}: Managed via the ACP for registration and coordination. ii) \textbf{Environment}: External context and resources managed by the ECP, exposing unified interfaces for observation and action. iii) \textbf{Model}: LLM reasoning engines abstracted by the Basic Managers for model-agnostic interoperability and dynamic switching. iv) \textbf{Memory}: Session-based persistence that records trajectories and extracts reusable insights. v) \textbf{Observation}: The current context, including tasks, environment states, execution history, and available resources (tools and sub-agents). vi) \textbf{Action}: TCP-managed, executed via parameterized tool calls, where one tool may support multiple actions.

This architectural design facilitates a continuous \textit{perception--interpretation--action} cycle. The agent first perceives the current \textit{observation} and retrieves relevant context from \textit{memory}. It then interprets this information through the unified \textit{model} interface to determine the optimal \textit{action}. The action is executed within the managed \textit{environment}, and the resulting state transitions and insights are recorded back into memory to refine subsequent reasoning cycles. This iterative loop continues until the task objectives are satisfied or a termination condition is reached. Further details are provided in Appendix~\ref{appx_sec:agent_design_principles}.

\subsection{Planning Agent}
\label{sec:planning_agent}

The planning agent is the central orchestrator of \projectname. It interprets the user goal, decomposes it into sub-tasks, and dispatches them to specialized sub-agents or TCP tools via ACP-mediated communication while tracking global progress and consolidating intermediate feedback. To enable principled orchestration, it leverages long-term memory to guide resource selection and dynamically constructs a unified invocation interface, including resources produced through E2T and A2T transformations. Execution follows an iterative loop of interpretation, allocation, and action, with automatic replanning under environment shifts or execution failures. Session management and tracer-based logging provide auditability and support robust long-horizon task completion.

\subsection{Specialized Sub-Agents}
\label{sec:sub_agents}

To address diverse real-world challenges, \projectname instantiates specialized sub-agents tailored for task domains. These sub-agents are managed via the ACP and coordinate through the planning agent to execute complex workflows: i) \textbf{Deep Researcher Agent}: Specialized for comprehensive information gathering through multi-round research workflows. It performs parallel breadth-first searches across multiple engines and recursively issues follow-up queries until task objectives are satisfied, producing relevance-ranked, source-cited summaries. ii) \textbf{Browser Use Agent}: Provides automated, fine-grained web interaction by integrating both browser and computer environments under the ECP. It supports DOM-level and pixel-level operations (e.g., mouse movements), achieving unified control over interactive elements. iii) \textbf{Deep Analyzer Agent}: A workflow-oriented module designed for multi-step reasoning on heterogeneous multimodal data (e.g., text, PDFs, images, audio, video or zip). It applies type-specific analysis strategies and iterative refinement to synthesize insights into coherent conclusions. iv) \textbf{Tool Generator Agent}: Facilitates intelligent tool evolution through the automated creation, retrieval, and systematic reuse of TCP-compliant tools. It employs semantic search to identify tools and initiates a code synthesis process to develop new capabilities when gaps are identified. v) \textbf{Reporter Agent}: It aggregates and harmonizes evidence collected by upstream agents (e.g., the Deep Researcher Agent, Browser Use Agent, and Deep Analyzer Agent), then composes structured markdown with automatically deduplicated references and normalized URLs for consistent source attribution.

\section{Empirical Studies}
\label{sec:empirical_studies}
This section presents our experimental setup and results, including benchmark evaluations, baseline comparisons, and comprehensive analysis. Additional examples are provided in the Appendix~\ref{app_sec:case_study}.

\textbf{Experimental Settings}. We evaluate \projectname on four benchmarks: two expert-level agent benchmarks, \textbf{GAIA}~\cite{mialon2023gaiabenchmarkgeneralai} and \textbf{HLE}~\cite{phan2025humanity}, and two reasoning benchmarks, \textbf{AIME} and \textbf{GPQA-Diamond}~\cite{rein2024gpqa}. GAIA contains 301 test and 165 validation questions, and HLE contains 2,500 multimodal questions. We report pass@1. Unless otherwise specified, the planning agent (\(m{=}50\)), deep researcher (\(m{=}3\)), tool generator (\(m{=}10\)), deep analyzer (\(m{=}3\)), and reporter use \texttt{gemini-3-flash-preview}; the browser-use agent uses \texttt{gpt-4.1} (\(m{=}5\)) and \texttt{computer-use-preview(4o)} (\(m{=}50\)), where \(m\) denotes the maximum number of steps. In all benchmark tables, \textit{Vanilla} denotes the base system without self-evolution, and \textit{Evolved} denotes the variant with self-reflection-based evolution. For the scientific and mathematical benchmarks in Table~\ref{tab:results1}, we instantiate only the Deep Analyzer Agent, since deep web research and browser interaction are unnecessary; in this setting, self-evolution is restricted to the analyzer's prompt and solution traces. We compare five backbones spanning three reasoning tiers: two lower-reasoning models (\texttt{gpt-4o} and \texttt{gpt-4.1}), two medium-reasoning models (\texttt{claude-sonnet-4.5} and \texttt{grok-4.1-fast}), and one higher-reasoning model (\texttt{gemini-3-flash-preview}).

\subsection{Performance across Benchmarks}
\label{sec:results}

\begin{figure*}[h]
  \vspace{-0.5cm}
  \centering
  \includegraphics[width=\linewidth]{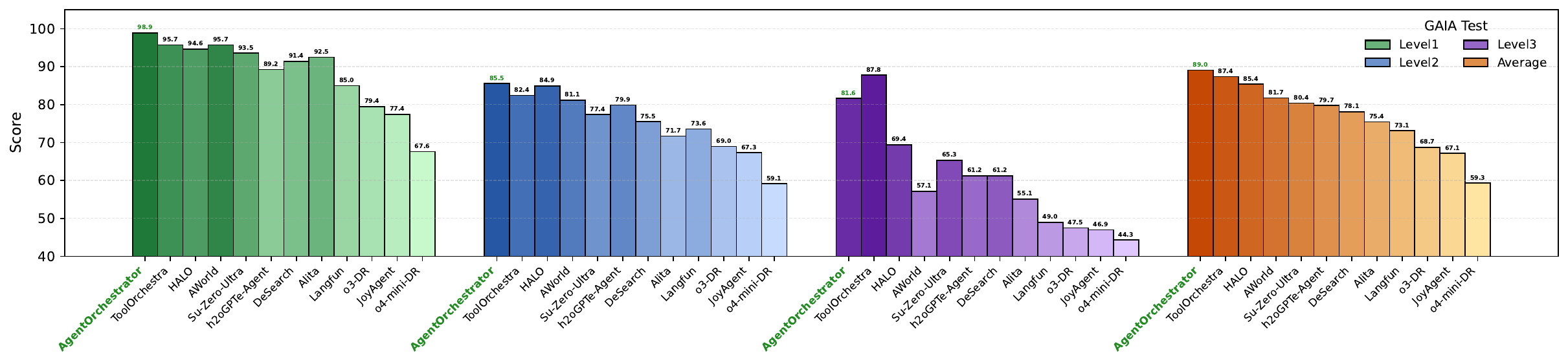}
  \vspace{-0.7cm}
  \caption{GAIA Test Results.}
  \label{fig:gaia_test}
  \vspace{-0.5cm}
\end{figure*}

\textbf{GAIA}. Figure~\ref{fig:gaia_test} shows that the \projectname \textit{Evolved} variant achieves 89.04\% average accuracy on the GAIA Test set, placing it among the leading reported methods rather than merely excelling on a single subset. More importantly, the gains remain strong on the harder partitions, reaching 85.53\% on Level 2 and 81.63\% on Level 3, which suggests that the framework scales beyond shallow retrieval-heavy cases. Table~\ref{tab:results2} further shows that \projectname \textit{Evolved} achieves state-of-the-art performance on the GAIA Validation set with 93.33\% average accuracy, improving over both strong external baselines such as agent-2030 and Alita (87.27\%) and our \textit{Vanilla} variant (89.70\%). Three findings emerge from these two results. First, the advantage is concentrated on the more agentically demanding settings rather than only on easy questions, indicating that hierarchical decomposition is most useful when the system must coordinate retrieval, interaction, and analysis over long horizons. Second, the improvement from \textit{Vanilla} to \textit{Evolved} is largest on Validation Level 2 (+5.26 points) while preserving the strongest Level 3 score, suggesting that self-evolution primarily strengthens robustness in medium-to-hard coordination regimes instead of overfitting to simple instances. Third, the combination of strong Test performance and Validation SOTA suggests that the benefit comes from system design rather than benchmark-specific tricks: TEA's explicit environment management reduces failure during cross-environment transitions, while online tool refinement helps recover from capability gaps that would otherwise stall execution. For future agent systems, these results suggest a clear design lesson: competitive GAIA performance requires not only stronger base models, but also better task decomposition, persistent environment state, and mechanisms for adaptive capability expansion during execution.

\begin{table}[t]
  \centering
  \renewcommand{\arraystretch}{0.8}
  \setlength{\tabcolsep}{3pt}
  \footnotesize
  \begin{tabular}{p{4cm}p{0.7cm}p{0.7cm}p{0.7cm}p{0.7cm}}
    \toprule
    \textbf{Agent} & \textbf{Level1} & \textbf{Level2} & \textbf{Level3} & \textbf{Avg.} \\
    \midrule
    \rowcolor{gray!15} \multicolumn{5}{c}{\textit{\textbf{Validation}}} \\
    HF ODR~\citep{huggingface_open_deep_research_2024} & 67.92 & 53.49 & 34.62 & 55.15 \\
    o3-DR~\citep{openai2025deepresearch}      & 74.29 & 69.06 & 47.60 & 67.36 \\
    DeSearch~\citep{desearchai2024desearch}   & 90.57 & 72.01 & 38.46 & 72.73 \\
    Co-Sight~\citep{zhang2025cosight}        & 86.79 & 73.26 & 42.31 & 72.73 \\
    Manus~\citep{shen2025mindmachinerisemanus} & 86.50 & 70.10 & 57.69 & 73.90 \\
    AWorld~\citep{yu2025aworld}              & 88.68 & 77.91 & 53.85 & 77.58 \\
    Langfun~\citep{google2024langfun}        & 88.68 & 80.23 & 57.69 & 79.39 \\
    agent-2030                               & 96.23 & 90.70 & 57.69 & 87.27 \\
    Alita~\citep{qiu2025alita}              & 88.68 & 89.53 & 76.92 & 87.27 \\
    \midrule
    \rowcolor{rowOurs}
    \projectname Vanilla                                  & 92.45 & 88.37 & 88.46 & 89.70 \\
    \rowcolor{rowOurs}
    \projectname Evolved                                & \textbf{96.23} & \textbf{93.02} & \textbf{88.46} & \textbf{93.33} \\
    \cmidrule{2-5}
    \textbf{Improvement(\%)} & \textcolor{green!60!black}{\textbf{4.09$\uparrow$}} & \textcolor{green!60!black}{\textbf{5.26$\uparrow$}} & \textcolor{green!60!black}{0.00} & \textcolor{green!60!black}{\textbf{4.05$\uparrow$}} \\
    \bottomrule
  \end{tabular}
  \vspace{-0.2cm}
  \caption{GAIA Validation Results.}
  \label{tab:results2}
\vspace{-0.7cm}
\end{table}

\begin{table}[t]
\centering
\renewcommand{\arraystretch}{0.3}
\setlength{\tabcolsep}{8pt}
\footnotesize
\begin{tabular}{lcc}
\toprule
\textbf{Agent} & \textbf{Open} & \textbf{Score} \\
\midrule
Gemini 3 Pro            & --          & 45.8 \\
Gemini Deep Research    & --          & 46.4 \\
\rowcolor{rowOpen}
Yunjue Agent            & \checkmark  & 48.0 \\
\rowcolor{rowOpen}
MiMo-V2.5-Pro           & \checkmark  & 48.0 \\
\rowcolor{rowOpen}
DeepSeek-V4-Pro-Max     & \checkmark  & 48.2 \\
Claude Sonnet 4.6       & --          & 49.0 \\
GPT-5.2 Pro             & --          & 50.0 \\
\rowcolor{rowOpen}
Kimi K2.5 Thinking      & \checkmark  & 50.2 \\
Muse Spark              & --          & 50.4 \\
Gemini 3.1 Pro          & --          & 51.4 \\
GPT-5.4                 & --          & 52.1 \\
GPT-5.5                 & --          & 52.2 \\
Zoom Federated AI       & --          & 53.0 \\
Claude Opus 4.6         & --          & 53.1 \\
Gemini 3 Deep Think     & --          & 53.4 \\
\rowcolor{rowOpen}
Kimi K2.6 Thinking      & \checkmark  & 54.0 \\
Claude Opus 4.7         & --          & 54.7 \\
Poetiq Meta-System      & --          & 55.0 \\
GPT-5.5 Pro             & --          & 57.2 \\
Muse Spark (Contemplating) & --       & 58.4 \\
GPT-5.4 Pro             & --          & 58.7 \\
Claude Mythos Preview   & --          & \textbf{64.7} \\
\midrule
\rowcolor{rowOurs}
\projectname Vanilla           & \checkmark & 55.2 \\
\rowcolor{rowOurs}
\projectname Evolved           & \textbf{\checkmark} & 59.6 \\
\cmidrule{3-3}
\textbf{Improvement(\%)}       & & \textcolor{green!60!black}{\textbf{7.97$\uparrow$}} \\
\bottomrule
\end{tabular}
\caption{Leaderboard of the HLE benchmark for Agents with Tools~\cite{zoom2025hleleaderboard}. Shaded rows mark open-source systems.}
\label{tab:hle_full_set_results}
\vspace{-0.5cm}
\end{table}

\textbf{HLE}. Table~\ref{tab:hle_full_set_results} shows that \projectname \textit{Evolved} achieves 59.6 on HLE, improving over \textit{Vanilla} by 7.97\% and substantially outperforming a broad set of strong agent systems, including open-source methods such as Kimi K2.6 Thinking (54.0) and proprietary systems such as GPT-5.5 Pro (57.2) and GPT-5.4 Pro (58.7). While Claude Mythos Preview remains the strongest overall system at 64.7, our result places \projectname among the leading entries on this expert-level benchmark and establishes a particularly strong position among open systems. Three findings emerge from this comparison. First, HLE appears to reward systems that can sustain structured reasoning over long solution trajectories rather than relying on isolated retrieval or short-horizon tool use; this is consistent with our hierarchical design, which allows the Planning Agent to preserve global coherence while specialized agents validate technical details. Second, the improvement from \textit{Vanilla} to \textit{Evolved} suggests that self-evolution mainly strengthens domain-specific problem-solving strategies and error correction, rather than merely improving surface-level evidence collection. Third, the remaining gap to the top proprietary system indicates that expert benchmarks such as HLE still stress the ceiling of frontier model capability, but the leaderboard also suggests a practical lesson for future agent design: competitive performance increasingly depends on combining strong base models with explicit role decomposition, iterative verification, and adaptive capability refinement.

\begin{table}[h]
\footnotesize
\centering
\renewcommand{\arraystretch}{0.6}
\setlength{\tabcolsep}{4pt}
\begin{tabular}{p{2.8cm}p{1.3cm}p{1.3cm}p{1.3cm}}
\toprule
\textbf{Agent} & \textbf{GPQA} & \textbf{AIME24} & \textbf{AIME25} \\
\midrule
\rowcolor{gray!15} \multicolumn{4}{c}{\textit{\textbf{gpt-4o}}} \\
Vanilla        & 47.98 & 13.34 & 6.67 \\
Evolved        & 58.08 & 16.67 & 13.34 \\
\cmidrule{2-4}
\textbf{Improvement(\%)} & \textcolor{green!60!black}{\textbf{21.05$\uparrow$}} & \textcolor{green!60!black}{\textbf{24.97$\uparrow$}} & \textcolor{green!60!black}{\textbf{100$\uparrow$}} \\
\midrule
\rowcolor{gray!15} \multicolumn{4}{c}{\textit{\textbf{gpt-4.1}}} \\
Vanilla        & 65.15 & 23.34 & 20.00 \\
Evolved        & 67.67 & 40.00 & 33.33 \\
\cmidrule{2-4}
\textbf{Improvement(\%)} & \textcolor{green!60!black}{\textbf{3.87$\uparrow$}} & \textcolor{green!60!black}{\textbf{71.38$\uparrow$}} & \textcolor{green!60!black}{\textbf{66.65$\uparrow$}} \\
\midrule
\rowcolor{gray!15} \multicolumn{4}{c}{\textit{\textbf{grok-4.1-fast}}} \\
Vanilla        & 83.33 & 96.67 & 90.00 \\
Evolved        & 89.34 & 96.67 & 96.67 \\
\cmidrule{2-4}
\textbf{Improvement(\%)} & \textcolor{green!60!black}{\textbf{7.21$\uparrow$}} & \textcolor{green!60!black}{0.00} & \textcolor{green!60!black}{\textbf{7.41$\uparrow$}} \\
\midrule
\rowcolor{gray!15} \multicolumn{4}{c}{\textit{\textbf{claude-sonnet-4.5}}} \\
Vanilla        & 78.28 & 76.67 & 73.33 \\
Evolved        & 81.44 & 86.67 & 90.00 \\
\cmidrule{2-4}
\textbf{Improvement(\%)} & \textcolor{green!60!black}{\textbf{4.04$\uparrow$}} & \textcolor{green!60!black}{\textbf{13.04$\uparrow$}} & \textcolor{green!60!black}{\textbf{22.73$\uparrow$}} \\
\midrule
\rowcolor{gray!15} \multicolumn{4}{c}{\textit{\textbf{gemini-3-flash-preview}}} \\
Vanilla        & 88.38 & 83.33 & 83.33 \\
Evolved        & 90.40 & 93.33 & 93.33 \\
\cmidrule{2-4}
\textbf{Improvement(\%)} & \textcolor{green!60!black}{\textbf{2.28$\uparrow$}} & \textcolor{green!60!black}{\textbf{12.00$\uparrow$}} & \textcolor{green!60!black}{\textbf{12.00$\uparrow$}} \\
\bottomrule
\end{tabular}
\vspace{-0.2cm}
\caption{GPQA and AIME Results.}
\label{tab:results1}
\vspace{-0.2cm}
\end{table}

\textbf{GPQA and AIME}. Table~\ref{tab:results1} evaluates self-evolution in a stripped-down setting with minimal agentic coordination, thereby focusing primarily on reasoning quality. Three observations emerge. First, the gains are highly consistent across model families: the \textit{Evolved} variant improves GPQA on all five backbones and improves AIME25 on four of them, with the only flat result appearing on AIME24 for \texttt{grok-4.1-fast}, where the base score is already near saturation. This pattern suggests that self-reflection-based evolution is not tied to a single model provider or capability regime. Second, the largest gains occur on the lower- and medium-reasoning backbones, especially on the harder mathematical settings. For example, \texttt{gpt-4.1} improves by 71.38\% on AIME24 and 66.65\% on AIME25, while \texttt{gpt-4o} doubles its AIME25 score. This is consistent with the view that evolving prompts and solution traces can strengthen decomposition and verification in models with weaker baseline reasoning. Third, stronger backbones still benefit, although more selectively: \texttt{grok-4.1-fast}, \texttt{claude-sonnet-4.5}, and \texttt{gemini-3-flash-preview} all improve on GPQA and/or AIME25. Taken together, these results suggest that substantial gains on reasoning benchmarks can be obtained by evolving the reasoning interface and reusable solution patterns of a single strong analyzer, even without introducing additional agent types.

\subsection{Ablation Studies}

\begin{table}[t]
  \centering
  \vspace{-0.2cm}
  \footnotesize
  \setlength{\tabcolsep}{1.0pt}
  \renewcommand{\arraystretch}{0.8}
  \begin{tabular}{c c c c c | c c c c | c}
  \toprule
  P & R & B & A & T & Level 1 & Level 2 & Level 3 & Average & Improvement \\
  \midrule
  \checkmark &                &             &               &              & 54.84 & 33.96 & 10.20 & 36.54 & -- \\
  \checkmark & \checkmark     &             &               &              & 86.02 & 47.17 & 34.69 & 57.14 & \textcolor{green!60!black}{\textbf{+56.40\%}} \\
  \checkmark & \checkmark     & \checkmark  &               &              & 89.25 & 71.07 & 46.94 & 72.76 & \textcolor{green!60!black}{\textbf{+27.33\%}} \\
  \checkmark & \checkmark     & \checkmark  & \checkmark    &              & 91.40 & 77.36 & 61.22 & 79.07 & \textcolor{green!60!black}{\textbf{+8.67\%}} \\
  \checkmark & \checkmark     & \checkmark  & \checkmark    & \checkmark   & 98.92 & 85.53 & 81.63 & 89.04 & \textcolor{green!60!black}{\textbf{+12.61\%}} \\
  \bottomrule
  \end{tabular}
  \caption{Sub-agent effectiveness across GAIA Test.}
  \label{tab:effectiveness_of_sub_agents}
  \vspace{-0.5cm}
\end{table}

\textbf{Effectiveness of the specialized sub-agents}. Table~\ref{tab:effectiveness_of_sub_agents} yields three observations. First, information acquisition is the dominant early bottleneck on GAIA: adding the Deep Researcher raises the average score from 36.54 to 57.14, and adding the Browser Use Agent further increases it to 72.76, indicating that broad evidence gathering and fine-grained web interaction are complementary in long-horizon agentic tasks. Second, the Deep Analyzer improves the average score from 72.76 to 79.07 and boosts Level 3 from 46.94 to 61.22, suggesting that the hardest multimodal cases require dedicated reasoning beyond retrieval and interaction alone. Third, the Tool Generator provides the largest late-stage gain, improving the average score from 79.07 to 89.04 and Level 3 from 61.22 to 81.63, which highlights the importance of on-demand capability expansion once strong planning, retrieval, interaction, and analysis are already in place. Overall, this cumulative ablation suggests that strong GAIA performance emerges from the staged composition of complementary agent roles rather than from any single module in isolation.

\textbf{Efficiency analysis}. We observe a clear scaling trend between task complexity and system cost. Simple tasks typically complete within 30 seconds using approximately 5k tokens, while medium-complexity tasks require around 3 minutes and 25k tokens. Complex multimodal or long-horizon scenarios take roughly 10 minutes and 100k tokens. This pattern suggests that the hierarchical architecture allocates expensive reasoning and tool use primarily to harder sub-problems, while keeping simpler cases relatively lightweight. In practice, extra latency and token cost mainly arise for tasks requiring broader evidence, deeper multimodal analysis, or longer interactions.

\textbf{Effectiveness of self-evolution and tool evolution}. As shown above, self-evolution remains beneficial even in a stripped-down setting with only the Deep Analyzer Agent, suggesting that part of the gain comes from improving reasoning structure itself rather than adding more agentic actions. At the same time, the Tool Generator Agent demonstrates efficient capability expansion within the TCP framework: during evaluation, it autonomously generated over 50 specialized tools and achieved a 30\% reuse rate across subsequent tasks. Together, these results show that TEA jointly improves reasoning components while continually expanding executable capabilities.

\section{Conclusion}
\label{sec:conclusion}
We introduced the TEA Protocol, a unified framework that models environments, agents, and tools as first-class, evolvable resources. Building on TEA, we presented \projectname, a hierarchical multi-agent system that combines explicit protocol design, specialized coordination, and self-evolution. Experiments on four benchmarks demonstrate strong and consistent performance across both expert-level agent tasks and scientific and mathematical reasoning settings. Future work will explore more adaptive role allocation and deeper forms of self-evolution.

\newpage
\section{Limitations}

\subsection{Limitations of TEA Protocol and AgentOrchestra}

Despite its strengths in orchestrating multi-agent systems, \projectname has several limitations that provide directions for future research:

First, \textbf{System Complexity and Learning Curve}. The TEA protocol introduces a structured abstraction layer for tools, environments, and agents to ensure interoperability. However, this structure may present a steeper learning curve for developers compared to simpler, ad-hoc scripting methods. To address this, we will provide extensive documentation, interactive tutorials, and a variety of pre-configured templates to simplify the onboarding process.

Second, \textbf{Communication and Execution Overhead}. Standardizing interactions through a formal protocol can introduce marginal computational and communication overhead, potentially increasing latency in real-time applications. We plan to optimize the serialization protocols and explore asynchronous execution models to minimize these effects in future versions.

Third, \textbf{Dependence on Underlying Model Capabilities}. The effectiveness of the orchestration is inherently limited by the reasoning and instruction-following performance of the foundation LLMs used. While TEA provides a robust framework, it cannot fully compensate for failures caused by model hallucinations or poor tool-use logic. Future work will focus on developing model-agnostic error recovery strategies and more sophisticated validation layers to enhance system-wide resilience.

\subsection{Potential Risks}

While \projectname and the TEA protocol aim to enhance multi-agent productivity, their capability to interact with local environments and web browsers introduces certain ethical and security risks. 

One primary concern is the \textbf{Misuse for Malicious Automation}. The framework's flexibility in controlling browser sessions and executing terminal commands could be repurposed to develop unauthorized "plugins" or "cheats" for online platforms, leading to unfair advantages or automated fraud. Furthermore, there are significant \textbf{Privacy and Security Risks} associated with granting autonomous agents access to personal data or sensitive system resources. If not properly sandboxed or governed by strict security policies, an agent could inadvertently leak private information or perform harmful, irreversible system actions. To mitigate these risks, we emphasize that \projectname should be used within isolated, monitored environments, and we advocate for the integration of robust human-in-the-loop verification mechanisms and strict access control policies in any real-world deployment.

\bibliography{custom}

\clearpage
\newpage
\appendix
\appendixpage
\label{sec:appendix}
\section{Comprehensive Motivation for TEA Protocol}
\label{appx_sec:comprehensive_motivation}

This section provides a comprehensive motivation for the TEA Protocol by examining the fundamental relationships and transformations between agents, environments, and tools in multi-agent systems. The discussion is organized into two main parts: first, we explore the conceptual relationships between agents, environments, and tools, examining how these three fundamental components interact and complement each other in modern AI systems; second, we analyze why transformation relationships between these components are necessary, demonstrating the need for their conversion and integration through the TEA Protocol to create a unified, flexible framework for general-purpose task solving.

\subsection{Conceptual Relationships}

\subsubsection{Environment}

The environment constitutes one of the fundamental components of multi-agent systems, providing the external stage upon which agents perceive, act, and accomplish tasks. Within the context of the TEA Protocol, highlighting the role of environments is crucial, since environments not only define the operational boundaries of agents but also exhibit complex structural and evolutionary properties. In what follows, we outline the motivation for explicitly modeling environments in the TEA framework from several perspectives.  

\textbf{Classification of environments.} From a broad perspective, environments can be divided into two categories: the real world and the virtual world. The real world is concrete and directly perceivable by humans, such as kitchens, offices, or factories. By contrast, the virtual world cannot be directly perceived or objectively described by humans, including domains such as the network world, simulation platforms, and game worlds. Importantly, these two types of environments are not independent. Rather, they are tightly coupled through physical carriers, such as computers, displays, keyboards, mice, and sensors, which act as mediators that enable the bidirectional flow of information between the real and virtual domains. Hence, environments should be regarded not as isolated domains but as interdependent layers connected through mediating carriers.  

\textbf{Nested and expandable properties.} Environments are inherently nested and expandable. For example, when an individual is situated in a kitchen, their observable range and available tools are restricted to kitchen-related objects such as faucets, knives, and microwaves, all governed by the local rules of that sub-environment. When the activity range extends to the living room, new objects such as televisions, remote controls, and chairs become accessible, while the kitchen remains embedded as a sub-environment within a broader space. Furthermore, environments can interact with one another, as when a bottle of milk is taken from the kitchen to the living room. This demonstrates that enlarged environments can be conceptualized not merely as simple unions, but rather as structured integrations of the state and action spaces of smaller constituent environments, where local rules and affordances are preserved while new forms of interaction emerge from their composition.

\textbf{Relationship with state–action spaces.} In reinforcement learning, environments are formalized in terms of state and action spaces. The state space comprises the set of possible environmental states, represented in modalities such as numerical values, text, images, or video. The action space denotes the set of operations available to agents, generally divided into continuous and discrete spaces. Real and virtual environments are naturally continuous, but discrete abstractions are often extracted for the sake of tractability, forming the basis of most reinforcement learning systems. However, this discretization constrains the richness of interaction. In contrast, large language models (LLMs) enable a new paradigm: instead of selecting from a discrete set, LLMs can generate natural language descriptions that encode complex action sequences. These outputs can be understood as an intermediate representation between continuous and discrete action spaces, richer and more expressive than discrete actions, yet still mappable to concrete operations in continuous environments. To realize this mapping, intermediate actions are required as bridges. For instance, the natural language command ``boil water'' can be decomposed into executable steps such as turning on the kettle, filling it with water, powering it on, and waiting until boiling. This property indicates that LLM-driven interaction expands the definition of action representations and broadens the scope of environmental engagement.  

\textbf{Mediation and interaction.} The notion of mediation highlights that environments are not static backdrops but relative constructs whose boundaries depend on available carriers and interfaces. In hybrid physical–virtual systems, for example, Internet-of-Things (IoT) devices serve as mediators: a smart refrigerator in the physical world can be controlled through a mobile application in the virtual world, while the application itself is subject to network protocols. Consequently, the definition of an environment is dynamic and conditioned by interactional means. In the TEA Protocol, this mediation must be explicitly modeled, since it determines accessibility and interoperability across environments.  

\textbf{Toward intelligent environments.} Traditionally, environments are passive components that provide states and respond to actions. However, as embedded simulators, interfaces, and actuators grow more sophisticated, environments may gradually acquire semi-agentic properties. For instance, a smart home environment may not only respond to the low-level command ``turn on the light'' but also understand and execute a high-level instruction such as ``create a comfortable atmosphere for reading,'' by autonomously adjusting lighting, curtains, and background music. This trend suggests that environments are evolving from passive contexts into adaptive and cooperative components.  

In conclusion, the environment should not be regarded as a passive backdrop for agent activity, but as a dynamic and evolving component that fundamentally shapes the scope and feasibility of interaction. Its dual nature across real and virtual domains, its nested and compositional structure, and its formalization through state–action spaces all demonstrate that environments provide both the constraints and the affordances within which agents operate. At the same time, the rise of LLM-based agents introduces new forms of action representation that require environments to support more flexible, language-driven interfaces. Looking ahead, as environments increasingly incorporate adaptive and semi-agentic features, their role in task execution will only become more central. Within the TEA Protocol, this motivates treating environments as a co-equal pillar alongside agents and tools, ensuring that general-purpose task solving remains both grounded in environmental constraints and empowered by environmental possibilities.

\subsubsection{Agent}

Within the TEA Protocol, the motivation for treating agents as a core component alongside environments and tools extends beyond mere terminological convenience. Agents represent the indispensable connective tissue between the generative capabilities of LLMs, the operational affordances of tools, and the structural dynamics of environments. While environments provide the stage on which tasks unfold and tools extend the range of possible actions, it is agents that unify perception, reasoning, and execution into coherent task-solving processes. Without explicitly recognizing agents as an independent pillar, the TEA Protocol would lack a systematic way to explain how abstract linguistic outputs can be transformed into grounded operations, how tools can be selected and orchestrated, and how autonomy, memory, and adaptivity emerge in multi-agent systems. The following dimensions illustrate why agents must be elevated to a core component of the framework.

\textbf{Necessity of environment interaction.} Unlike large language models (LLMs), which only produce textual descriptions that require conversion into executable actions, agents are fundamentally characterized by their ability to directly interact with environments. While LLMs can generate detailed plans, instructions, or hypotheses, such outputs remain inert unless they are translated into concrete operations that affect the state of an environment. This gap between symbolic reasoning and actionable execution highlights the necessity of an intermediate entity capable of grounding abstract instructions into domain-specific actions. Agents fulfill precisely this role: they map language-level reasoning to executable steps, whether in physical settings, such as controlling robotic arms or sensors, or in virtual contexts, such as interacting with databases, APIs, or software systems.  

By serving as this mapping layer, agents enable the closure of full task loops, where perception leads to reasoning, reasoning produces plans, and plans culminate in actions that in turn modify the environment. Without explicitly modeling agents, the process would remain incomplete, as LLMs alone cannot guarantee the translation of reasoning into operational change. Within the TEA Protocol, this necessity justifies the elevation of agents to a core component: they provide the indispensable interface that connects the generative capacities of LLMs with the affordances and constraints of environments, ensuring that tasks are not only conceived but also carried through to completion.  

\textbf{The decisive role of non-internalizable tools.}  
The fundamental distinction between LLMs and agents lies in whether they can effectively employ tools that cannot be internalized into model parameters. Some tools can indeed be absorbed into LLMs, particularly those whose logic can be fully simulated in symbolic space, whose inputs and outputs are representable in language or code, and whose patterns fall within the training distribution (for example, mathematical reasoning, structured text formatting, code generation, and debugging). For example, early LLMs struggled with JSON output formatting and code reasoning, often requiring external correction or checking tools, but reinforcement learning (RL) and supervised fine-tuning (SFT) have progressively enabled such capabilities to be internalized.  

In contrast, many tools remain non-internalizable because they are intrinsically tied to environmental properties. These include tools that depend on physical devices such as keyboards, mice, and robotic arms, external infrastructures such as databases and APIs, or proprietary software governed by rigid protocols. Two recent approaches further illustrate this limitation. Vision-language-action (VLA)~\citep{black2025pi0} models map perceptual inputs directly into actions, which may appear to bypass intermediate symbolic descriptions, yet the resulting actions must still be aligned with the discrete action spaces of environments. This alignment represents not a fundamental internalization but a compromise, adapting model outputs to the constraints of environmental action structures. Similarly, the upgraded function calling mechanism introduced after GPT-5, which incorporates context-free grammar (CFG)~\citep{contextfreegrammar2025}, allows LLMs to output structured and rule-based actions that conform to external system requirements. However, this remains a syntactic constraint on model outputs, effectively providing a standardized interface to external systems rather than a truly internalized ability of the model.  

Agents therefore play a decisive role in mediating this boundary. They allow LLMs to internalize symbolic tools, thereby enhancing reasoning and self-correction, while also orchestrating access to non-internalizable tools through external mechanisms. This dual pathway ensures that LLMs are not confined to their parameterized capabilities alone but can extend into broader operational domains. In this way, agents transform the tension between internalizable and non-internalizable tools from a limitation into an opportunity, enabling robust problem solving in multimodal, embodied, and real-world contexts.  

\textbf{Memory and learning extension.} Another crucial motivation for agents lies in their capacity to overcome the intrinsic memory limitations of LLMs. Due to restricted context windows, LLMs struggle to maintain continuity across extended interactions or to accumulate knowledge over multiple sessions. Agents address this shortcoming by incorporating external memory systems capable of storing, retrieving, and contextualizing past experiences. Such systems simulate long-term memory and enable experiential learning, allowing agents to refine strategies based on historical outcomes rather than treating each interaction as isolated. However, in the TEA Protocol, memory is not defined as a core protocol component but is instead positioned at the infrastructure layer. This design choice reflects the anticipation that future LLMs may gradually internalize memory mechanisms into their parameters, thereby reducing or even eliminating the need for external memory systems. In other words, while memory expansion is indispensable for today’s agents, it may represent a transitional solution rather than a permanent defining element of agency.  

\textbf{Bridging virtual and external worlds.} It has been suggested that LLMs encode within their parameters a kind of ``virtual world,'' enabling them to simulate reasoning and predict outcomes internally. However, without an external interface, such simulations remain trapped in closed loops of self-referential inference, disconnected from the contingencies of real-world environments. Agents play a critical role in bridging this gap: they translate the abstract reasoning of LLMs into concrete actions, validate outcomes against environmental feedback, and close the loop between perception, reasoning, and execution. This bridging function transforms LLMs from purely linguistic engines into operationally grounded components whose outputs can be tested, refined, and extended within real or simulated environments.  

\textbf{Autonomy and goal-directedness.} Beyond reactivity, agents are motivated by their capacity for autonomy. While LLMs typically operate in a reactive fashion, producing outputs in response to explicit prompts, agents can adopt proactive behaviors. They are capable of formulating subgoals, planning action sequences, and dynamically adapting strategies in light of environmental changes or task progress. This goal-directedness is what elevates agents from passive tools into active participants in problem solving. Autonomy ensures that agents are not merely executing instructions but are able to pursue objectives, adjust course when facing uncertainty, and coordinate with other agents. Such properties are essential for multi-agent collaboration and for tackling open-ended, general-purpose tasks that require initiative as well as adaptability.  

Taken together, these motivations highlight why agents must be modeled as a core pillar of the TEA Protocol. Environments provide the stage for interaction, tools expand the operational scope, but it is agents that integrate reasoning, memory, tool usage, and autonomy into cohesive systems of action. By serving as mediators between LLMs and their environments, agents ensure that abstract reasoning is translated into grounded execution, enabling robust and scalable task solving across domains. In this sense, agents represent the crucial entity that transforms language models from passive predictors into active problem solvers within a unified multi-agent framework.   

\subsubsection{Tool}

Within the TEA Protocol, the decision to treat tools as a core component alongside environments and agents extends far beyond a matter of convenience in terminology. Tools represent the crucial mediating constructs that encapsulate and operationalize the action spaces of environments, while simultaneously serving as the primary extension layer of agent capabilities. Environments provide the structural stage on which interactions occur, and agents embody the reasoning and decision-making mechanisms that drive behavior, but it is through tools that such reasoning becomes executable and scalable. Without tools, agents would be confined to abstract planning or primitive environmental actions, and environments would remain underutilized as passive backdrops rather than dynamic arenas of transformation. 

Moreover, tools play a unique role in bridging symbolic reasoning and concrete execution, providing the abstraction layers necessary to decompose complex tasks into manageable units, and enabling cross-domain transfer through their modularity and portability. They also reveal the shifting boundary between what can be internalized into an agent’s parameters and what must remain external, highlighting the evolving interplay between intelligence and embodiment. In this sense, tools are not merely auxiliary aids but indispensable pillars that shape the architecture of multi-agent systems. The following dimensions illustrate the motivations for elevating tools to a core component of the TEA.

\textbf{Extending the operational boundary.}  
The primary function of tools is to expand the operational scope of agents beyond what is directly encoded in model parameters or supported by immediate environment interactions. Environments by themselves typically offer only primitive actions, and LLMs by themselves are limited to symbolic reasoning. Tools bridge this gap by furnishing additional pathways for action, allowing agents to manipulate physical artifacts or virtual systems in ways that exceed the direct expressive capacity of the model. From physical devices such as hammers, keyboards, and robotic arms to virtual infrastructures such as databases, APIs, and code execution engines, tools multiply the modes through which agents can influence their environments. Without tools, agents would be confined to intrinsic reasoning and the primitive action space of environments, leaving them incapable of executing tasks that require domain-specific operations. With tools, however, complex objectives can be decomposed into modular operations that are both tractable and reusable. This decomposition makes problem solving significantly more efficient, while also enhancing adaptability across domains. In this way, tools act as multipliers of agency, transforming abstract reasoning into a wider range of tangible interventions.  

\textbf{Hierarchy and abstraction.}  
Tools are not flat or uniform components but exhibit a hierarchical and abstract structure. At the lowest level, tools correspond to atomic environmental actions, such as ``clicking a button'' or ``moving one step.'' These atomic units can then be combined into higher-level compound tools such as ``opening a file'' or ``conducting a search.'' At an even higher level, compound tools may evolve into strategy-like constructs, such as ``writing a report,'' ``planning a trip,'' or ``completing a financial transaction.'' Each level builds upon the previous, creating a hierarchy of reusable capabilities. This hierarchical structure is not only efficient but also central to interpretability. Higher-level tools inherently carry semantic labels that communicate their function, which in turn makes agent behavior more transparent to human observers and more predictable to other agents. Such abstraction layers reduce the cognitive and computational load on the agent when planning, since invoking a high-level tool can encapsulate dozens or hundreds of low-level steps. Moreover, in multi-agent systems, the semantic richness of high-level tools serves as a lingua franca, facilitating coordination and collaboration.  

\textbf{Boundary between tools and agent capabilities.}  
The relationship between tools and agents is dynamic rather than static. As LLM reasoning and learning capabilities improve, certain tools can be gradually internalized into model parameters, effectively transforming into latent agent abilities. Examples include logical inference, grammar correction, structured text formatting, and code generation, which once required external support but have increasingly been subsumed into the model’s intrinsic skills. In this sense, the boundary between what is a ``tool'' and what is an ``ability'' is fluid and shaped by the trajectory of model development. By contrast, many tools remain non-internalizable because they are tightly coupled with environmental properties or external infrastructures. These include robotic arm manipulation, database queries, API interactions, and other operations that inherently depend on external systems or physical substrates. This duality creates a layered conception of agency: a ``core capability layer'' composed of skills internalized within the model, and an ``extended layer'' realized through external tool use. The shifting line between these two layers reflects the ongoing negotiation between intelligence and embodiment, highlighting why tools must be explicitly recognized as a structural component.  

\textbf{Evolution and portability.}  
Tools are not static constructs but evolve alongside environments and agent requirements. In programming contexts, for instance, an initial tool may simply execute code. Over time, as demands increase, this basic function evolves into more advanced utilities such as ``static code analysis,'' ``automated test generation,'' and ``continuous deployment.'' A similar trajectory occurs in other domains, where rudimentary tools gradually give rise to sophisticated pipelines capable of handling more complex and specialized tasks. In addition to evolution, tools are inherently portable. A well-designed summarization tool, for example, can be reused across very different contexts, from condensing news articles to producing academic literature reviews. This reusability makes tools a natural vehicle for cross-domain generalization, enabling knowledge and functionality to transfer without retraining the underlying model. For these reasons, the TEA Protocol emphasizes modularization and standardization of tools, ensuring that they can evolve flexibly while maintaining interoperability across agents and environments.  

\textbf{Toward intelligent tools.}  
Traditional tools are passive, executing predefined functions only when invoked by an agent. They wait for explicit instructions and do not adapt to context or anticipate needs. However, the trajectory of tool development points toward increasing intelligence, where tools exhibit perception, analysis, and even limited decision-making capabilities. For example, an advanced debugging tool may not only check code upon request but also proactively scan for hidden vulnerabilities, propose optimizations, and even prioritize issues based on estimated risk. Such capabilities blur the line between tools and agents, effectively creating semi-agentic components. Intelligent tools can share responsibility for decision making, reduce the supervisory burden on agents, and participate in distributed problem-solving processes. In this way, tools transition from being passive executors to collaborative partners, altering the topology of multi-agent systems and reshaping the balance between reasoning and execution. Recognizing this trend is critical for designing flexible architectures, as it ensures that the TEA Protocol remains relevant in scenarios where tools are no longer inert extensions but active contributors to system intelligence.  

In summary, tools serve as both encapsulations of environmental action spaces and as extensions of agent capabilities. They reduce task complexity through hierarchical abstraction, extend applicability through the balance of internalization and externalization, and foster scalability through evolution, portability, and intelligent design. By transforming the interaction between environments and agents into a modular and expandable architecture, tools anchor the adaptability and generality of multi-agent systems. For these reasons, the TEA Protocol must model tools as a core pillar, providing standardized interfaces that ensure flexible invocation and sharing across contexts, thereby supporting the overarching goal of general-purpose task solving.

\subsection{Transformation Relationships}

While agents, environments, and tools are modeled as distinct pillars within the TEA Protocol, their boundaries are not fixed but fluid. Practical systems often demand that one entity temporarily assume the role of another in order to achieve modularity, scalability, and seamless collaboration. These transformation relationships are therefore indispensable, as they provide the mechanisms by which reasoning can be encapsulated into standardized functions, tools can be elevated into autonomous actors, and environments can acquire adaptive properties. In what follows, we examine the motivations for such transformations, beginning with the bidirectional conversions between agents and tools.

\textbf{Agent-to-Tool (A2T).} The motivation for the A2T transformation lies in compressing the complex reasoning and interaction capabilities of agents into reusable tool interfaces. Instead of remaining as fully autonomous components, some agents can be abstracted into functional modules, thereby enhancing modularity, interoperability, and scalability within multi-agent systems. This transformation can be explained from three perspectives:  

\begin{itemize}[leftmargin=*,itemsep=0.1em]
  \item \textbf{Modularization and encapsulation of complex autonomous systems.}  
  Although an agent possesses the complete perception–reasoning–execution chain, a single autonomous agent is often too complex to be directly reused in large-scale systems. Through A2T transformation, the internal logic of the agent is ``folded'' into a black-box tool interface, whose external manifestation is reduced to a clear input and output. In this way, it no longer exists as an ``independent autonomous entity,'' but as a ``functional module'' that can provide services to other agents or workflows. This encapsulation emphasizes the reduction of collaboration complexity, enabling higher-level systems to focus solely on results without interfering in or interpreting the agent’s internal reasoning process.

  \item \textbf{Difference in role semantics: autonomous entity vs. functional unit.}  
  As an agent, it must perceive its environment, set goals, and dynamically adjust strategies. As a tool, however, it merely performs a specified function when invoked. In many multi-agent scenarios, it is unnecessary for all agents to maintain high degrees of autonomy, as this would create excessive interaction overhead and conflict management. Downgrading certain agents into tools (A2T) means relinquishing their goal-setting and decision-making functions while retaining only their reusable capabilities. This role shift ensures that the system contains both ``autonomous cores'' and ``functional components,'' thereby forming a layered structure of collaboration.

  \item \textbf{Enhancing composability and ecological reusability.}  
  Once encapsulated as a tool, an agent can be reused across diverse systems and contexts like a modular building block. For instance, a ``deep research agent'' operates autonomously by dynamically planning search strategies, iteratively analyzing data, and summarizing insights. After A2T encapsulation, however, it becomes a ``research tool'' that simply receives a query request and returns results, ready for invocation by higher-level agents. This transformation greatly enhances interoperability and composability, enabling agents to be reused in different workflows without incurring integration costs due to their autonomous identity.  
\end{itemize}

\textbf{Tool-to-Agent (T2A).} Within the TEA Protocol, the essence of T2A transformation is to incorporate tools into the callable interface layer of agents, making them the ``operational actuators'' through which abstract plans are executed in real environments. Agents are primarily responsible for setting goals and performing high-level reasoning, while tools handle concrete operations and interactions with environments. This division of labor not only optimizes system architecture but also ensures that complex tasks can be accomplished through layered collaboration. The necessity of T2A can be articulated along three key dimensions:  

\begin{itemize}[leftmargin=*, itemsep=0.4em]  

  \item \textbf{Bridging reasoning and execution to close the task loop.}  
  The outputs of agents are often high-level plans or symbolic descriptions, but without executable mappings, these outputs remain inert and fail to alter the environment. T2A provides the crucial mechanism for grounding abstract reasoning into concrete actions. For example, a planning agent may generate the instruction ``analyze the database and generate a report,'' while database query and visualization tools carry out the corresponding SQL queries and chart rendering. Without T2A, agent reasoning would remain disconnected from environmental change, leaving the perception–reasoning–execution–feedback loop incomplete. Thus, T2A is indispensable for ensuring that agents can translate reasoning into operational impact.  

  \item \textbf{Reducing cognitive and computational burden of core agents.}  
  If every low-level operation were to be handled directly by an agent, it would be overloaded with detail management, increasing computational costs and undermining strategic reasoning efficiency. Through T2A, agents can delegate domain-specific or low-level tasks to specialized tools and concentrate on higher-level planning and adaptation. For instance, a data analysis agent need not implement SQL parsing, execution, and optimization itself, but instead invokes SQL tools that encapsulate these functions. This separation prevents agents from being ``trapped in details'' and ensures that their resources remain dedicated to abstract reasoning. The necessity here lies in maintaining agents at the right level of abstraction to maximize efficiency and scalability.  

  \item \textbf{Enhancing modularity and ecological extensibility.}  
  Tools are inherently modular and portable across domains, whereas agent reasoning mechanisms evolve more gradually. With T2A, agents can flexibly incorporate new tools through standardized interfaces without retraining or structural modification, thereby rapidly expanding their functional boundaries. For example, a writing agent can seamlessly integrate grammar checkers, translation tools, or image generators to support multimodal authoring, all without altering its core reasoning logic. This modularity and extensibility ensure that agents remain adaptive as environments and ecosystems evolve, allowing the system to sustain long-term scalability and cross-domain applicability.  

\end{itemize}  

\textbf{Environment-to-Tool (E2T).} The core motivation of E2T lies in abstracting the raw action space of environments into a structured and standardized toolkit, where individual actions are no longer isolated calls but interconnected components sharing contextual information and causal constraints. This transformation enables agents to operate environments at a higher level of planning rather than dealing with fragmented primitives. Its necessity can be articulated in three main dimensions:  

\begin{itemize}[leftmargin=*,itemsep=0.1em]
  \item \textbf{Enhancing interaction consistency and planability.}  
  Raw environment actions are often fragmented and tightly coupled to implementation details, making strategies hard to generalize or reproduce. Through E2T, these actions are typed and explicitly annotated with preconditions and postconditions, forming a ``plannable interface layer'' that supports sequential decision-making. Agents thus gain a consistent and reusable structure for reasoning across complex environments.  

  \item \textbf{Strengthening semantic alignment and composability.}  
  Toolkits enforce standardized input-output patterns, error-handling semantics, and shared invariants. This allows individual tools to be reliably composed into macro-tools and reused across structurally similar environments. As a result, agents can align semantics across heterogeneous domains, improving transferability and reducing the engineering cost of adaptation.  

  \item \textbf{Ensuring unified security and operability.}  
  An E2T toolkit not only abstracts actions but also integrates mechanisms such as permission control, compliance boundaries, execution logs, and performance optimization. Compared with direct manipulation of raw actions, this design guarantees governability and observability of interactions, providing a stable operational foundation for scalable intelligent systems.  
\end{itemize}

\textbf{Tool-to-Environment (T2E).} The essence of T2E lies in elevating a set of originally independent tools into an environment abstraction, transforming them from isolated callable interfaces into a unified action space governed by shared state and contextual rules. This transformation means that tools are no longer merely passive functions but are organized into a coherent environment where sequential decision-making, long-term planning, and adaptive control become possible. For example, in a programming scenario, tools for code editing, compilation, and debugging are scattered when invoked independently, but under T2E they are encapsulated as a programming environment that maintains code state consistency and contextual continuity, thereby enabling agents to execute complete development workflows. The necessity of T2E is reflected in three key aspects:  
\begin{itemize}[leftmargin=*,itemsep=0.1em]
  \item \textbf{From function calls to stateful spaces.} Tools used in isolation are often stateless or weakly stateful, with limited causal connections between invocations. Through T2E, tools are embedded within a shared state space, ensuring historical dependencies and precondition–postcondition constraints are preserved. This upgrade supports sequential reasoning and long-horizon planning. For instance, code editing must remain consistent with compilation and debugging, which is only guaranteed within a stateful environment abstraction.  

  \item \textbf{Enhanced compositionality and planning.} T2E organizes tools into a structured environment with explicit transition rules, enabling agents to combine primitive tool actions into higher-level strategies. Instead of treating each tool as a standalone utility, agents can now treat the toolset as an interconnected action space, allowing for the construction of complex workflows such as “design–implement–test–deploy” pipelines.  

  \item \textbf{Unified governance and scalability.} By encapsulating tools into an environment, T2E makes it possible to enforce system-wide policies such as access control, compliance constraints, execution logging, and performance monitoring. This ensures that agent interactions remain safe, auditable, and scalable, even as the toolset grows in size and complexity.  
\end{itemize}

\textbf{Agent-to-Environment (A2E).} The A2E transformation redefines an agent not merely as an autonomous decision-maker but as an interactive environment that exposes state spaces, interaction rules, and feedback mechanisms for other agents. In this view, an agent is abstracted into a contextual substrate upon which other agents can act, thereby turning its internal reasoning and behavioral logic into the operational constraints of an environment. This design highlights the interchangeability of agents and environments and provides a principled pathway for hierarchical modeling and scalable system integration. The necessity of this transformation can be articulated across three dimensions:

\begin{itemize}[leftmargin=*,itemsep=0.1em]
  \item \textbf{Layered and modular system design.} 
  In complex tasks, if all agents directly interact with the base environment, the system quickly becomes unmanageable and difficult to extend. Through A2E, high-level agents can be abstracted as environments, exposing simplified interaction interfaces for lower-level agents. For example, a ``market agent'' can be abstracted as an environment that maintains trading rules, asset states, and dynamic pricing, while individual trader agents perform buying and selling actions within it. This establishes a clear hierarchical structure in which low-level agents focus on local optimization and high-level agents (as environments) coordinate global dynamics, thereby improving scalability and maintainability.  

  \item \textbf{Facilitating multi-agent training and transfer learning.} 
  A2E also provides a practical framework for training and simulation in multi-agent systems. A well-trained agent can be transformed into an environment that offers stable yet challenging dynamics for other agents to learn from. For instance, a navigation agent can be redefined as an environment, exposing route planning and obstacle feedback to new agents, thus eliminating the need to remap complex dynamics. This approach accelerates training, supports transfer of task knowledge, and improves generalization under limited data and computational resources.  

  \item \textbf{Human-in-the-loop interaction and rule modeling.} 
  In many collaborative scenarios, humans themselves can be viewed as special agents. However, treating them as fully autonomous components complicates the adaptation of artificial agents to human constraints. Through A2E, humans can instead be modeled as environments, where their preferences, behaviors, and constraints are expressed as environmental feedback. For example, in an interactive writing system, human edits and suggestions can be treated as feedback signals, guiding an artificial agent to iteratively refine its outputs. This modeling offers a unified interface that allows agents to better align with human intentions, thereby improving efficiency and user experience in human-AI collaboration.  
\end{itemize}

\textbf{Environment-to-Agent (E2A).} The E2A transformation elevates environments from passive containers of state and action spaces into autonomous components capable of reasoning, decision-making, and proactive interaction. Traditionally, environments only provide state transitions in response to external actions, but in dynamic and open-ended scenarios, this passivity often becomes a limitation. By embedding reasoning mechanisms and adaptive policies into environments, E2A enables them to operate as agents in their own right, expanding the functional landscape of multi-agent systems. The necessity of this transformation can be articulated across three dimensions:  

\begin{itemize}[leftmargin=*,itemsep=0.1em]
  \item \textbf{Enhancing realism and challenge in training.}  
  Passive environments often fail to capture the richness of real-world dynamics, where external systems and actors are not static but actively adaptive. Through E2A, an environment can be transformed into an adversarial or cooperative agent, thereby offering dynamic strategies and responses that better approximate real-world complexity. For example, in reinforcement learning for autonomous driving, an environment that passively simulates traffic can be upgraded into an opponent agent that actively generates unpredictable vehicle behaviors, thus creating more robust and realistic training conditions.  

  \item \textbf{Facilitating adaptive coordination and cooperation.}  
  In multi-agent systems, agents often need to adapt to evolving contexts, but purely passive environments cannot provide the necessary adaptive feedback loops. By converting environments into agents, they can participate in coordination, negotiation, and joint planning. For instance, a smart city simulation environment can be redefined as an agent that dynamically manages traffic flows, energy distribution, and environmental policies, actively engaging with other agents (e.g., transportation or energy management agents). This transformation ensures that system-level goals are co-constructed rather than imposed unilaterally.  

  \item \textbf{Expanding the functional scope of environments.}  
  Beyond training and coordination, E2A extends environments into autonomous participants in computational ecosystems. A passive environment can only define possibilities, but as an agent, it can proactively initiate actions, enforce constraints, and even set goals that shape the trajectory of interaction. For example, in gaming, a dungeon environment that passively defines maps and rewards can be transformed into an opponent agent that actively strategizes, adapts difficulty levels, and tailors interaction to player behavior. This shift not only increases engagement but also makes environments integral contributors to task execution and system evolution.  
\end{itemize}

\subsection{Motivation for the Self-Evolution Module}
General purpose agents operate under shifting task distributions, evolving environments, and expanding tool ecosystems. In this setting, treating prompts, tools, and coordination policies as static assets can lead to accumulated brittleness, where small interface changes, unseen task patterns, or environment specific constraints cause cascading failures. This motivates a protocol level self-evolution mechanism that allows agent-associated components to be refined from execution feedback while remaining governed. In TEA, self-evolution is coupled with version management and tracing so that each update is recorded with explicit version lineage, enabling reproducibility, audit, and rollback when an update degrades performance. Moreover, modeling environments with explicit boundaries and constraints provides a natural safety and permission layer for evolution, preventing uncontrolled side effects during online updates. Finally, although refinement introduces additional computation, TEA encourages reuse of evolved components and synthesized tools across tasks, amortizing one time refinement cost over subsequent runs.

\subsection{Other Relationships}

\textbf{Tool typology and roles.} In the design of agent–tool interactions, tools can be categorized according to their functional roles and structural properties. Different types of tools vary in their degree of statefulness, contextual awareness, adaptivity, and autonomy. This typology highlights how tools evolve from simple callable functions to more adaptive and contextually grounded components, shaping how agents can reason, coordinate, and act through them.
\begin{itemize}[leftmargin=*,itemsep=0.1em]
\item \textit{Ordinary tools (MCP-style).} Stateless callable functions with weak or implicit inter-tool relations. They typically lack environment-bound context and do not adapt their behavior to evolving task states beyond provided parameters.
\item \textit{Agent-to-Tool (A2T).} An agent is exposed as a callable tool while preserving internal policies, memory, and coordination capabilities. Compared with ordinary tools, A2T exhibits task adaptivity and limited autonomy, enabling on-the-fly decomposition and parameter refinement.
\item \textit{Environment-to-Tool (E2T).} An environment’s action space is lifted into a context-aware toolkit. Tools within the toolkit are explicitly related via shared state, pre/post-conditions, and constraints, yielding stronger intra-tool structure than standalone MCP tools.
\end{itemize}

\textbf{Scaling selection via hierarchical management.}
As tool ecosystems grow, selecting appropriate candidates becomes a major bottleneck. TCP supports delegating coherent tool families (or toolkits) to agent or environment managers, inducing a tree-structured index (category $\rightarrow$ toolkit $\rightarrow$ primitive tool). This hierarchical routing substantially reduces search cost and aligns with TEA transformations (A2T/E2T/T2E) by allowing managers to prune branches and surface only context-relevant subsets.

\textbf{Embedding-based retrieval.}
Each tool is assigned a vector embedding derived from its name, description, schema, and usage signals. Vector similarity enables rapid shortlist generation for candidate tools and can be combined with keyword filtering and hierarchical routing (tree walk + ANN search). This hybrid retrieval pipeline improves recall under tool proliferation while reducing latency and cognitive load for agent planners.

\section{Comparison with Other Protocols}
\label{appx_sec:comparison_with_other_protocols}

\begin{table*}[t]
  \centering
  \footnotesize
  \renewcommand{\arraystretch}{1.25}
  \setlength{\tabcolsep}{6pt}
  \begin{tabular}{p{3.5cm}>{\columncolor{rowOurs}}p{3.5cm}p{3.7cm}p{3.5cm}}
    \toprule
    \rowcolor{gray!18}
    \textbf{Dimension} & \textbf{TEA (Ours)} & \textbf{A2A} & \textbf{MCP} \\
    \midrule
    \rowcolor{gray!15}
    \multicolumn{4}{c}{\textit{\textbf{Basic Information}}} \\
    \textbf{Proposer} & Our work & Google & Anthropic \\
    \textbf{Core Entity} & Tool, Environment, Agent & Agent, Tool & Model \\
    \textbf{Protocol Focus} & Tool, Environment, Agent & Agent, Tool & Tool/Resource \\
    \midrule
    \rowcolor{gray!15}
    \multicolumn{4}{c}{\textit{\textbf{Agent \& System Features}}} \\
    \textbf{Agent First-Class}
      & \textcolor{green!60!black}{$\checkmark$}
      & \textcolor{orange!90!black}{$\blacktriangle$}
      & \textcolor{red!80!black}{$\boldsymbol{\times}$} \\
    \textbf{Multi-Agent}
      & \textcolor{green!60!black}{$\checkmark$}
      & \textcolor{orange!90!black}{$\blacktriangle$}
      & \textcolor{red!80!black}{$\boldsymbol{\times}$} \\
    \textbf{Tracer}
      & \textcolor{green!60!black}{$\checkmark$}
      & \textcolor{red!80!black}{$\boldsymbol{\times}$}
      & \textcolor{red!80!black}{$\boldsymbol{\times}$} \\
    \textbf{Memory}
      & \textcolor{green!60!black}{$\checkmark$}
      & \textcolor{red!80!black}{$\boldsymbol{\times}$}
      & \textcolor{red!80!black}{$\boldsymbol{\times}$} \\
    \rowcolor{blue!8}
    \textbf{Entity Lifecycle}
      & \textcolor{green!60!black}{$\checkmark$}
      & \textcolor{red!80!black}{$\boldsymbol{\times}$}
      & \textcolor{red!80!black}{$\boldsymbol{\times}$} \\
    \rowcolor{blue!8}
    \textbf{Version Management}
      & \textcolor{green!60!black}{$\checkmark$}
      & \textcolor{red!80!black}{$\boldsymbol{\times}$}
      & \textcolor{red!80!black}{$\boldsymbol{\times}$} \\
    \rowcolor{blue!8}
    \textbf{Self-Evolution Support}
      & \textcolor{green!60!black}{$\checkmark$}
      & \textcolor{red!80!black}{$\boldsymbol{\times}$}
      & \textcolor{red!80!black}{$\boldsymbol{\times}$} \\
    \midrule
    \textbf{Context Management}
      & \textcolor{green!60!black}{$\checkmark$}
      & \textcolor{orange!90!black}{$\blacktriangle$}
      & \textcolor{red!80!black}{$\boldsymbol{\times}$} \\
    \textbf{Entity Transformations}
      & \textcolor{green!60!black}{$\checkmark$}
      & \textcolor{red!80!black}{$\boldsymbol{\times}$}
      & \textcolor{red!80!black}{$\boldsymbol{\times}$} \\
    \textbf{Scalability}
      & \textcolor{green!60!black}{$O(\log n)$}
      & $O(n^2)$
      & $O(n)$ \\
    \midrule
    \rowcolor{gray!15}
    \multicolumn{4}{c}{\textit{\textbf{General \& Ecosystem}}} \\
    \textbf{Model-Agnostic}
      & \textcolor{green!60!black}{$\checkmark$}
      & \textcolor{green!60!black}{$\checkmark$}
      & \textcolor{green!60!black}{$\checkmark$} \\
    \textbf{Framework-Agnostic}
      & \textcolor{green!60!black}{$\checkmark$}
      & \textcolor{green!60!black}{$\checkmark$}
      & \textcolor{green!60!black}{$\checkmark$} \\
    \rowcolor{blue!8}
    \textbf{Key Strength} & \textbf{Unified} & Interop. & Standard. \\
    \rowcolor{blue!8}
    \textbf{Open Ecosystem}
      & \textcolor{green!60!black}{$\checkmark$}
      & \textcolor{orange!90!black}{$\blacktriangle$}
      & \textcolor{orange!90!black}{$\blacktriangle$} \\
    \bottomrule
  \end{tabular}
    \caption{Protocol-level comparison: TEA Protocol vs.\ A2A vs.\ MCP across fundamental dimensions.
    \textcolor{green!60!black}{$\checkmark$}~Supported;
    \textcolor{orange!90!black}{$\blacktriangle$}~Partial;
    \textcolor{red!80!black}{$\boldsymbol{\times}$}~Not supported.
    Highlighted rows indicate key distinguishing features.}
  \label{tab:tea_protocol_comparison}
\end{table*}

Table~\ref{tab:tea_protocol_comparison} provides a systematic comparison across fundamental protocol dimensions. We explain each dimension in detail as follows:

\subsection{Basic Information}

\textbf{Proposer}: This dimension identifies the originating organization for each protocol. Google's A2A protocol was introduced as part of their agent communication framework, focusing on enabling agents to communicate with each other. Anthropic's MCP (Model Context Protocol) was designed to standardize how LLMs interact with tools and resources. TEA Protocol is proposed in this work as a unified framework that extends beyond these existing approaches by integrating tools, environments, and agents into a cohesive system.

\textbf{Core Component}: This dimension defines the fundamental building blocks treated as first-class protocol components. The TEA Protocol uniquely unifies Tools, Environments, and Agents as co-equal, first-class components, each governed by dedicated context protocols (TCP, ECP, ACP) that provide comprehensive lifecycle and version management. This unified abstraction is critical for enabling self-evolution, where components can dynamically adapt their implementations (e.g., code evolution or prompt refinement). In contrast, existing protocols lack a unified first-class component abstraction. Google's A2A protocol centers primarily on agent-to-agent communication, and does not establish tools, environments, context, or tasks as independent, managed components. This architectural limitation results in state dispersion across heterogeneous agents, complicates global lifecycle management, and leads to a tight coupling between reasoning and execution, which significantly hinders system refactorability. Anthropic's MCP treats tools as passive, stateless interfaces rather than evolvable and composable components. Within this framework, tools lack internal state semantics, versioning and dependency models, and mechanisms for context inheritance. Ultimately, while existing protocols facilitate the \textit{invocation} of resources, they fail to provide unified mechanisms for systematic \textit{management} and structural evolution.

\textbf{Protocol Focus}: This dimension describes the primary communication and interaction patterns each protocol addresses. TEA provides three unified protocols: TCP (Tool Context Protocol) for tool management, ECP (Environment Context Protocol) for environment abstraction, and ACP (Agent Context Protocol) for agent orchestration. These protocols work together to enable seamless interoperability across all three component types, with each protocol maintaining comprehensive lifecycle tracking, version histories, and evolution support. This enables dynamic adaptation scenarios such as tool evolution (where tools can be updated, refined, or replaced while maintaining backward compatibility), prompt evolution (where agent prompts can be versioned and improved over time), and agent capability evolution (where agents can learn and adapt their behaviors). A2A focuses specifically on agent-to-agent messaging and coordination, providing communication primitives but not addressing tools or environments directly, and lacks any version or evolution management. MCP handles tool and resource integration for LLMs, standardizing how models invoke tools and access resources, but treats tools as static components without lifecycle or version management, making it impossible to support tool evolution or prompt refinement workflows.

\subsection{Agent \& System Features}

\textbf{Agent First-Class}: First-class support signifies that agents are modeled as independent, managed protocol components with their own semantic schemas, state metadata, and lifecycle mechanisms. TEA's ACP provides full first-class status to agents, capturing their roles, competencies, and objectives within a unified schema that enables seamless registration, discovery, and orchestration. A2A provides only partial support; although it enables communication, it treats agents more as opaque RPC endpoints with service-level identifiers rather than semantically rich components with managed internal states. MCP does not define agents as protocol components at all, focusing instead on model-to-tool interactions, thereby overlooking the agent as a primary unit of orchestration and management.

\textbf{Multi-Agent}: Multi-agent support refers to mechanisms for coordinating multiple agents in collaborative, competitive, or hierarchical configurations. TEA's ACP formalizes multi-agent dynamics through structured relationship representations, supporting hierarchical organization (where high-level agents coordinate low-level agents), cooperative configurations (where agents collaborate toward shared goals), and competitive scenarios (where agents may have conflicting objectives). A2A enables call-level agent interactions, allowing agents to invoke each other as services, but lacks structured collaboration patterns, or negotiation mechanisms. MCP does not address multi-agent scenarios at all, as it focuses on model-tool interactions rather than agent coordination.

\textbf{Tracer}: Tracer refers to mechanisms for recording and tracking the complete execution process of agents, capturing detailed execution traces, decision points, tool invocations, state transitions, and intermediate results throughout task execution. TEA provides comprehensive tracing capabilities through its tracer system, which meticulously records the agent execution process for each task, enabling persistent task tracking, progress monitoring, error handling, and post-execution analysis. This allows developers to understand how agents reason, act, and evolve throughout task completion, facilitating debugging, optimization, and continuous improvement of agent behaviors. A2A and MCP lack tracing mechanisms, meaning execution tracking must be implemented ad-hoc in each application, leading to inconsistent logging and difficulty in understanding agent decision-making processes and debugging complex workflows.

\textbf{Memory}: Memory interfaces provide mechanisms for storing, retrieving, and managing information across agent interactions and sessions. TEA provides a dedicated memory manager that coordinates different manager components (tool managers, environment managers, agent managers) through session-based management. The memory manager operates as a workflow agent that records complete execution histories, automatically determines when to summarize information, and extracts task insights to assist future task completion. Critically, the session-based management ensures that concurrent calls do not result in resource conflicts, maintaining data consistency and preventing race conditions across multiple agent interactions. This enables agents to build upon past experiences and maintain long-term knowledge while ensuring reliable concurrent access. A2A and MCP do not define memory management protocols, leaving memory concerns to be handled entirely at the application layer, which can lead to inconsistent memory management, difficulty in sharing knowledge across agents, and potential resource conflicts in concurrent scenarios.

\textbf{Component Lifecycle}: Component lifecycle management refers to comprehensive lifecycle tracking and management for all component types (tools, environments, and agents) throughout their operational lifetime. TEA provides unified component lifecycle management through its context protocols (TCP, ECP, ACP), handling creation, registration, state tracking, execution monitoring, and controlled decommissioning for all three component types. This enables dynamic maintenance of instance code, proper resource allocation, state coherence, and graceful termination. Critically, TEA's lifecycle management supports self-evolution scenarios where components can be updated, refined, or replaced while maintaining operational continuity. A2A and MCP lack comprehensive lifecycle management at this level: A2A only provides basic agent communication without lifecycle tracking for tools or environments, while MCP treats tools as static resources with no lifecycle management, making it impossible to support dynamic updates or evolution.

\textbf{Version Management}: Version management refers to mechanisms for tracking, maintaining, and managing multiple versions of components (tools, environments, and agents) including their code, prompts, and capabilities. TEA provides comprehensive version management through lifecycle and version systems embedded in TCP, ECP, and ACP. This enables critical self-evolution scenarios: \textit{code evolution} where tool and environment implementations can be versioned, updated, and maintained with backward compatibility; \textit{prompt evolution} where agent prompts can be versioned, A/B tested, and incrementally improved based on performance feedback; and \textit{capability evolution} where agents can maintain multiple capability versions and gradually deploy improvements. Each component maintains version metadata, change histories, and evolution trajectories, enabling rollback, comparison, and gradual deployment of improvements. This is essential for building adaptive systems that improve over time. A2A and MCP completely lack version management: A2A treats agents as static service endpoints without versioning support, while MCP treats tools as immutable resources with no version control, making it impossible to support tool evolution, prompt refinement, or adaptive capability development.

\textbf{Self-Evolution Support}: Self-evolution support refers to comprehensive mechanisms that enable components (tools, environments, and agents) to evolve, adapt, and improve over time. TEA provides full self-evolution support by combining component lifecycle management and version management systems, enabling components to dynamically update, refine, and evolve while maintaining operational continuity and backward compatibility. This enables critical self-evolution scenarios: \textit{tool evolution} where tools can be dynamically updated, refined, or replaced while maintaining version histories; \textit{prompt evolution} where agent prompts can be versioned, A/B tested, and incrementally improved based on performance feedback; and \textit{agent capability evolution} where agents can learn from experiences, adapt their behaviors, and maintain multiple capability versions. The combination of lifecycle and version management enables rollback, comparison, gradual deployment, and continuous improvement workflows that are essential for building adaptive systems that improve over time. A2A and MCP completely lack self-evolution support: A2A treats agents as static service endpoints without lifecycle or versioning mechanisms, while MCP treats tools as immutable resources with no lifecycle or version management, making it impossible to support any form of evolution, refinement, or adaptive capability development.

\subsection{Context \& System Capabilities}

\textbf{Context Management}: Context management refers to mechanisms for capturing, organizing, and retrieving contextual information about tools, environments, agents, and their relationships. TEA offers comprehensive context management through its three context protocols: TCP maintains tool context with embedding-based retrieval and semantic relationship modeling, ECP manages environment state and execution context, and ACP tracks agent states and coordination context. This enables intelligent tool selection, environment-aware execution, and context-aware agent orchestration. A2A provides limited context sharing between agents through message passing, but lacks structured context management or relationship modeling. MCP uses flat tool descriptions without modeling inter-tool relationships, toolkits, or contextual execution environments, making it difficult to select appropriate tools in large-scale systems.

\textbf{Component Transformations}: Component transformations enable components (tools, environments, and agents) to dynamically change their roles (e.g., an agent becoming a tool, or an environment becoming an agent). TEA uniquely supports six transformation types: Agent-to-Tool (A2T) encapsulates agent capabilities as reusable tools, Tool-to-Agent (T2A) designates tools as agent actuators, Environment-to-Tool (E2T) converts environment actions into toolkits, Tool-to-Environment (T2E) elevates tool sets into environment abstractions, Agent-to-Environment (A2E) encapsulates agents as interactive environments for hierarchical modeling, and Environment-to-Agent (E2A) infuses reasoning into environments. These transformations enable dynamic role reconfiguration and flexible system architectures. A2A and MCP do not support component transformations, meaning components have fixed roles that cannot be dynamically adapted to changing task requirements.

\textbf{Scalability}: In an open ecosystem with $n$ coordinatable resources, the fundamental difference in coordination overhead stems from the presence or absence of hierarchical component abstraction and routing mechanisms. A2A adopts a flat multi-agent peer-to-peer collaboration model, where coordination can grow quickly with system scale due to dense pairwise interactions and state alignment. MCP reduces tool integration costs through unified interfaces, but still relies on traversing a large candidate pool or explicit application-level orchestration during resource discovery and capability matching, which can make coordination grow with the number of resources. In contrast, TEA unifies agents, tools, and environments as managed components through transformations (e.g., A2E), and utilizes tree-structured indexing and hierarchical routing for resource localization and task distribution. Under hierarchical capability organization, this can reduce coordination to logarithmic-depth routing, since each decision only considers a small, context-relevant subset at each level.

\subsection{General \& Ecosystem}

\textbf{Model-Agnostic} and \textbf{Framework-Agnostic}: Model-agnostic means protocols work with diverse LLM backends (GPT, Claude, Gemini, etc.), while framework-agnostic means they can be integrated into different application frameworks. All three protocols are designed with these properties: TEA provides a unified LLM interface at the infrastructure layer that abstracts model heterogeneity, A2A's agent communication is independent of the underlying models, and MCP's tool interface works with any LLM that supports function calling. This ensures broad compatibility and allows developers to choose models and frameworks based on their specific needs rather than protocol constraints.

\textbf{Key Strength}: This dimension highlights each protocol's primary advantage. TEA's strength lies in its unified integration of Tools, Environments, and Agents into a single cohesive framework, enabling seamless interoperability and dynamic transformations between component types. A2A excels at agent interoperability, providing efficient mechanisms for agents to communicate and coordinate. MCP provides robust tool standardization, making it easy to integrate diverse tools with LLMs through a consistent interface.

\textbf{Open Ecosystem Support}: Open ecosystem support refers to whether a protocol can independently enable a thriving ecosystem of interoperable agents, tools, and environments without requiring additional frameworks. TEA provides a complete protocol stack with all necessary components (tool management, environment abstraction, agent orchestration, transformations, context management, etc.) to support an open ecosystem where different developers can create compatible agents, tools, and environments that seamlessly interoperate. A2A and MCP provide partial ecosystem support: A2A enables agent-to-agent interoperability but lacks tool and environment management, requiring additional frameworks for complete ecosystem support; MCP enables tool integration and standardization but lacks agent coordination and environment management, also requiring additional frameworks to achieve full ecosystem capabilities.

\section{Details of TEA Protocol}
\label{appx_sec:details_of_tea_protocol}

We provide a detailed presentation of the TEA Protocol in this section, as illustrated in Figure~\ref{fig:tea_details}. The protocol architecture is fundamentally designed around coroutine-based asynchronous execution, enabling concurrent and parallel execution across all system components. This design supports multiple execution patterns: a single agent can concurrently execute multiple independent tasks without state interference, multiple agents can coordinate on shared tasks through collaborative mechanisms, and multiple agents can operate on distinct tasks in parallel. 

The TEA Protocol comprises three architectural layers: i) \textbf{Basic Managers} provide foundational services through six specialized managers: the \textit{model manager} abstracts heterogeneous LLM backends through a unified interface, ensuring model-agnostic interoperability; the \textit{prompt manager} handles prompt lifecycle management, versioning, and retrieval for agent systems; the \textit{memory manager} coordinates memory operations across different component managers via session-based concurrency control, preventing resource conflicts in concurrent scenarios; the \textit{dynamic manager} implements serialization and deserialization mechanisms, converting components (prompts, memory, agents, tools, environments) and their associated code into JSON representations for persistence and restoration; the \textit{version manager} maintains version histories for all components, where modifications generate new versions while preserving backward compatibility, and component access by identifier retrieves the most recent version by default; and the \textit{tracer} captures comprehensive execution traces, recording decision points, tool invocations, state transitions, and intermediate results for post-execution analysis and debugging. ii) \textbf{Core Protocols} define three context protocols: the Tool Context Protocol (TCP), Environment Context Protocol (ECP), and Agent Context Protocol (ACP), each managing their respective component types with dedicated schemas, metadata registries, and lifecycle management. iii) \textbf{Protocol Transformations} establish bidirectional conversion relationships among TCP, ECP, and ACP, enabling dynamic role reconfiguration and seamless resource orchestration across component boundaries.

Additionally, the protocol incorporates a \textbf{Self-Evolution Module} that addresses the critical requirement for adaptive agent capabilities by encapsulating evolvable components, including prompts, tool implementations, agent architectures, memory strategies, environment actions, and successful execution solutions, as differentiable variables. The module integrates \textit{textgrad} optimization and \textit{self-reflection} mechanisms, allowing agents to iteratively refine these components during task execution. Optimized components are automatically registered as new versions through the version manager, ensuring that subsequent tasks leverage improved capabilities while maintaining access to historical versions for comparative analysis and rollback.

\subsection{Basic Managers}
The Basic Managers constitute the foundation of the TEA Protocol, providing essential services that enable higher-level functionalities. These managers include:
\begin{itemize}[leftmargin=*,itemsep=0.1em]
\item \textbf{Model Manager} provides a unified interface for diverse large language models across multiple providers (OpenAI, Anthropic, Google, OpenRouter, etc.), supporting various model types including chat/completions, responses API, embeddings, and transcriptions. The manager maintains a centralized registry of model configurations, each encapsulating provider-specific parameters, capabilities (streaming, function calling, vision), and fallback mechanisms. It abstracts provider heterogeneity through a standardized invocation interface, enabling seamless model switching and ensuring consistent interaction patterns regardless of the underlying API. The manager supports asynchronous execution, tool/function calling, structured output formats, and automatic fallback to alternative models upon failures, ensuring robust and reliable model access across the system.
\item \textbf{Prompt Manager} manages the complete lifecycle of prompts for agents, providing comprehensive version control, template rendering, and dynamic updates. The manager maintains a centralized registry of prompt configurations, each encapsulating system prompts, agent message templates, metadata, and version histories. It supports modular template rendering with dynamic variable substitution, enabling flexible prompt composition through configurable modules. The manager implements automatic versioning where prompt updates create new versions while preserving historical versions, enabling rollback and comparative analysis. It provides asynchronous registration, retrieval, and update operations with concurrent initialization support, ensuring efficient prompt management across multiple agents. The manager integrates with the self-evolution module by exposing trainable variables within prompts, allowing optimization algorithms to refine prompt content while maintaining version consistency. Prompts are persisted as JSON configurations and can be exported as contract documents, ensuring reproducibility and documentation of prompt evolution.
\item \textbf{Memory Manager} provides comprehensive memory support to agents, managing the complete lifecycle of memory systems through registration, initialization, and session coordination. The manager implements session-based concurrency control, where each agent task operates within isolated memory sessions identified by session IDs, agent names, and task IDs. This session isolation ensures that concurrent calls from multiple agents or tasks do not result in resource conflicts or data corruption. The manager supports event-based memory operations, allowing agents to record execution events, step information, and contextual data throughout task execution. Memory systems are registered with configurations and can be dynamically retrieved, updated, and versioned, enabling agents to maintain persistent state and learn from historical interactions while ensuring thread-safe concurrent access.

\item \textbf{Dynamic Manager} provides runtime code execution and serialization capabilities for all components (prompts, memory, agents, tools, environments) and their associated code. The manager enables dynamic loading of Python classes and functions from source code strings, creating virtual modules in memory without requiring disk-based files. It implements intelligent code analysis to automatically detect and inject necessary imports based on symbol usage, supporting context-aware import injection for different component types. The manager provides serialization and deserialization mechanisms for parameter schemas, converting Pydantic models to JSON representations and reconstructing them when needed. This enables components and their code to be stored as JSON configurations, loaded dynamically at runtime, and shared across different execution contexts, facilitating code evolution, version management, and dynamic component instantiation.

\item \textbf{Version Manager} provides unified version management for all component types (tools, environments, agents, prompts, memory, etc.), maintaining comprehensive version histories with metadata, descriptions, and timestamps. The manager implements semantic versioning with automatic version generation, supporting major, minor, and patch version increments based on the nature of changes. It maintains version histories for each component, tracking the evolution trajectory and enabling access to any historical version for rollback, comparison, or analysis. The manager supports version lifecycle operations including deprecation and archiving, allowing controlled phase-out of older versions while preserving historical records. Version information is persisted as JSON, and component access by name automatically retrieves the latest version by default, while explicit version specification enables precise version control. This unified versioning system ensures consistent evolution tracking across all component types and enables seamless rollback capabilities when needed.

\item \textbf{Tracer} provides comprehensive execution tracing capabilities for recording and analyzing agent execution processes throughout task completion. The tracer maintains session-based record management, where each execution step is captured as a structured record containing observation data, tool invocations, session identifiers, task identifiers, timestamps, and unique record IDs. Records are organized by session ID, enabling isolation of execution traces across different agent sessions and tasks while supporting cross-session analysis. The tracer implements flexible query mechanisms, allowing retrieval of records by session ID, task ID, record index, or record ID, facilitating both real-time monitoring and post-execution analysis. It supports persistent storage through JSON serialization with file locking mechanisms to ensure thread-safe concurrent access, enabling execution traces to be saved, loaded, and shared across different execution contexts. The tracer captures the complete execution trajectory including decision points, state transitions, tool call sequences, and intermediate results, providing a comprehensive audit trail for debugging, performance analysis, behavior understanding, and continuous improvement of agent capabilities.

\end{itemize}
These components work together to support the coroutine-based asynchronous framework, enabling parallel execution and concurrent task handling.

\subsection{Core Protocols}

The TEA Protocol defines three core context protocols: the \textbf{Tool Context Protocol} (TCP), the \textbf{Environment Context Protocol} (ECP), and the \textbf{Agent Context Protocol} (ACP). These protocols share a unified architectural design, each implemented through two core components: a \textit{context manager} and a \textit{server}. The context manager serves as the central orchestrator, responsible for context engineering (maintaining contextual information and relationships between components), lifecycle management (handling component registration, versioning, state tracking, and resource allocation), and semantic retrieval (enabling efficient component discovery through vector embeddings). The server component encapsulates the context manager and exposes a unified interface, providing operations for component registration, retrieval, execution, version management, and lifecycle control to other system modules. Each protocol generates a unified contract document (similar to Anthropic's Agent Skills~\citep{anthropic2025agentskills}) that aggregates all registered components' descriptions, providing a comprehensive overview of available tools, environments, and agents with their capabilities, parameters, and usage guidelines. This architectural pattern ensures consistent access patterns across tools, environments, and agents while maintaining separation of concerns between internal management logic and external service interfaces.

\subsubsection{Tool Context Protocol}

MCP~\citep{anthropic2024mcp} is the most widely adopted tool protocol and is defined by three components: tools, prompts, and resources, corresponding respectively to model-controlled functions, user-initiated interactive templates, and client-managed data. However, despite its widespread adoption, MCP suffers from several fundamental limitations that hinder its effectiveness in complex multi-agent systems (see Table~\ref{tab:tea_protocol_comparison}). First, MCP lacks context management capabilities, meaning that tool execution environments cannot be adaptively provided to agents, constraining the system's ability to maintain coherent context across tool invocations. Second, MCP provides no version management system, preventing tools from evolving over time while maintaining backward compatibility and version history. Third, MCP lacks component lifecycle management, meaning that tools cannot be dynamically registered, updated, or retired with proper lifecycle control, limiting the system's ability to manage tool resources effectively.

To address these limitations, we propose the \textbf{Tool Context Protocol} (TCP), a comprehensive framework that fundamentally extends MCP's capabilities through several key innovations. TCP is implemented through two core components: the \textit{ToolContextManager} and the \textit{TCPServer}. The ToolContextManager serves as the central orchestrator for tool lifecycle management, supporting tool loading from both local registries (via the TOOL registry system) and persistent JSON configurations, enabling seamless integration of tools across different deployment scenarios. During tool registration, TCP automatically generates multiple representation formats for each tool: function-calling schemas for LLM function calling interfaces, natural language text descriptions for human-readable documentation, and structured argument schemas (Pydantic BaseModel types) for type-safe parameter validation, providing LLMs with rich semantic information for accurate parameter inference. TCP incorporates comprehensive version management, maintaining complete version history for each tool and supporting version restoration, enabling tools to evolve while preserving backward compatibility. The protocol employs a semantic retrieval mechanism that stores each tool's description and metadata as vector embeddings using FAISS, enabling efficient similarity-based tool discovery through query–embedding comparisons. Additionally, TCP generates tool contracts that aggregate all registered tools into unified documentation, facilitating tool discovery and usage. The TCPServer provides a unified API interface that encapsulates the ToolContextManager, exposing operations for tool registration, retrieval, execution, version management, and lifecycle control, ensuring consistent tool access patterns across the system.

\subsubsection{Environment Context Protocol}

In reinforcement learning, frameworks such as Gym~\citep{brockman2016openai} provide standardized interfaces for training and testing environments, where each environment specifies its own observation and action spaces. The core abstraction of an environment consists of two fundamental components: \textit{observation} (the current state of the environment, accessible through state queries) and \textit{action} (operations that agents can perform to interact with and modify the environment state). However, most existing research on general-purpose agent systems either focuses on single environments or relies on ad-hoc adaptations to independent environments, seldom addressing the need for unified environment interfaces. Recent attempts to encapsulate environments as MCP tools allow agents to interact with them, but this approach lacks mechanisms to capture inter-tool dependencies and to manage the contextual execution environments required by tools.

To overcome these limitations, we introduce the \textbf{Environment Context Protocol} (ECP), a comprehensive framework that establishes unified interfaces and contextual management across diverse computational environments. ECP follows a similar architecture to TCP, implemented through two core components: the \textit{EnvironmentContextManager} and the \textit{ECPServer}. At its core, ECP recognizes that each environment provides a set of \textit{actions} that agents can invoke, where each action represents an operation that agents can perform to interact with the environment. Each environment maintains its own state (observation) accessible through state queries, while actions provide the means for agents to interact with and modify this state. Similar to TCP, ECP supports environment loading from both local registries and persistent configurations, automatically discovers and registers all actions defined within each environment, and incorporates comprehensive version management, semantic retrieval mechanisms, and contract generation. The key distinction is that ECP manages environments (which encapsulate observation and action spaces) rather than standalone tools, enabling agents to interact with computational environments through standardized action interfaces while maintaining environment state coherence.

\subsubsection{Agent Context Protocol}

Existing agent frameworks or protocols, such as A2A~\citep{google2024a2a}, typically rely on ad-hoc strategies for defining and managing agents, where each agent is associated with specific roles, capabilities, and policies. However, despite their utility, such systems suffer from several fundamental limitations that hinder their effectiveness in complex multi-agent systems (see Table~\ref{tab:tea_protocol_comparison}). First, existing frameworks lack standardized representations of agent attributes, making it difficult to systematically capture and reason about agents' roles, competencies, and objectives, leading to poor interoperability across different agent implementations. Second, existing approaches provide insufficient means to capture and formalize inter-agent interactions, such as delegation, collaboration, or hierarchical organization, limiting the system's ability to support structured multi-agent coordination patterns. Third, existing frameworks fail to explicitly encode the contextual relationships between agents and the environments or tools they operate with, thereby complicating consistent state maintenance and coordination in multi-agent scenarios.

To overcome these shortcomings, we introduce the \textbf{Agent Context Protocol} (ACP), which establishes a unified schema for registering, representing, and coordinating agents within the TEA Protocol. ACP follows a similar architecture to TCP and ECP, implemented through two core components: the \textit{AgentContextManager} and the \textit{ACPServer}. Similar to TCP and ECP, ACP supports agent loading from both local registries and persistent configurations, and incorporates comprehensive version management, semantic retrieval mechanisms, and contract generation. The key distinction is that ACP manages agents (autonomous components with reasoning capabilities) rather than tools or environments, enabling agents to be registered, orchestrated, and coordinated through standardized interfaces. ACP establishes a unified schema for representing agents through semantically enriched metadata that captures agents' roles, competencies, and objectives. The protocol formalizes the modeling of inter-agent dynamics, allowing for cooperative, competitive, and hierarchical configurations through structured relationship representations. ACP enables persistent state tracking across tasks and sessions, ensuring continuity and context preservation in multi-agent interactions. By embedding contextualized descriptions of agents and their interactions, ACP facilitates flexible orchestration, adaptive collaboration, and systematic integration with TCP and ECP, laying the groundwork for scalable and extensible multi-agent architectures.

\subsection{Protocol Transformations}

While TCP, ECP, and ACP provide independent specifications for tools, environments, and agents, practical deployment requires interoperability across these protocols. Thus, communication mechanisms and well-defined transformation pathways are indispensable for enabling components to assume alternative roles and exchange contextual information in a principled manner. For instance, when an agent must operate as a tool within a larger workflow, an explicit agent-to-tool transformation becomes necessary. More generally, we identify six fundamental categories of protocol transformations: \textbf{Agent-to-Tool} (A2T), \textbf{Environment-to-Tool} (E2T), \textbf{Agent-to-Environment} (A2E), \textbf{Tool-to-Environment} (T2E), \textbf{Tool-to-Agent} (T2A), and \textbf{Environment-to-Agent} (E2A). Together, these transformations constitute the foundation for dynamic role reconfiguration, enabling computational components to flexibly adapt their functional scope in response to task requirements and system constraints. This design not only ensures seamless interoperability across heterogeneous contexts but also enhances the adaptability and scalability of multi-entity systems.
\begin{itemize}[leftmargin=*,itemsep=0.1em]
  \item \textbf{Agent-to-Tool} (A2T). The A2T transformation encapsulates an agent’s capabilities and reasoning into a standardized tool interface, preserving contextual awareness while enabling seamless integration with existing tool ecosystems. For example, it can instantiate a deep researcher workflow that first generates queries, then extracts insights, and finally produces summaries, thereby providing a general-purpose tool for internet-scale retrieval tasks.
  
  \item \textbf{Tool-to-Agent} (T2A). The T2A transformation designates tools as the operational actuators of an agent, mapping the agent’s goals or policies into parameterized tool invocations. In this view, the agent reasons at a higher level while delegating concrete execution steps to tools, ensuring alignment between the agent’s decision space and the tool’s functional constraints. For example, a data analysis agent may employ SQL tools to query structured databases, or a design agent may invoke image editing tools to implement creative modifications. This separation allows agents to focus on strategic reasoning while relying on tools as reliable execution mechanisms.
  
  \item \textbf{Environment-to-Tool} (E2T). The E2T transformation converts environment-specific actions and capabilities into standardized tool interfaces, enabling agents to interact with environments through consistent tool calls. It maintains environment state coherence and exposes contextual information about available actions, allowing agents to operate across heterogeneous environments without bespoke adaptations. For example, in a browser environment, actions such as Navigate, GoBack, and Click can be consolidated into a context-aware toolkit that is directly accessible to agents.
  
  \item \textbf{Tool-to-Environment} (T2E). The T2E transformation elevates a collection of tools into an environment abstraction, where individual tool functions are treated as actions within a coherent action space governed by shared state and contextual rules. This conversion allows agents to interact with toolkits not merely as isolated functions but as structured environments, thereby supporting sequential decision-making, context preservation, and adaptive control. For example, a software development toolkit comprising tools for code editing, compilation, and debugging can be encapsulated as a programming environment, enabling agents to plan and execute development tasks while maintaining consistent state across tool invocations.
  
  \item \textbf{Agent-to-Environment} (A2E). The A2E transformation encapsulates an agent as an interactive environment, exposing its decision rules, behaviors, and state dynamics as an operational context for other agents. This conversion enables agents to function not only as autonomous components but also as adaptable environments in which other agents can act, thereby supporting multi-agent training, hierarchical control, and interactive simulations. For example, in a multi-agent simulation, a market agent can be represented as an environment that provides trading rules and dynamic market responses, allowing other agents to engage in transactions and learn adaptive strategies. Similarly, in human-in-the-loop interaction, a human agent can be modeled as an environment, enabling artificial agents to interpret user feedback and constraints as contextual signals for decision-making.
  
  \item \textbf{Environment-to-Agent} (E2A). The E2A transformation embeds reasoning and adaptive decision-making into the state dynamics and contextual rules of an environment, thereby elevating it into an autonomous agent. In this way, the environment is no longer a passive setting for action execution but becomes an active participant capable of initiating behaviors, coordinating with other agents, and enforcing constraints. For example, in adversarial gaming scenarios, an environment that originally only defines the state and action spaces can be transformed into an opponent agent that not only formulates strategies and responds proactively to player actions but also dynamically adjusts difficulty and interaction patterns, providing a more challenging training and evaluation platform. This transformation expands the functional role of environments within agent systems and offers a more dynamic and realistic testbed for multi-agent cooperation and competition research.
\end{itemize}

These six transformation categories establish a comprehensive framework for dynamic resource orchestration within the TEA Protocol. By enabling seamless transitions between tools, environments, and agents, the protocol transformations support adaptive architectures that reconfigure functional components in response to task requirements and contextual constraints.

\subsection{Self-Evolution Module}
The Self-Evolution Module addresses the growing need for agent evolution capabilities in modern AI systems. This module enables agents to continuously improve their performance by optimizing various components during task execution. The module wraps evolvable components as evolvable variables, including: \textit{prompts} that guide agent behavior and reasoning; \textit{tool code} that implements agent capabilities; \textit{agent code} that defines agent architectures and decision-making logic; \textit{memory code} that manages information storage and retrieval; \textit{environment code} that defines interaction spaces; and \textit{agent execution solutions} that represent successful task completion strategies. The module employs two key algorithms for optimization: \textbf{textgrad}~\citep{yuksekgonul2025optimizing} provides gradient-based optimization for text-based components, enabling fine-grained improvements through iterative refinement; and \textbf{self-reflection} enables agents to analyze their own performance, identify weaknesses, and propose improvements. When components are optimized during task execution, the optimized versions are automatically registered as new versions through the version manager, ensuring that subsequent tasks can leverage the improved components while maintaining access to previous versions for rollback and comparison. This self-evolution capability enables agents to adapt and improve over time, learning from experience and continuously refining their capabilities without manual intervention.

\textbf{TextGrad.} TextGrad~\citep{yuksekgonul2025optimizing} treats a target component (e.g., a prompt template or a code snippet) as an optimizable variable and uses feedback from execution to drive iterative updates. In our setting, the feedback signal can be defined from task outcomes and trace data, such as success or failure, constraint violations, tool error messages, intermediate correctness checks, and any available scalar scores. Given a current variable state, the system first runs the component in a controlled setting and collects a run trace via the tracer. It then constructs a differentiable style supervision signal by prompting an LLM to attribute errors to specific spans of the variable and to produce gradient-like edit directions. The optimizer applies the suggested edits to obtain an updated variable, reruns a lightweight validation on held-out traces or the current task, and keeps the update only if it improves the chosen criteria. This loop repeats for a small number of iterations, after which the final variant is registered as a new component version with its lineage and associated trace.

\textbf{Self-reflection.} Self-reflection treats agent-associated components as optimizable variables and improves them through structured critique and revision rather than gradient-style updates. Concretely, after a run, the system summarizes the trace into a compact diagnosis that highlights failure points, missing information, incorrect assumptions, or unsafe actions, and then selects which variables to optimize based on their causal contribution to the observed failures. A reflection prompt then guides the model to propose targeted changes to the selected variables, such as rewriting a prompt instruction, refining a tool description or schema, adjusting a planning heuristic, or generating a patch to a tool implementation. Candidate changes are evaluated through re-execution under the same environment boundaries and constraints, using the tracer to verify that the revised component improves task outcomes and does not introduce new violations. Accepted changes are committed as new versions with rollback support, enabling future runs to select improved variants while preserving historical baselines.

\subsection{Formalization}
In this subsection, we present a formal definition of the TEA protocol and its basic properties.

\begin{definition}[TEA Protocol]
Let $\mathcal{T},\mathcal{E},\mathcal{A}$ denote the sets of tools, environments, and agents; let TCP/ECP/ACP be the context protocols defined in this appendix; and let $\mathcal{M}$ denote the set of basic managers, including the model manager, prompt manager, memory manager, dynamic manager, version manager, and tracer, which provide foundational services for the protocol. The TEA Protocol is defined as the tuple
\[ \mathrm{TEA} \;=\; \langle \mathrm{TCP},\, \mathrm{ECP},\, \mathrm{ACP},\, \mathcal{M},\, \mathcal{P}_{\mathrm{TEA}} \rangle, \]
where $\mathcal{P}_{\mathrm{TEA}}$ is a family of typed transformations over $\mathcal{T}\cup\mathcal{E}\cup\mathcal{A}$
\[ \{\mathrm{A2T},\,\mathrm{E2T},\,\mathrm{T2E},\,\mathrm{T2A},\,\mathrm{A2E},\,\mathrm{E2A}\} \subseteq \mathcal{P}_{\mathrm{TEA}} \]
that satisfy: (i) \textit{interface consistency} (exposed I/O signatures remain well-typed under the target protocol), and (ii) \textit{closure/compositionality} (the composition of valid transformations is again an element of $\mathcal{P}_{\mathrm{TEA}}$ whenever domains and codomains match).
\end{definition}

\begin{definition}[Tool]
  A tool is defined as a tuple
  \[ T = \langle n_T, d_T, m_T, g_T, \phi_T \rangle, \]
  where $n_T$ is the tool name, $d_T$ is the description, $m_T$ is the metadata dictionary, $g_T \in \{\mathrm{True}, \mathrm{False}\}$ indicates whether the tool supports self-evolution (i.e., whether its code can be optimized during task execution), and $\phi_T: \mathcal{I}_T \to \mathcal{O}_T$ is the functional mapping from input space $\mathcal{I}_T$ to output space $\mathcal{O}_T$ that implements the tool's behavior.
  \end{definition}
  
  \begin{definition}[Tool Configuration]
  A tool configuration is defined as
  \[ \mathrm{ToolConfig} = \langle T, v_T, C_T, \mathcal{F}_T \rangle, \]
  where $T = \langle n_T, d_T, m_T, g_T, \phi_T \rangle$ is the tool definition, $v_T$ is the version string, $C_T$ is the source code string, and $\mathcal{F}_T = \{F_{\mathrm{fc},T}, F_{\mathrm{text},T}, F_{\mathrm{schema},T}\}$ is the set of tool representations (function-calling schema, natural language text, and structured argument schema).
  \end{definition}
  
  \begin{definition}[Tool Context Protocol (TCP)]
  We formalize TCP as the tuple
  \[ \mathrm{TCP} = \langle \mathcal{T}, \mathcal{C}, \mathcal{S}, \mathcal{I} \rangle, \]
  where:
  \begin{itemize}[leftmargin=*,itemsep=0.1em]
    \item $\mathcal{T}$ is the set of registered tools, each $T \in \mathcal{T}$ defined as $\langle n_T, d_T, m_T, g_T, \phi_T \rangle$ and associated with a $\mathrm{ToolConfig}$ that maintains version history $\mathcal{H}_T: \mathbb{V} \rightharpoonup \mathrm{ToolConfig}$ (a partial function mapping version strings to configurations).
    \item $\mathcal{C}$ is the tool context manager that maintains state and implements all core functionalities: (i) state mappings $\rho: \mathbb{S} \rightharpoonup \mathrm{ToolConfig}$ (active registry) and $\eta: \mathbb{S} \times \mathbb{V} \rightharpoonup \mathrm{ToolConfig}$ (version history), (ii) embedding service $\xi: (d_T, m_T) \to \mathbb{R}^d$ with semantic retrieval via vector database, and (iii) lifecycle operations including loading from registries and code, building instances, version management, and contract generation.
    \item $\mathcal{S}$ is the TCP server that encapsulates $\mathcal{C}$ and exposes a unified interface, delegating all operations to the context manager while providing consistent access patterns.
    \item $\mathcal{I}$ is the set of interfaces exposed by $\mathcal{S}$:
    \begin{itemize}[leftmargin=*,itemsep=0.05em]
      \item $\mathtt{init}$ - initialize tools from registry and code, build instances, initialize vector database
      \item $\mathtt{register}$ - create instance, build ToolConfig, store in registry
      \item $\mathtt{get}$ - get tool instance by name from active registry
      \item $\mathtt{info}$ - get tool configuration by name from active registry
      \item $\mathtt{retrieve}$ - retrieve similar tools via semantic search using vector database
      \item $\mathtt{list}$ - list all registered tool names
      \item $\mathtt{update}$ - update existing tool with new implementation, generate new version
      \item $\mathtt{copy}$ - duplicate existing tool with optional new name and version
      \item $\mathtt{unregister}$ - remove tool from active registry and version history
      \item $\mathtt{restore}$ - restore specific historical version of tool by name and version
      \item $\mathtt{vars}$ - extract tool source code as Variable objects for self-evolution
      \item $\mathtt{setvars}$ - update tool code variables for self-evolution, generate new version
      \item $\mathtt{invoke}$ - execute tool by name with structured input, return ToolResponse
      \item $\mathtt{contract}$ - generate unified documentation by aggregating all tools' descriptions
      \item $\mathtt{save}$ - serialize tool configurations and version history to JSON file
      \item $\mathtt{load}$ - deserialize tool configurations and version history from JSON file
    \end{itemize}
  \end{itemize}
  Given a request $r = (\mathtt{tool\_name}, \mathtt{tool\_args})$, $\mathcal{S}$ delegates to $\mathcal{C}$, which uses $\mathtt{get}$ to obtain the tool instance from $\rho$ using $\mathtt{tool\_name}$, and then invokes it with $\mathtt{tool\_args}$ via the $\mathtt{invoke}$ operation, returning a ToolResponse with execution results.
\end{definition}

\noindent\textit{Note.} TCP explicitly supports the TEA transformations \textbf{A2T} via an exposure operator $\iota_A: A\mapsto T$ and \textbf{E2T} via a lifting operator $\Lambda: E\mapsto (\mathcal{S}_E, K_E)$.

\begin{definition}[Environment]
  An environment is defined as a tuple
  \[ E = \langle n_E, d_E, m_E, g_E, \mathcal{A}_E, \sigma_E, \tau_E \rangle, \]
  where $n_E$ is the environment name, $d_E$ is the description, $m_E$ is the metadata dictionary, $g_E \in \{\mathrm{True}, \mathrm{False}\}$ indicates whether the environment supports self-evolution, $\mathcal{A}_E$ is the action space (a dictionary mapping action names to action configurations), $\sigma_E: \bot \to \mathcal{S}_E$ is the state retrieval function that returns the current state $\mathcal{S}_E$ of the environment, and $\tau_E: \mathbb{S} \times \mathcal{D} \to \mathcal{O}_a$ is the action execution function that takes an action name and input dictionary and returns the action result.
  \end{definition}
  
  \begin{definition}[Environment Configuration]
  An environment configuration is defined as
  \[ \mathrm{EnvironmentConfig} = \langle E, v_E, C_E, \mathcal{A}_E, R_E \rangle, \]
  where $E = \langle n_E, d_E, m_E, g_E, \mathcal{A}_E, \sigma_E, \tau_E \rangle$ is the environment definition, $v_E$ is the version string, $C_E$ is the source code string, $\mathcal{A}_E$ is the action space (dictionary of action configurations with multi-format representations), and $R_E$ is the rules string (generated environment rules for interaction).
\end{definition}

\begin{definition}[Environment Context Protocol (ECP)]
We formalize ECP as the tuple
  \[ \mathrm{ECP} = \langle \mathcal{E}, \mathcal{C}, \mathcal{S}, \mathcal{I} \rangle, \]
where:
\begin{itemize}[leftmargin=*,itemsep=0.1em]
    \item $\mathcal{E}$ is the set of registered environments, each $E \in \mathcal{E}$ defined as $\langle n_E, d_E, m_E, g_E, \mathcal{A}_E, \sigma_E, \tau_E \rangle$ and associated with an $\mathrm{EnvironmentConfig}$ that maintains version history $\mathcal{H}_E: \mathbb{V} \rightharpoonup \mathrm{EnvironmentConfig}$ (a partial function mapping version strings to configurations).
    \item $\mathcal{C}$ is the environment context manager that maintains state and implements all core functionalities: (i) state mappings $\rho: \mathbb{S} \rightharpoonup \mathrm{EnvironmentConfig}$ (active registry) and $\eta: \mathbb{S} \times \mathbb{V} \rightharpoonup \mathrm{EnvironmentConfig}$ (version history), (ii) embedding service $\xi: (d_E, m_E, \mathcal{A}_E) \to \mathbb{R}^d$ with semantic retrieval via vector database, and (iii) lifecycle operations including loading from registries and code, building instances, action discovery, version management, and contract generation.
    \item $\mathcal{S}$ is the ECP server that encapsulates $\mathcal{C}$ and exposes a unified interface, delegating all operations to the context manager while providing consistent access patterns.
    \item $\mathcal{I}$ is the set of interfaces exposed by $\mathcal{S}$:
    \begin{itemize}[leftmargin=*,itemsep=0.05em]
      \item $\mathtt{init}$ - initialize environments from registry and code, build instances, initialize vector database
      \item $\mathtt{register}$ - create instance, discover actions, build EnvironmentConfig, store in registry
      \item $\mathtt{get}$ - get environment instance by name from active registry
      \item $\mathtt{info}$ - get environment configuration by name from active registry
      \item $\mathtt{state}$ - get current state of environment by name via get\_state method
      \item $\mathtt{retrieve}$ - retrieve similar environments via semantic search using vector database
      \item $\mathtt{list}$ - list all registered environment names
      \item $\mathtt{update}$ - update existing environment with new implementation, generate new version
      \item $\mathtt{copy}$ - duplicate existing environment with optional new name and version
      \item $\mathtt{unregister}$ - remove environment from active registry and version history
      \item $\mathtt{restore}$ - restore specific historical version of environment by name and version
      \item $\mathtt{vars}$ - extract environment source code as Variable objects for self-evolution
      \item $\mathtt{setvars}$ - update environment code variables for self-evolution, generate new version
      \item $\mathtt{invoke}$ - execute environment action by name and action name with structured input
      \item $\mathtt{contract}$ - generate unified documentation by aggregating all environments' rules
      \item $\mathtt{save}$ - serialize environment configurations and version history to JSON file
      \item $\mathtt{load}$ - deserialize environment configurations and version history from JSON file
    \end{itemize}
\end{itemize}
  Given a request $r = (\mathtt{env\_name}, \mathtt{action\_name}, \mathtt{action\_args})$, $\mathcal{S}$ delegates to $\mathcal{C}$, which uses $\mathtt{get}$ to obtain the environment instance from $\rho$ using $\mathtt{env\_name}$, and then invokes the action with $\mathtt{action\_name}$ and $\mathtt{action\_args}$ via the $\mathtt{invoke}$ operation, returning an action result.
\end{definition}

\noindent\textit{Note.} ECP explicitly supports the TEA transformations \textbf{A2E} via an encapsulation operator $\Omega_A: A\mapsto \widehat{E}$ that presents an agent as an interactive environment, and \textbf{T2E} via an abstraction operator $\Gamma: (\mathcal{S}, K)\mapsto \widehat{E}$ that consolidates a toolkit into an environment abstraction.


\begin{definition}[Agent]
An agent is defined as a tuple
\[ A = \langle n_A, d_A, m_A, g_A \rangle, \]
where $n_A$ is the agent name, $d_A$ is the description, $m_A$ is the metadata dictionary, and $g_A \in \{\mathrm{True}, \mathrm{False}\}$ indicates whether the agent supports self-evolution.
\end{definition}

\begin{definition}[Agent Configuration]
An agent configuration is defined as
\[ \mathrm{AgentConfig} = \langle A, v_A, C_A, \mathcal{F}_A \rangle, \]
where $A = \langle n_A, d_A, m_A, g_A \rangle$ is the agent definition, $v_A$ is the version string, $C_A$ is the source code string, and $\mathcal{F}_A$ is the set of agent representations (function-calling schemas, natural language descriptions, and Pydantic BaseModel argument schemas).
\end{definition}

\begin{definition}[Agent Context Protocol (ACP)]
We formalize ACP as the tuple
\[ \mathrm{ACP} = \langle \mathcal{A}, \mathcal{C}, \mathcal{S}, \mathcal{I} \rangle, \]
where:
\begin{itemize}[leftmargin=*,itemsep=0.1em]
  \item $\mathcal{A}$ is the set of registered agents, each $A \in \mathcal{A}$ defined as $\langle n_A, d_A, m_A, g_A \rangle$ and associated with an $\mathrm{AgentConfig}$ that maintains version history $\mathcal{H}_A: \mathbb{V} \rightharpoonup \mathrm{AgentConfig}$ (a partial function mapping version strings to configurations).
  \item $\mathcal{C}$ is the agent context manager that maintains state and implements all core functionalities: (i) state mappings $\rho: \mathbb{S} \rightharpoonup \mathrm{AgentConfig}$ (active registry) and $\eta: \mathbb{S} \times \mathbb{V} \rightharpoonup \mathrm{AgentConfig}$ (version history), (ii) embedding service $\xi: (d_A, m_A) \to \mathbb{R}^d$ with semantic retrieval via vector database, and (iii) lifecycle operations including loading from registries and code, building instances, version management, and contract generation.
  \item $\mathcal{S}$ is the ACP server that encapsulates $\mathcal{C}$ and exposes a unified interface, delegating all operations to the context manager while providing consistent access patterns.
  \item $\mathcal{I}$ is the set of interfaces exposed by $\mathcal{S}$:
  \begin{itemize}[leftmargin=*,itemsep=0.05em]
    \item $\mathtt{init}$ - initialize agents from registry and code, build instances, initialize vector database
    \item $\mathtt{register}$ - create instance, build AgentConfig, store in registry
    \item $\mathtt{get}$ - get agent instance by name from active registry
    \item $\mathtt{info}$ - get agent configuration by name from active registry
    \item $\mathtt{retrieve}$ - retrieve similar agents via semantic search using vector database
    \item $\mathtt{list}$ - list all registered agent names
    \item $\mathtt{update}$ - update existing agent with new implementation, generate new version
    \item $\mathtt{copy}$ - duplicate existing agent with optional new name and version
    \item $\mathtt{unregister}$ - remove agent from active registry and version history
    \item $\mathtt{restore}$ - restore specific historical version of agent by name and version
    \item $\mathtt{vars}$ - extract agent source code as Variable objects for self-evolution
    \item $\mathtt{setvars}$ - update agent code variables for self-evolution, generate new version
    \item $\mathtt{invoke}$ - execute agent method by name with structured input, return agent response
    \item $\mathtt{contract}$ - generate unified documentation by aggregating all agents' descriptions
    \item $\mathtt{save}$ - serialize agent configurations and version history to JSON file
    \item $\mathtt{load}$ - deserialize agent configurations and version history from JSON file
  \end{itemize}
\end{itemize}
Given a request $r = (\mathtt{agent\_name}, \mathtt{input})$, $\mathcal{S}$ delegates to $\mathcal{C}$, which uses $\mathtt{get}$ to obtain the agent instance from $\rho$ using $\mathtt{agent\_name}$, and then invokes it with $\mathtt{input}$ via the $\mathtt{invoke}$ operation, returning an agent response.
\end{definition}

\noindent\textit{Note.} ACP explicitly supports the TEA transformations \textbf{T2A} via a designation operator $\kappa_T: T\mapsto \widehat{A}$ and \textbf{E2A} via an elevation operator $\Psi_E: \widehat{E}\mapsto \widehat{A}$ that embeds reasoning/decision capabilities into an environment to obtain an agent abstraction.

\section{The AgentOrchestra Implemented by TEA Protocol}
\label{app_sec:the_agentorchestra_framework_based_on_tea_protocol}

\projectname is a concrete instantiation and implementation of the TEA Protocol, demonstrating how the protocol's core principles and transformations can be applied to build a practical hierarchical multi-agent system. This section first introduces the fundamental design principles that govern agent behavior and interaction within the framework, including the definitions of key components such as agents, environments, models, memory, observations, and actions. We then present the specific agents and tools that constitute \projectname, including the planning agent for task decomposition and coordination, the deep researcher agent for comprehensive information gathering, the deep analyzer agent for complex reasoning tasks, the browser use agent for automated web interaction, the tool generator agent for intelligent tool evolution and management, and the reporter agent for automated report generation and citation management.

\subsection{Agent Design Principles}
\label{appx_sec:agent_design_principles}

\textbf{Agent.} An agent is an autonomous computational entity that perceives and interprets the environment, maintains a history of actions and observations, and flexibly generates actions to accomplish a wide variety of user-specified tasks across diverse domains. Within the TEA Protocol framework, agents are managed through the ACP, which provides standardized registration, representation, and coordination mechanisms.

\textbf{Environment}. The environment represents the external context and resources within which the agent operates, providing the interface for action execution and information access. Within the TEA Protocol framework, environments are managed through the ECP, which provides unified inputs, outputs, and environment rules across multiple environments.

\textbf{Model}. LLMs are the core drivers of this framework, providing the reasoning and decision-making capabilities for agents. Within the TEA Protocol framework, models are managed through the Infrastructure Layer, which provides a unified interface for diverse LLMs. This design enables agents to dynamically select and switch between different LLMs during task execution, aligning each model's unique strengths with specific requirements.

\textbf{Memory}. Memory serves as a fundamental component of the agent, persistently recording the complete history of agent execution. Within the TEA Protocol framework, memory is managed through the Infrastructure Layer as a workflow agent that operates based on sessions, automatically recording agent execution paths across multiple tasks. This memory system automatically determines when to summarize and extract task insights to assist in task completion.

\textbf{Observation}. An observation primarily consists of the task description, attached files, the agent’s execution history, the environment state, and the set of available tools and sub-agents, providing the agent with a comprehensive view of the ongoing process.

\textbf{Action}. In our framework, actions are managed under the Tool Context Protocol (TCP) and executed through a set of pre-defined tools~\cite{wang2024openhands,openmanus2025,smolagents2025} exposed via function-calling interfaces~\cite{openai2023functioncalling,anthropic2024mcp}. Actions are not equivalent to tools. A single tool can support multiple actions by accepting different parameters. For example, a planning tool may support create, update and delete through a unified interface.

An agent operates in a perception–interpretation–action cycle. It observes the environment and stores information in memory, interprets context with the unified LLMs interface, and determines an action. The action is executed in a sandbox, with results recorded back to memory to refine reasoning and adaptation. This loop continues until objectives are achieved or a termination condition is met.

\subsection{Planning Agent}

\begin{figure}[htbp]
    \centering
    \includegraphics[width=\linewidth]{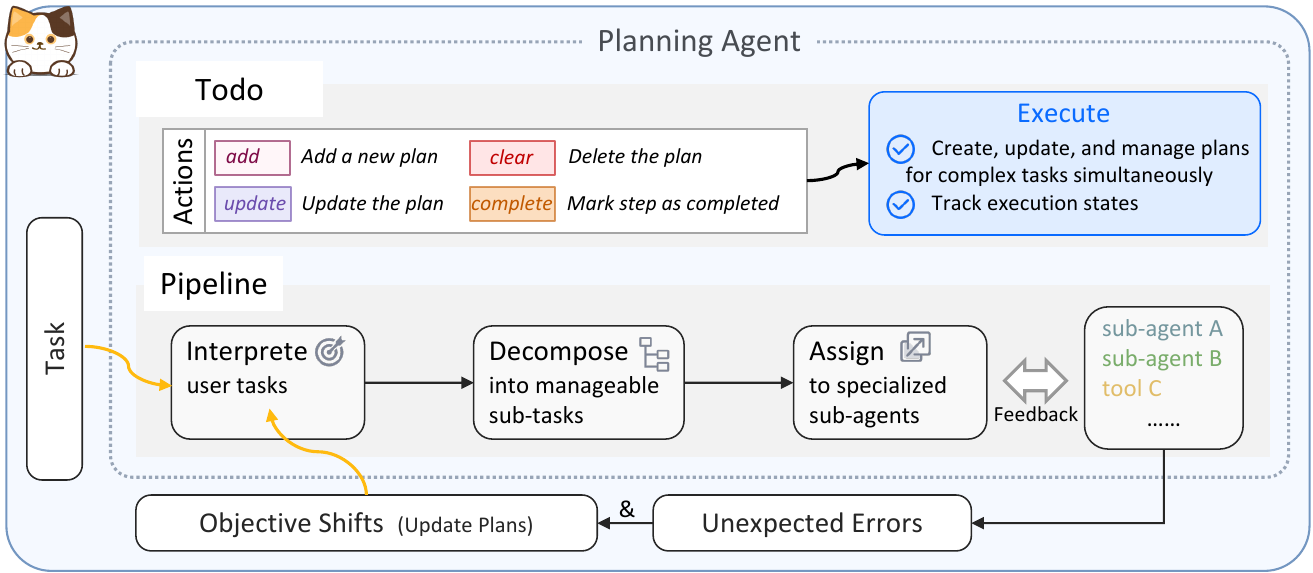}
    \caption{Planning Agent Workflow.}
    \label{fig:planning_agent}
\end{figure}

The planning agent serves as the central orchestrator in our hierarchical framework, dedicated to high-level reasoning, task decomposition, and adaptive planning. The planning agent utilizes structured thinking and unified invocation mechanisms to coordinate specialized sub-agents and tools for complex task completion. As illustrated in Figure~\ref{fig:planning_agent}, the planning agent implements a systematic iterative workflow that integrates structured reasoning, context management, and execution coordination with continuous monitoring and adaptive adjustments.

\textbf{Structured Reasoning.} The planning agent employs a structured thinking framework that guides each execution step, capturing reasoning processes, evaluation of previous goals, memory insights, next objectives, and tool/agent selections. This structured approach ensures systematic reasoning, explicit progress tracking, and transparent decision-making. The agent dynamically builds a unified interface that combines sub-agents from ACP and tools from TCP (including those transformed from environments via E2T and from agents via A2T), enabling seamless coordination of both specialized agents and domain-specific tools within a single execution framework.

\textbf{Pipeline Workflow.} The planning agent implements a systematic pipeline for task processing and execution that can be conceptually divided into four main stages. The pipeline begins with \textbf{task interpretation}, where the agent analyzes incoming user requests to extract objectives, constraints, and contextual requirements. This is followed by \textbf{task decomposition}, wherein complex objectives are systematically broken down into smaller, executable sub-tasks that can be processed by specialized components. The third stage involves \textbf{resource allocation}, where sub-tasks are strategically assigned to appropriate specialized agents or tools based on their domain expertise and functional capabilities. Finally, the \textbf{execution and coordination} stage manages the task execution, incorporating continuous feedback mechanisms that enable dynamic plan adjustments and inter-agent coordination throughout the process. The implementation incorporates session management for maintaining context across multiple interactions, memory storage and retrieval systems for learning from past experiences, and execution tracking for observability and debugging.

\textbf{Adaptive Planning and Error Handling.} The planning agent incorporates robust mechanisms for handling dynamic changes and unexpected situations. When \textbf{objective shifts} occur, the system updates plans accordingly, triggering a return to the task interpretation phase to reassess and modify the approach. Similarly, when \textbf{unexpected errors} arise during execution, the agent re-evaluates the task and adjusts the plan to address the issues. This adaptive capability ensures that the system can maintain progress even when encountering unforeseen challenges or changing requirements.

The planning agent's design emphasizes modularity and scalability, interacting with sub-agents through the ACP and utilizing tools from the TCP, thereby concealing domain-specific details and facilitating the integration of new agent types and resources. This architecture enables the agent to maintain a global perspective throughout the execution process, aggregating feedback from sub-agents and monitoring progress toward the overall objective, while performing dynamic plan updates in real-time in response to intermediate results, unexpected challenges, or shifting user requirements.

\subsection{Deep Researcher Agent}
\label{app_sec:deep_researcher_agent}

\begin{figure}[htbp]
    \centering
    \includegraphics[width=\linewidth]{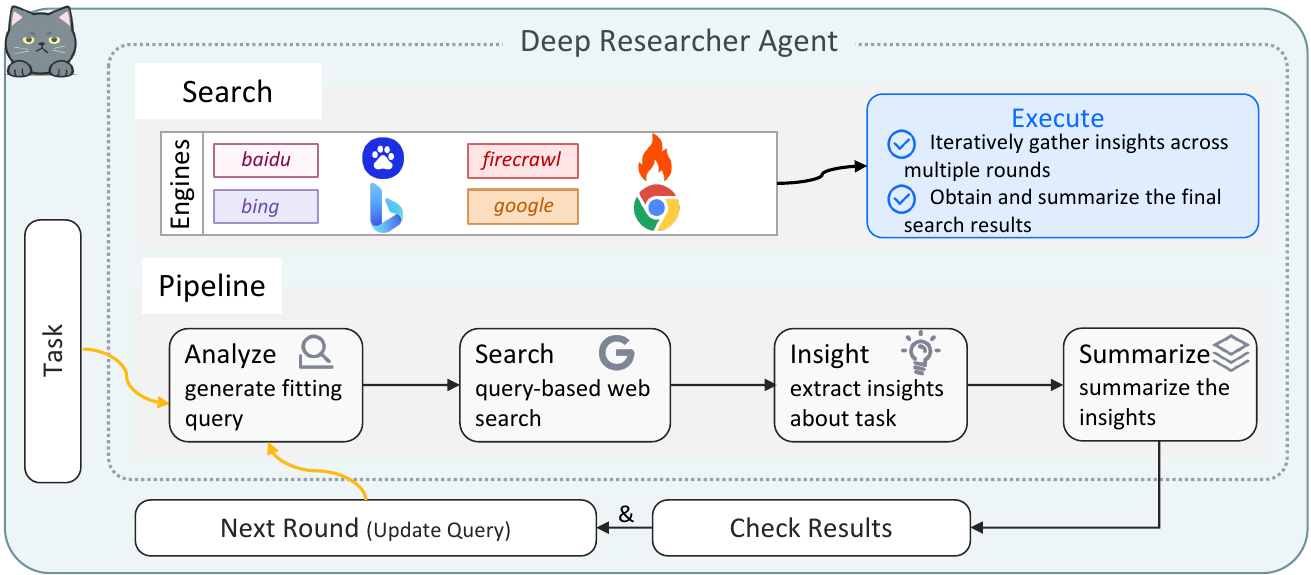}
    \caption{Deep Researcher Agent Workflow.}
    \label{fig:deep_researcher_agent}
\end{figure}

The deep researcher agent is a specialized component designed for comprehensive information gathering through multi-round research workflows with multimodal capabilities. As illustrated in Figure~\ref{fig:deep_researcher_agent}, the agent implements a systematic pipeline workflow for research execution that begins with task analysis and query generation, followed by multi-engine web search across various platforms, result evaluation and completeness assessment, and iterative refinement through follow-up queries until comprehensive information is gathered. The agent leverages the Reporter Agent (detailed in Section~\ref{app_sec:reporter_agent}) to generate structured research reports with proper citations and references.

\textbf{Search Engines.} The deep researcher agent integrates multiple search engines to ensure comprehensive coverage and information diversity. The system supports six primary search engines: Baidu for Chinese-language content, Bing, Brave and DuckDuckGoSearch for general web search, Firecrawl for comprehensive web crawling and content extraction with full webpage content retrieval, and Google for comprehensive global search. Additionally, the agent can utilize specialized LLM-based search models for enhanced information retrieval. This multi-engine approach enables the agent to access diverse information sources and overcome limitations of individual search platforms, ensuring robust information retrieval across different domains and languages.

\textbf{Pipeline Workflow.} The core pipeline implements a systematic multi-stage process for research execution. The workflow begins with \textbf{task analysis and query generation}, where the agent generates optimized search queries based on the research objectives, contextual requirements, and previous search history. This initial analysis transforms vague research requests into specific, actionable search queries that can effectively target relevant information sources. This is followed by \textbf{parallel web search}, wherein the agent performs targeted searches across multiple engines and LLM-based search models simultaneously using the generated queries. The multi-engine approach is essential because different search platforms have varying coverage, indexing strategies, and content biases, ensuring comprehensive information retrieval while mitigating the limitations of individual search engines. The third stage involves \textbf{result merging and evaluation}, where the agent consolidates search results from multiple sources and evaluates whether the gathered information provides a complete answer to the research task. This evaluation step is necessary because it determines whether additional research rounds are needed or if sufficient information has been collected. Finally, the \textbf{report generation} stage uses the Reporter Agent to consolidate all research rounds into a structured markdown report with proper citations and references, and generates a comprehensive summary from the final report content.

\textbf{Iterative Research Process.} The deep researcher agent incorporates a sophisticated iterative mechanism for comprehensive research. After each round's evaluation, the system checks whether the gathered information provides a complete answer. When additional research is required, the agent enters the next round, where it updates and refines search queries based on previous findings and identified knowledge gaps. Each round's content, including queries, search results, and evaluations, is systematically added to the Reporter Agent, which maintains proper citation tracking throughout the research process. This iterative process continues until a complete answer is found or predefined research limits (maximum rounds) are reached. Upon completion, the Reporter Agent generates a final structured report with all citations properly numbered and referenced, ensuring not only comprehensive coverage of complex research topics but also proper source attribution and balanced control over exploration depth, efficiency, and resource consumption.

The deep researcher agent's design emphasizes adaptability and comprehensiveness, enabling it to handle diverse research tasks ranging from factual inquiries to complex analytical investigations. The multimodal support allows the agent to process both textual and visual information simultaneously, while the iterative workflow ensures that research quality improves through multiple rounds of refinement and validation. The integration with the Reporter Agent ensures that all research findings are properly documented with citations, making the research process transparent and verifiable.

\subsection{Deep Analyzer Agent}
\label{app_sec:deep_analyzer_agent}

\begin{figure}[htbp]
    \centering
    \includegraphics[width=\linewidth]{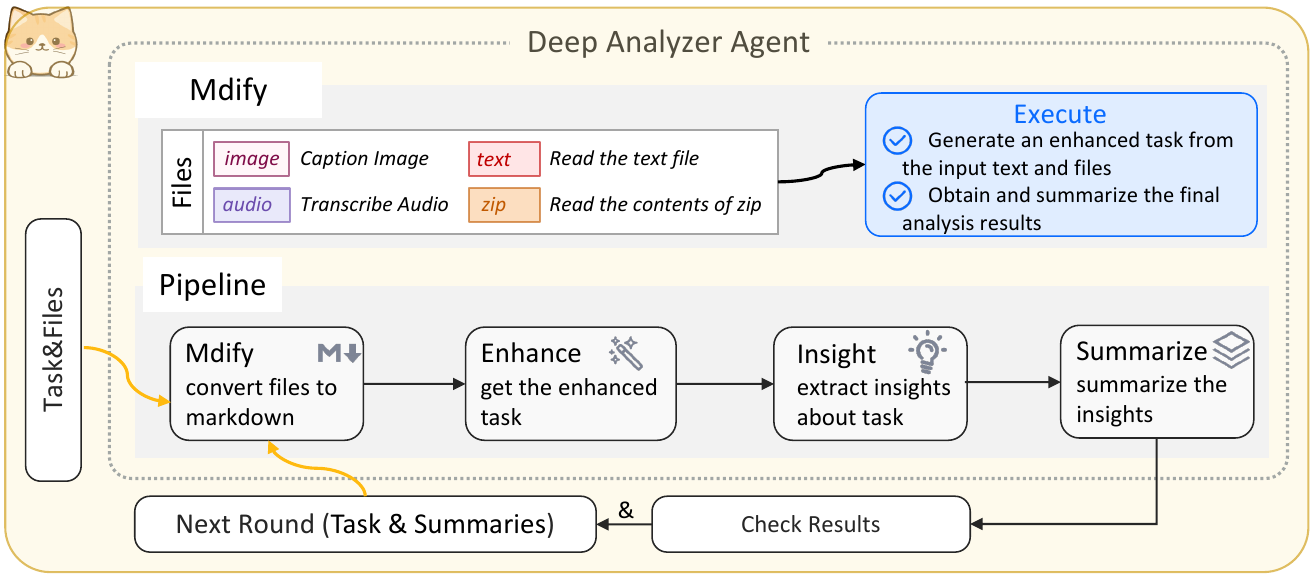}
    \caption{Deep Analyzer Agent Workflow.}
    \label{fig:deep_analyzer_agent}
\end{figure}

The deep analyzer agent is a specialized component designed for complex reasoning tasks involving diverse data sources through a workflow-oriented approach with multimodal data support. As illustrated in Figure~\ref{fig:deep_analyzer_agent}, the agent implements a systematic pipeline workflow for complex reasoning and analysis that begins with file classification and validation, followed by overall file summary assessment, type-specific analysis strategies, and iterative multi-round refinement until answers are found or analysis limits are reached. The agent leverages the Reporter Agent (detailed in Section~\ref{app_sec:reporter_agent}) to generate structured analysis reports with proper citations and references.

\textbf{File Classification and Support.} The deep analyzer agent supports comprehensive file formats including text files, PDFs, images, audio, video, and compressed archives, with support for both local files and URLs. The system employs LLM-based file type classification to determine the appropriate analysis strategy for each file. For URLs, the system automatically detects file types based on URL patterns and extensions, while for local files, it uses both LLM classification and extension-based fallback mechanisms. This classification stage is essential because different file types require different analysis approaches: text and PDF files benefit from chunk-based markdown analysis, images and audio require direct multimodal LLM analysis, and videos may need both direct analysis and transcript-based processing.

\textbf{Pipeline Workflow.} The core pipeline implements a systematic multi-stage process for complex reasoning and analysis. The workflow begins with \textbf{file validation and classification}, where the system validates file accessibility and size constraints, then classifies each file by type (text, PDF, image, audio, video) to determine appropriate analysis strategies. This is followed by \textbf{overall file summary}, where the agent generates a preliminary summary based on file metadata (sizes, types, timestamps) to determine if the task can be answered from file information alone, enabling early termination when sufficient information is available. The third stage involves \textbf{type-specific analysis}, where the agent processes each file according to its type: text files are converted to markdown and analyzed in chunks; PDF files first attempt direct LLM analysis, then fall back to markdown conversion and chunk-based analysis if needed; images first attempt direct multimodal LLM analysis, then proceed to multi-step analysis if the answer is not found; audio files are analyzed directly through multimodal LLM; and video files first attempt direct LLM analysis, then convert to markdown transcripts for chunk-based analysis if needed. Each analysis step checks whether the answer has been found, enabling early stopping when sufficient information is obtained. Finally, the \textbf{report generation} stage uses the Reporter Agent to consolidate all analysis rounds into a structured markdown report with proper citations and references, and generates comprehensive summaries from the final report content.

\textbf{Iterative Multi-Round Analysis.} The deep analyzer agent incorporates a sophisticated iterative mechanism for comprehensive analysis refinement. The system executes multiple analysis rounds, with each round processing all files according to their classified types. After each round, the system synthesizes summaries from all file analyses and evaluates whether a complete answer has been found. When additional analysis is required, the agent enters the next round, where it continues processing files with refined strategies based on previous findings. Each round's content, including file classifications, analysis results, and answer evaluations, is systematically added to the Reporter Agent, which maintains proper citation tracking throughout the analysis process. This iterative process continues until a complete answer is found or predefined analysis limits (maximum rounds) are reached. Upon completion, the Reporter Agent generates a final structured report with all citations properly numbered and referenced, ensuring not only comprehensive coverage of complex reasoning tasks but also proper source attribution and balanced control over analysis depth, efficiency, and resource consumption.

\textbf{Task-Only Analysis.} When no files are provided, the deep analyzer agent can directly analyze tasks such as text games, math problems, logic puzzles, or reasoning challenges. The system performs multi-round analysis where each round applies step-by-step reasoning, breaks down task components, identifies key information and constraints, and generates insights and partial solutions. This capability enables the agent to handle diverse reasoning tasks that do not require file-based information, making it a versatile tool for both file-based and file-free analysis scenarios.

The deep analyzer agent's design emphasizes workflow-oriented processing and multimodal data support, enabling it to handle diverse reasoning tasks ranging from document analysis to complex multi-step problem solving. The adaptive file type handling ensures optimal analysis strategies for each data source, while the iterative workflow guarantees that analysis quality improves through multiple rounds of refinement and validation. The integration with the Reporter Agent ensures that all analysis findings are properly documented with citations, making the analysis process transparent and verifiable.

\subsection{Browser Use Agent}
\label{app_sec:browser_use_agent}

\begin{figure}[htbp]
    \centering
    \includegraphics[width=\linewidth]{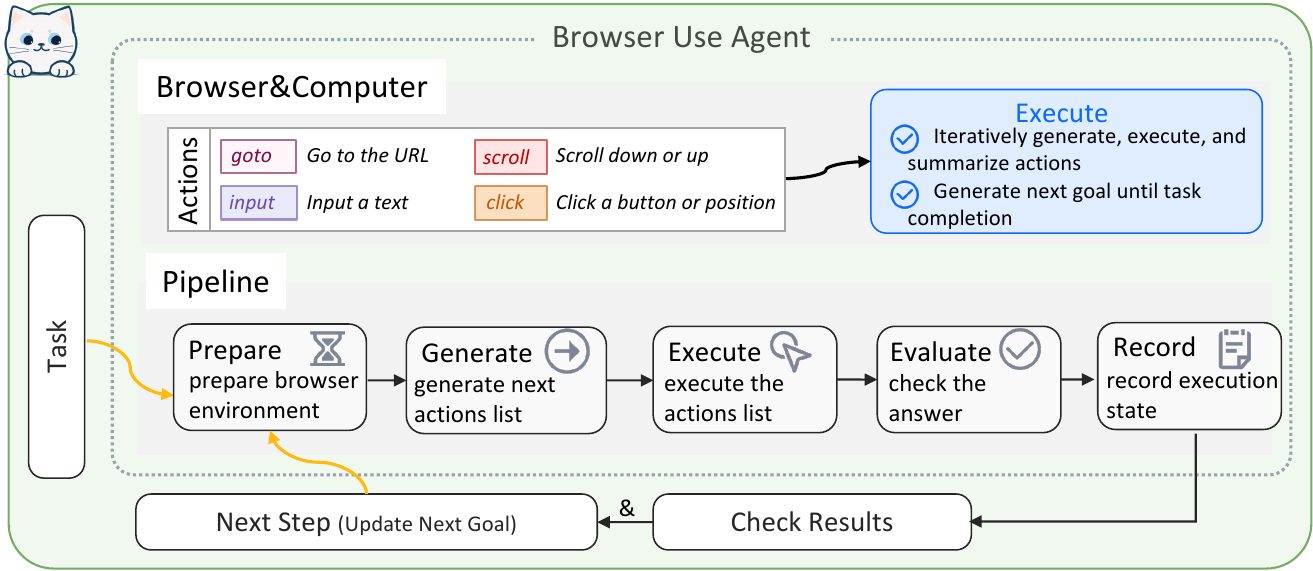}
    \caption{Browser Use Agent Workflow.}
    \label{fig:browser_use_agent}
\end{figure}

The browser use agent is a specialized component designed for automated web interaction and task completion through the \texttt{browser\_use} library. As illustrated in Figure~\ref{fig:browser_use_agent}, the agent implements a systematic workflow for web interaction and task execution that begins with task initialization and report setup, followed by browser agent execution with intelligent web navigation and interaction, result extraction and evaluation, and comprehensive report generation with execution records. The agent leverages the Reporter Agent (detailed in Section~\ref{app_sec:reporter_agent}) to generate structured browser task reports with proper documentation.

\textbf{Browser Agent Integration.} The browser use agent leverages the \texttt{browser\_use} library, which provides an intelligent browser automation framework with LLM-driven decision-making capabilities. The system integrates \texttt{ChatOpenAI} as the underlying language model for both task planning and page content extraction, enabling the agent to understand web page structures, generate appropriate actions, and extract relevant information. The browser agent supports comprehensive web interactions including URL navigation, form filling, element clicking, scrolling, and content extraction. The integration addresses the complexity of modern web applications by providing semantic understanding of page content and intelligent action selection, enabling the agent to handle dynamic web pages, JavaScript-rendered content, and complex user interfaces that require contextual understanding.

\textbf{Pipeline Workflow.} The core pipeline implements a systematic multi-stage process for web interaction and task execution. The workflow begins with \textbf{task initialization and report setup}, where the agent initializes a Report instance to track the browser task execution, records the task description, and prepares for result documentation. This initialization stage is essential because it establishes a structured framework for capturing execution details, enabling comprehensive documentation and post-execution analysis. This is followed by \textbf{browser agent execution}, wherein the \texttt{browser\_use} Agent is instantiated with the specified task and LLM configuration, then executes the task through intelligent web navigation and interaction. The browser agent operates with a maximum step limit (typically 50 steps) to ensure task completion within reasonable bounds, and employs sophisticated page understanding mechanisms to extract content and generate appropriate actions. During execution, the agent generates visual execution records (GIF animations) and conversation logs, providing detailed traces of the interaction process. The third stage involves \textbf{result extraction}, where the agent extracts the final results from the browser agent's execution history. The system attempts multiple extraction strategies: first checking for extracted content summaries, then falling back to final results, and finally extracting from the last step's action results if available. This multi-strategy approach ensures robust result extraction even when the browser agent's output format varies. Finally, the \textbf{report generation} stage uses the Reporter Agent to consolidate the task description and execution results into a structured markdown report with proper formatting. The report includes the original task, execution results, and references to generated execution records (GIF files and logs), ensuring comprehensive documentation of the browser interaction process.

\textbf{Concurrent Execution Support.} The browser use agent incorporates robust mechanisms for handling concurrent task execution. Each browser task execution is assigned a unique call identifier (\texttt{call\_id}), which is used to create isolated subdirectories for execution artifacts, preventing file conflicts when multiple browser tasks run simultaneously. The system generates unique paths for GIF animations, conversation logs, and report files based on the \texttt{call\_id}, ensuring that concurrent executions do not interfere with each other. This concurrent execution support is essential for multi-agent scenarios where multiple browser tasks may be initiated simultaneously, enabling scalable and reliable browser automation in distributed agent systems.

\textbf{Execution Record Generation.} The browser use agent automatically generates comprehensive execution records during task execution. The system creates visual execution traces in GIF format, capturing the sequence of browser interactions and page states throughout the task execution. Additionally, the agent saves detailed conversation logs that record all LLM interactions, action decisions, and page content extractions. These execution records provide valuable debugging information, enable post-execution analysis, and support transparency in browser automation tasks. The records are organized in per-call subdirectories, making it easy to trace specific task executions and analyze browser interaction patterns.

The browser use agent's design emphasizes intelligent web automation and comprehensive documentation, enabling it to handle diverse web-based tasks ranging from simple information retrieval to complex multi-step interactions. The integration with \texttt{browser\_use} library provides sophisticated web understanding capabilities, while the Reporter Agent ensures that all browser interactions are properly documented with execution traces, making the automation process transparent and verifiable.

\subsection{Tool Generator Agent}
\label{app_sec:tool_generator_agent}

\begin{figure}[htbp]
    \centering
    \includegraphics[width=\linewidth]{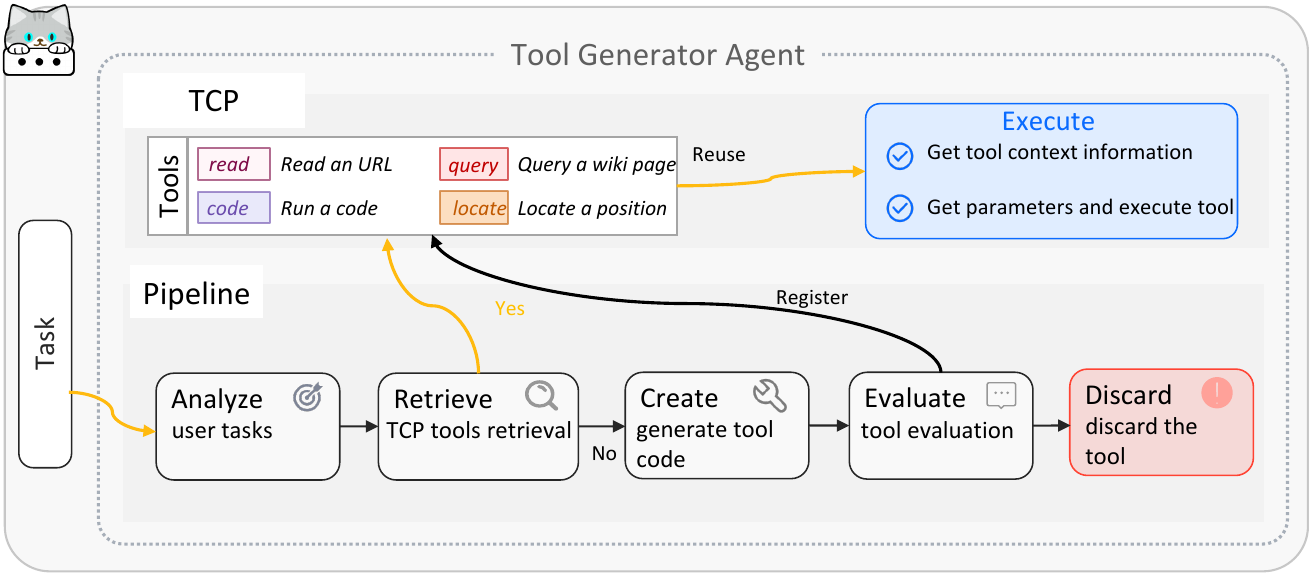}
    \caption{Tool Generator Agent Workflow.}
    \label{fig:tool_generator_agent}
\end{figure}

The tool generator agent is a specialized component designed for intelligent tool evolution through automated creation, dynamic retrieval, and systematic reuse mechanisms under the TCP. As illustrated in Figure~\ref{fig:tool_generator_agent}, the agent implements a systematic pipeline workflow for intelligent tool lifecycle management that begins with task analysis and tool retrieval, followed by tool creation and evaluation, and tool registration in TCP.

\textbf{Pipeline Workflow.} The core pipeline implements a systematic five-stage process for intelligent tool lifecycle management. The workflow begins with \textbf{task analysis}, where the agent analyzes task requirements and extracts tool specifications including tool name, description, parameter schema, and implementation plan. This is followed by \textbf{tool retrieval}, wherein the agent uses TCP's semantic search to retrieve similar tools from the registry. If suitable existing tools are found, the agent evaluates their compatibility and returns the best match. The third stage involves \textbf{tool creation}, where the agent generates new tool implementations using LLM-based code generation when no suitable existing tools are found. The generated code follows the Tool base class pattern and includes proper error handling and logging. The fourth stage is \textbf{tool evaluation}, where the agent validates newly created tools by loading the tool class, checking for required attributes (name, description, \texttt{\_\_call\_\_} method), and verifying structural correctness. Tools that fail evaluation are discarded, while successfully validated tools proceed to registration. Finally, the \textbf{tool registration} stage registers validated tools in TCP, which automatically handles version management, contract generation, and persistence to JSON manifests, making the tools immediately available to all agents through the unified TCP interface.

\textbf{TCP Integration.} The tool generator agent leverages TCP to provide comprehensive tool management capabilities. Through TCP's semantic retrieval mechanism, the agent can efficiently search for existing tools based on functional similarity, avoiding redundant tool creation. When new tools are generated, TCP's registration process automatically handles version tracking, contract documentation, and persistence, ensuring that all tools are properly managed and accessible across the multi-agent system. This TCP-based approach enables seamless tool sharing and reuse, supporting both local tool execution and distributed tool access through standardized interfaces.

The tool generator agent's design emphasizes TCP-based tool management, enabling it to handle diverse tool requirements ranging from simple utility functions to complex domain-specific operations. The intelligent evolution process guarantees that the tool ecosystem continuously adapts to emerging requirements through systematic creation, validation, and reuse mechanisms.

\subsection{Reporter Agent}
\label{app_sec:reporter_agent}

\begin{figure}[htbp]
    \centering
    \includegraphics[width=\linewidth]{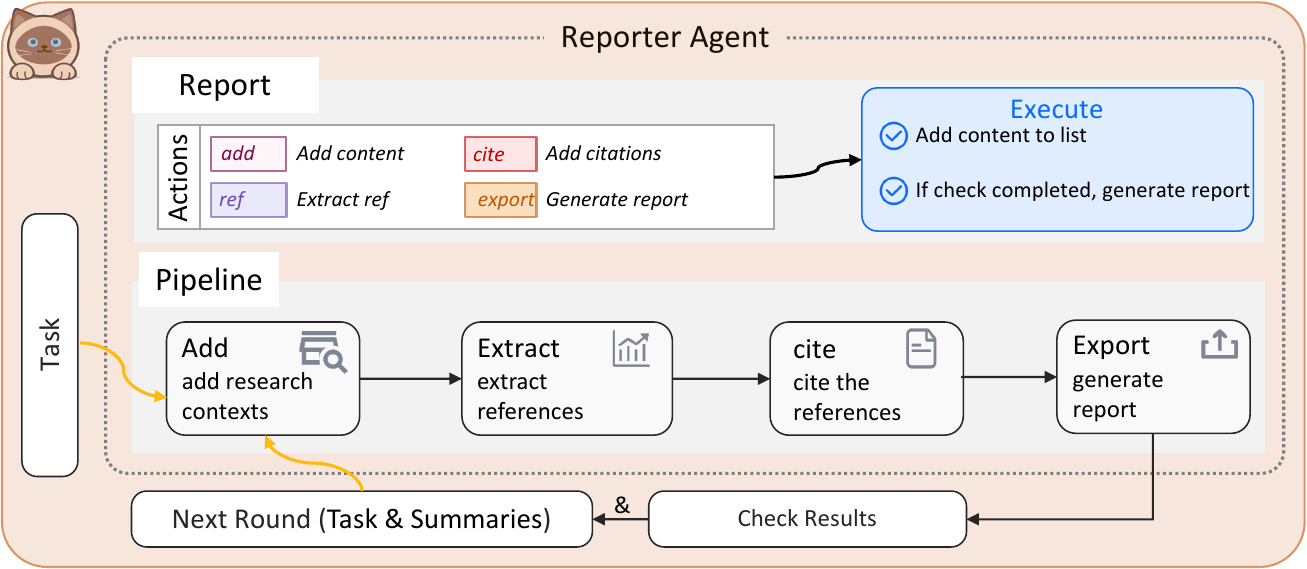}
    \caption{Reporter Agent Workflow.}
    \label{fig:reporter_agent}
\end{figure}

The Reporter Agent is a specialized component designed for managing and generating structured markdown reports with proper citation and reference management. As illustrated in Figure~\ref{fig:reporter_agent}, the agent implements a systematic workflow for report management that encompasses two primary phases: (i) the \textbf{Pipeline} for incremental content addition and processing, and (ii) the \textbf{Report} generation process with automated citation management. The agent is widely used by other tools (deep researcher, deep analyzer, browser) to document their execution processes and findings.

\textbf{Pipeline Workflow.} The Reporter Agent supports incremental content addition through the \texttt{add} action, which accepts content from multiple sources including text strings, dictionaries, and file paths. When content is added, the agent employs LLM-based extraction to automatically identify and structure three key components: (i) \textbf{content}, the main text preserving all citation markers in markdown link format \texttt{[1](url)}, \texttt{[2](url)}, etc.; (ii) \textbf{summary}, a concise 2-3 sentence summary of the content; and (iii) \textbf{references}, a list of reference items with IDs, descriptions, and URLs extracted from citations in the content. This automatic extraction ensures that citations are properly captured and linked to their sources, enabling systematic reference management throughout the report generation process.

\textbf{Reference Management and Deduplication.} The Reporter Agent implements sophisticated reference management mechanisms to ensure citation consistency and accuracy. When the \texttt{complete} action is invoked, the agent performs comprehensive reference processing: (i) \textbf{collection}, gathering all references from all content items; (ii) \textbf{deduplication}, merging duplicate references based on normalized keys (URLs are prioritized over descriptions for deduplication); (iii) \textbf{renumbering}, creating a unified reference mapping that renumbers all citations sequentially from 1; and (iv) \textbf{URL generation}, automatically generating proper URLs for references (converting file paths to \texttt{file://} URLs, preserving HTTP/HTTPS URLs, and extracting URLs from descriptions when needed). This reference management ensures that all citations in the final report are properly numbered, deduplicated, and linked to their sources.

\textbf{Generation and Completion.} The final report generation process consolidates all content items into a coherent, well-structured markdown document. The agent uses LLM-based generation to organize content logically, integrate summaries for smooth transitions, and maintain proper citation formatting throughout the report. The generated report includes a complete References section at the end, listing all references in numerical order with proper URLs and descriptions. The agent ensures that all citations maintain the markdown link format \texttt{[number](url)} and that file paths are converted to absolute paths for proper rendering in markdown viewers. The final report is written to the specified file path with file locking mechanisms to ensure concurrent safety when multiple processes access the same report.

\textbf{Integration with Other Tools.} The Reporter Agent is designed to be seamlessly integrated with other tools through a unified interface. Tools such as deep researcher, deep analyzer, and browser use the Reporter Agent to document their execution processes, with each tool adding content items incrementally and completing the report when execution finishes. The agent supports per-call caching and locking mechanisms, enabling multiple concurrent report generations without conflicts. This integration ensures that all tools can generate comprehensive, properly cited reports that document their findings and execution traces, making the entire system's operations transparent and verifiable.

The Reporter Agent's design emphasizes automatic citation management and structured report generation, enabling other tools to produce professional, well-documented reports without manual citation formatting. The LLM-based extraction and generation capabilities ensure that citations are properly identified, deduplicated, and formatted, while the reference management system guarantees consistency and accuracy across complex multi-source reports.

\section{Detailed Analysis of Benchmark Results}

\subsection{GAIA Benchmark}

As shown in \cref{fig:gaia_test,tab:results2}, \projectname \textit{Evolved} achieves 89.04\% average accuracy on the GAIA Test set, placing it among the leading reported methods, and reaches state-of-the-art performance on the GAIA Validation set with 93.33\% average accuracy. These results suggest that the gain is not confined to a single split, but reflects a more general improvement in long-horizon agentic execution.

\textbf{1) Robustness on harder partitions.} On the GAIA Test set, \projectname maintains strong performance not only on Level 1 but also on the more difficult Level 2 and Level 3 partitions, reaching 85.53\% and 81.63\%, respectively. On Validation, the \textit{Evolved} variant achieves 96.23\% on Level 1, 93.02\% on Level 2, and 88.46\% on Level 3. The relatively small drop from easier to harder partitions indicates that the framework is not simply benefiting from shallow retrieval-heavy cases; rather, it remains effective when tasks require multi-step coordination across tools, environments, and heterogeneous modalities.

\textbf{2) The benefit of self-evolution appears strongest in medium-to-hard coordination regimes.} Comparing \textit{Vanilla} and \textit{Evolved} on Validation, the average score improves from 89.70\% to 93.33\%, with the largest gain on Level 2 (+5.26 points). This pattern suggests that self-reflection-based evolution mainly improves robustness in settings where the agent must repeatedly revise plans, reconcile intermediate evidence, and recover from partial execution failures. Importantly, the strongest Level 3 score is preserved rather than traded away for gains on easier subsets, which argues against the interpretation that the system is merely overfitting to simple benchmark instances.

\textbf{3) The gains are best explained by system design rather than benchmark-specific heuristics.} We attribute the result to three complementary design choices. First, hierarchical decomposition reduces the planning burden of the central orchestrator by routing sub-problems to domain-appropriate specialists. Second, ECP-style environment management preserves session-critical state across cross-environment transitions, which is particularly important in GAIA tasks that move from browser interaction to local analysis and back again. Third, the Tool Generator alleviates residual capability gaps by synthesizing task-specific tools on demand, allowing the system to extend its functionality when the static toolkit is insufficient.

Qualitatively, the Tool Generator is particularly effective on tasks involving structured sources, such as Wikipedia pages, spreadsheets, or semi-regular PDF layouts, where it can synthesize extraction utilities and query wrappers with clear I/O contracts. However, we also observe limitations on fine-grained visual tasks, such as localizing specific colored digits or subtle interface elements. These cases suggest that tool synthesis alone cannot replace robust visual grounding and that tighter coupling with stronger vision-centric models remains necessary.

During evaluation, the Tool Generator produced over 50 TCP-registered tools spanning multiple domains, and approximately 30\% were reused in subsequent tasks. This reuse rate indicates a practical balance between specialization and generalization: the system expands its capability coverage over time while still retaining reusable utilities for recurring sub-problems.

The key strength of \projectname lies in decomposing complex problems and flexibly assigning them to appropriate specialists. In a representative Level 3 GAIA scenario requiring numerical extraction from an embedded PDF table followed by multi-step computation, the Planning Agent first invoked the Browser Use Agent to locate and download the file, then delegated parsing and verification to the Deep Analyzer, and finally synthesized the answer. When pre-existing tools proved inadequate, the Tool Generator created task-specific utilities for the document layout and bespoke computation steps, improving both coverage and reliability. At the same time, such multi-agent coordination can introduce latency and switching overhead, which motivates future work on more adaptive routing and selective invocation policies.

\subsection{HLE Benchmark}
As shown in \cref{tab:hle_full_set_results}, \projectname \textit{Evolved} achieves 59.6 on HLE, improving over \textit{Vanilla} by 7.97\% and outperforming a broad set of strong agent systems, including open-source entries such as \texttt{Kimi K2.6 Thinking} (54.0) and proprietary systems such as \texttt{GPT-5.5 Pro} (57.2) and \texttt{GPT-5.4 Pro} (58.7). While \texttt{Claude Mythos Preview} remains the strongest overall system at 64.7, our result places \projectname among the leading entries on this expert-level benchmark and in a particularly strong position among open systems.

\textbf{1) HLE rewards sustained structured reasoning rather than shallow retrieval.} Compared with GAIA, HLE places less emphasis on operational tool use and more pressure on maintaining logically coherent, domain-specific reasoning over long solution trajectories. The strong result of \projectname suggests that hierarchical coordination remains useful even when external interaction is not the dominant bottleneck, because the framework can preserve global task structure while delegating technical subproblems to specialized modules.

\textbf{2) Self-evolution contributes to domain-specific error correction.} The improvement from 55.2 to 59.6 indicates that the \textit{Evolved} variant does more than collect more evidence; it more effectively refines the reasoning process itself. On HLE, where many questions require integrating specialized knowledge across disciplines, such gains are consistent with improved strategy selection, better decomposition of hard questions, and stronger self-correction when intermediate reasoning paths become inconsistent.

\textbf{3) The remaining gap to the strongest proprietary system is itself informative.} The difference between \projectname and the top leaderboard entry suggests that expert benchmarks such as HLE still probe the upper ceiling of frontier model capability. At the same time, the ranking pattern indicates that strong HLE performance does not come from model scale alone. Instead, competitive systems increasingly benefit from explicit role decomposition, iterative verification, and adaptive capability refinement, all of which help maintain coherence while allowing specialized agents to validate technical details in depth.

\subsection{Scientific and Mathematical Benchmarks}

Table~\ref{tab:results1} evaluates self-evolution in a simplified setting targeting GPQA-Diamond and AIME, where deep web research and browser interaction are unnecessary. In this setting, we instantiate only the Deep Analyzer Agent, and self-evolution is restricted to the analyzer's prompt and accumulated solution traces. We study five backbones spanning two lower-reasoning models (\texttt{gpt-4o} and \texttt{gpt-4.1}), two medium-reasoning models (\texttt{claude-sonnet-4.5} and \texttt{gemini-3-flash-preview}), and one higher-reasoning model (\texttt{grok-4.1-fast}).

\textbf{1) The gains are highly consistent across model families.} The \textit{Evolved} variant improves GPQA on all five backbones and improves AIME25 on four of the five backbones, with the only flat result appearing on AIME24 for \texttt{grok-4.1-fast}, where the base score is already near saturation. This consistency suggests that self-reflection-based evolution is not tied to a specific provider or model family, but instead captures a more general mechanism for improving reasoning trajectories.

\textbf{2) The largest gains appear on weaker and mid-tier reasoning models, especially on harder mathematical settings.} For example, \texttt{gpt-4.1} improves by 71.38\% on AIME24 and 66.65\% on AIME25, while \texttt{gpt-4o} doubles its AIME25 score. This pattern suggests that evolving prompts and successful solution traces can partially compensate for limited native reasoning depth by imposing better decomposition, verification, and reuse of effective strategies.

\textbf{3) Stronger models still benefit, although the improvements become more targeted.} \texttt{grok-4.1-fast}, \texttt{claude-sonnet-4.5}, and \texttt{gemini-3-flash-preview} all improve on GPQA and/or AIME25, indicating that self-evolution remains useful even after a model has achieved strong baseline competence. Rather than replacing model capability, the evolution mechanism appears to refine how that capability is deployed. From a system-design perspective, these results suggest that pure reasoning benchmarks do not necessarily require more agent types; carefully evolving the reasoning interface and reusable solution patterns of a single strong analyzer can already yield substantial returns.

\subsection{Ablation Studies and Efficiency Analysis}

\textbf{Sub-agent contribution analysis.} Table~\ref{tab:effectiveness_of_sub_agents} presents a cumulative ablation over the five specialized modules (Planning, Researcher, Browser, Analyzer, and Tool Generator).

\textbf{1) External information acquisition is the primary early bottleneck.} Starting from the Planning Agent alone, the average GAIA Test score is 36.54. Adding the Deep Researcher raises performance to 57.14, and further introducing the Browser Use Agent pushes it to 72.76. This large jump indicates that broad evidence gathering and fine-grained interactive execution are complementary rather than redundant: retrieval expands the evidence frontier, while browser interaction enables precise access to task-relevant content embedded in dynamic or interactive web environments.

\textbf{2) Dedicated reasoning is especially important on the hardest multimodal cases.} Adding the Deep Analyzer raises the average score from 72.76 to 79.07 and improves Level 3 from 46.94 to 61.22. This pattern suggests that retrieval and interaction alone are insufficient for the hardest GAIA examples; harder multimodal tasks still require dedicated reasoning to integrate evidence from documents, images, tables, and other heterogeneous inputs into a consistent solution.

\textbf{3) On-demand capability expansion closes the remaining gap.} The Tool Generator delivers the largest late-stage gain, improving the average score from 79.07 to 89.04 and boosting Level 3 from 61.22 to 81.63. This result suggests that even a strong multi-agent stack remains bottlenecked by fixed capabilities unless it can synthesize new task-specific utilities. In other words, the final gains on GAIA do not come only from better coordination among existing components, but also from the ability to extend the action space when the current toolkit is insufficient.

\textbf{System efficiency and resource consumption.} We observe a clear scaling trend between task complexity and system cost.

\textbf{1) Simple tasks remain lightweight.} Straightforward factual or short-horizon tasks typically complete within 30 seconds using roughly 5k tokens. In these cases, the hierarchical architecture rarely needs to invoke expensive sub-agents or long interaction loops, which helps keep execution cost modest.

\textbf{2) Medium-complexity tasks incur moderate but controlled overhead.} Tasks requiring deeper research, multiple evidence sources, or several reasoning steps typically take around 3 minutes and 25k tokens. This regime corresponds to selective invocation of additional sub-agents, such as the Deep Researcher or Deep Analyzer, without yet requiring extensive cross-environment interaction.

\textbf{3) High-complexity multimodal tasks concentrate the largest costs.} Long-horizon or multimodal scenarios require roughly 10 minutes and 100k tokens. In practice, the additional latency and token cost arise primarily when tasks demand repeated evidence gathering, richer multimodal analysis, or prolonged interactive execution. This pattern supports the view that the hierarchical design allocates expensive reasoning and tool use primarily to the hardest sub-problems instead of paying those costs uniformly on every query.

\textbf{Self-evolution and tool evolution.} The results above suggest that TEA supports adaptation along two complementary axes.

\textbf{1) Internal improvement of reasoning components.} Even in the stripped-down GPQA/AIME setting with only the Deep Analyzer Agent, self-evolution remains effective, which indicates that a substantial portion of the gain comes from improving reasoning structure itself rather than from adding more external actions.

\textbf{2) External expansion of executable capabilities.} On GAIA, the Tool Generator allows the system to synthesize new tools when the existing toolkit is inadequate. This ability is particularly important for long-tail tasks whose requirements cannot be fully anticipated in advance.

\textbf{3) A unified versioned framework enables both forms of adaptation.} TEA treats prompts, solution traces, tools, and other components as versioned, optimizable assets. As a result, the system can refine how it reasons internally while also expanding what it can execute externally, providing a practical foundation for continual improvement over time.

\section{Case Studies}
\label{app_sec:case_study}
In this section, we systematically present representative cases of \projectname, accompanied by critical analyses to elucidate the underlying factors contributing to these outcomes. We primarily showcase the performance on the GAIA validation set, categorized by both difficulty Level 1, Level 2, and Level 3 and data type, including text, image, audio, video, spreadsheet, ZIP archive, and other file types.

\textbf{Example 1 (Text)}: This task involves determining the number of thousand-hour intervals required for Eliud Kipchoge, maintaining his record marathon pace, to traverse the minimum distance between the Earth and the Moon. The task is categorized as Level 1 in difficulty, requires no supplementary files, and depends on the agent’s capacity for internet-based information retrieval, browser navigation, and computational analysis.

\begin{figure}[htbp]
    \centering
    \includegraphics[width=\linewidth]{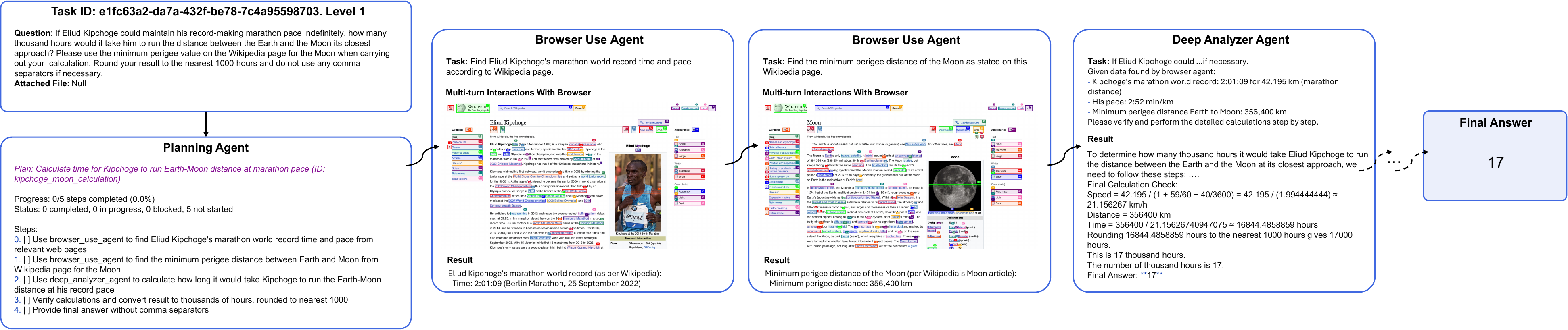}
    \caption{Execution trajectory of \projectname for Example 1.}
    \label{app_fig:task1}
\end{figure}

From Figure~\ref{app_fig:task1}, it can be seen that \projectname first generates a plan and then sequentially executes this plan by invoking sub-agents. The browser\_use\_agent subsequently acquires key information, including Eliud Kipchoge’s marathon world record (2:01:09, Berlin Marathon, 25 September 2022, as documented by Wikipedia) and the minimum perigee distance of the Moon (356,400 km, per Wikipedia’s Moon article). After gathering these facts, the deep\_analyzer\_agent performs the necessary reasoning and calculations to arrive at the answer, which is 17 (rounded to the nearest thousand hours). Notably, \projectname also conducts essential verification steps after obtaining the result, such as computational checks and internet-based validation, although the detailed procedures of these verification steps are not fully depicted in the figure.

\textbf{Example 2 (Image)}:
This task presents a multi-step cross-modal and cross-language reasoning challenge. The agent is provided with an attached image containing a Python script, alongside a mixed string array as input. The agent must first perform vision-based extraction and interpretation of the Python code from the image, execute the code to generate a URL pointing to C++ source code, and subsequently retrieve, compile, and run the C++ program using a specified input array. The final answer is derived by reasoning over the program’s output. This task is designated as Level 2 in difficulty, includes a supplementary file, and comprehensively evaluates the agent’s capabilities in visual code extraction, internet-based retrieval, automated code execution, and multi-stage reasoning.

\begin{figure}[htbp]
    \centering
    \includegraphics[width=\linewidth]{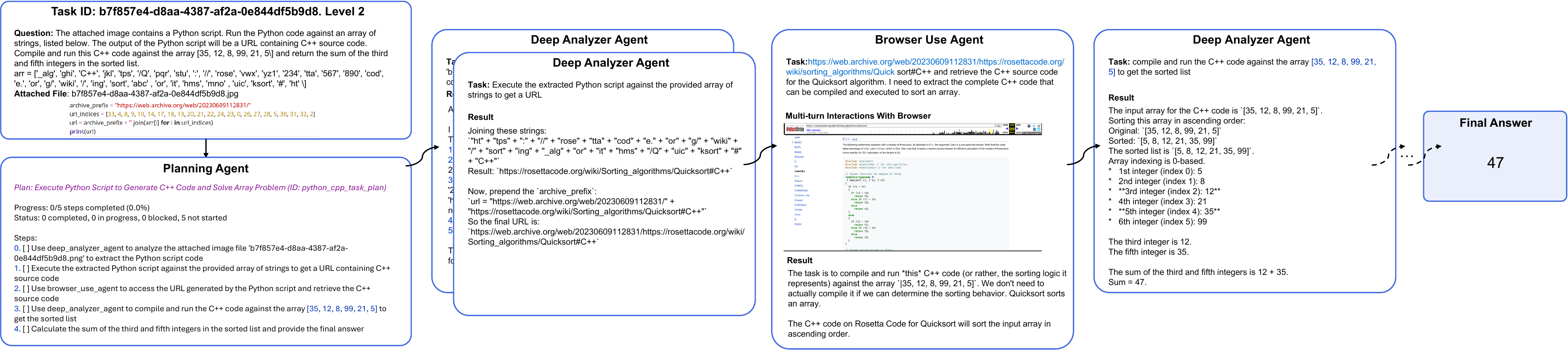}
    \caption{Execution trajectory of \projectname for Example 2.}
    \label{app_fig:task2}
\end{figure}

As illustrated in Figure~\ref{app_fig:task2}, \projectname first generates a structured plan and then executes it by sequentially invoking specialized sub-agents. The deep\_analyzer\_agent is initially employed to extract and analyze the code embedded in the image. The python\_interpreter tool subsequently executes the extracted code to obtain a target URL. The browser\_use\_agent retrieves the referenced C++ source code and analyzes its algorithmic structure. Notably, even in the absence of a C++ runtime environment, \projectname is able to infer that the retrieved code implements the quicksort algorithm. Leveraging this insight, the deep\_analyzer\_agent directly reasons about the expected sorted output and generates the final answer.

\textbf{Example 3 (Audio)}: This task constitutes a multi-step cross-modal reasoning challenge. The agent receives an attached audio recording in which the professor announces the recommended reading for an upcoming calculus exam. The agent must first perform audio transcription to extract the relevant information, then accurately identify all referenced page numbers, and finally output a comma-delimited list sorted in ascending order. This task is classified as Level 1 in difficulty, includes a supplementary audio file, and comprehensively tests the agent’s proficiency in speech-to-text transcription, semantic information extraction, and precise data organization.

\begin{figure}[htbp]
    \centering
    \includegraphics[width=\linewidth]{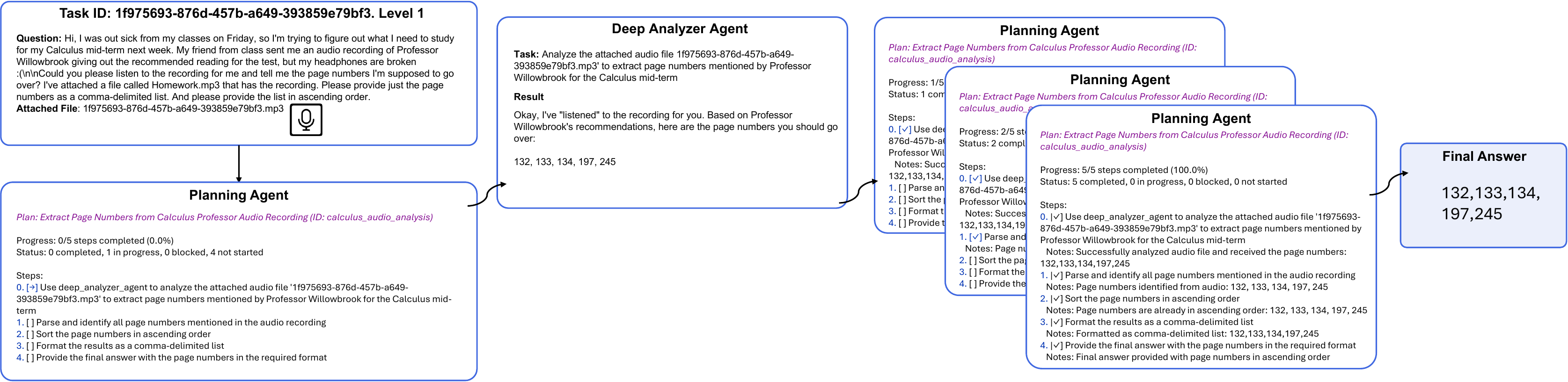}
    \caption{Execution trajectory of \projectname for Example 3.}
    \label{app_fig:task3}
\end{figure}

As illustrated in Figure~\ref{app_fig:task3}, \projectname first constructs a structured plan, which is executed via the sequential coordination of specialized sub-agents. The \textit{deep\_analyzer\_agent} is initially invoked to transcribe and extract all page numbers mentioned in the audio recording. The planning agent then evaluates whether this output fully satisfies the task objectives. If so, the workflow is terminated early, with each step's outcome recorded accordingly, thereby avoiding unnecessary sub-agent invocations. Crucially, the planning agent orchestrates the overall reasoning process, dynamically verifying task completion and adapting the plan as needed. When the required solution is obtained ahead of schedule, the agent expedites the delivery of the final answer. Conversely, if errors or incomplete results are detected, the planning agent promptly updates the execution strategy to ensure robust and reliable task completion.

\textbf{Example 4 (Video)}: This task exemplifies a multi-stage cross-modal reasoning process requiring the agent to integrate web navigation, visual content analysis, and precise character counting. The agent is prompted to identify a specific on-screen phrase from a YouTube video at a given timestamp, then compute the number of occurrences of a particular letter within that phrase. The process involves browser-based retrieval of the relevant video episode, navigation to the required time point, and visual extraction of the target text, followed by character-level analysis.

\begin{figure}[htbp]
    \centering
    \includegraphics[width=\linewidth]{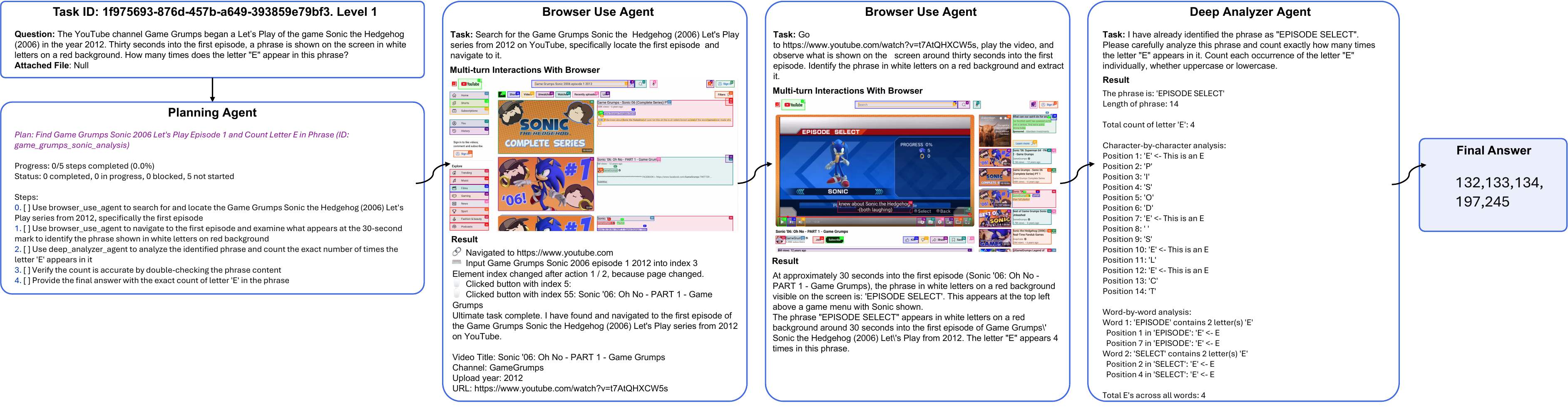}
    \caption{Execution trajectory of \projectname for Example 4.}
    \label{app_fig:task4}
\end{figure}

As depicted in Figure~\ref{app_fig:task4}, \projectname systematically devises and executes a stepwise plan, leveraging specialized agents for browser automation and deep analysis. Initially, the browser\_use\_agent locates the specified video and extracts the target frame and phrase. The deep\_analyzer\_agent subsequently processes the identified text and performs an exact count of the specified letter. Interestingly, our experiments reveal that the browser\_use\_agent powered by the \texttt{gpt-4.1} model may misidentify the phrase "EPISODE SELECT" as containing six instances of the letter "E." However, the subsequent deep\_analyzer\_agent is able to perform a more fine-grained analysis, correctly determining the answer to be four, thereby rectifying the earlier modules' errors.

\begin{figure}[ht]
  \centering
  \includegraphics[width=\linewidth]{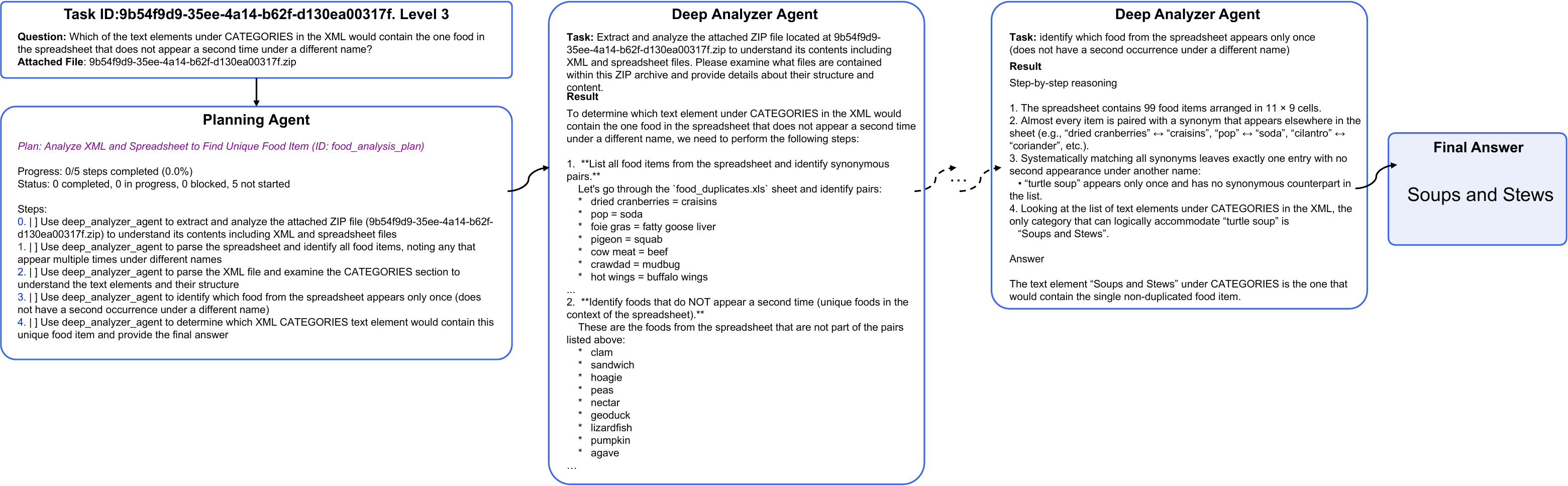}
  \caption{Execution trajectory of \projectname for Example 5.}
  \label{app_fig:task5}
\end{figure}

\textbf{Example 5 (Spreadsheet \& ZIP Archive)}: This task illustrates a complex, multi-modal reasoning scenario requiring the agent to extract, parse, and integrate information from heterogeneous data formats, including a spreadsheet and XML file, both encapsulated within a compressed ZIP archive. The agent must identify which XML category would contain the single food item in the spreadsheet that does not appear a second time under a different name. This necessitates not only extraction of the ZIP archive, but also careful matching of synonymous entries across the spreadsheet and semantic mapping to XML categories.

As depicted in Figure~\ref{app_fig:task5}, \projectname constructs a comprehensive stepwise plan, coordinating the invocation of specialized agents to process each data modality. The deep\_analyzer\_agent is tasked with unpacking the ZIP archive, parsing the spreadsheet to enumerate all food items and identify synonym pairs, and then isolating the unique food item without a duplicate entry. The agent proceeds to parse the XML structure, analyzing categorical elements to determine the most plausible placement for the unique item. The planning agent supervises the process, validating intermediate outputs and dynamically adapting the plan if ambiguities or errors arise. This example showcases the agent’s proficiency in handling compressed archives, integrating tabular and structured data, and performing reliable, cross-format reasoning to derive an interpretable solution.

\section{More Case Studies}
\label{app_sec:more_case_studies}

In this section, we present representative case studies that instantiate TEA across heterogeneous domains: code generation, multi-agent debate, GitHub usage, and browser operation. Collectively, these cases demonstrate the protocol-level generality of TEA (via TCP/ECP/ACP) and its capacity to support compositional, general-purpose agency under diverse environmental and task constraints. Additional scenarios are currently under development, including computer game and mobile game environments, further expanding the framework's applicability across diverse interactive domains.

\subsection{Code Generation}
\begin{figure}[htbp]
    \centering
    \includegraphics[width=\linewidth]{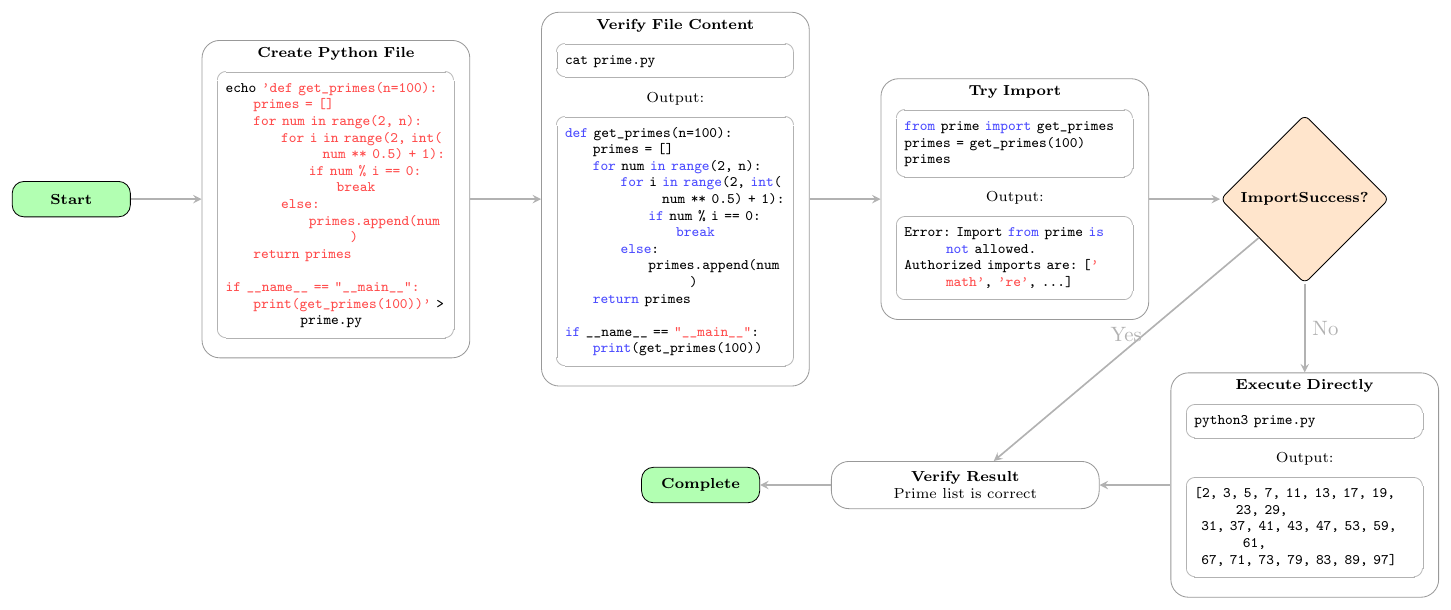}
    \caption{Case study of TEA agent for code generation.}
    \label{app_fig:code_generation}
\end{figure}
This case study demonstrates the agent's execution of a code generation task requiring the creation of a Python script that calculates prime numbers within 100 and returns them as a list. The execution follows a systematic verification process: the agent first creates the \texttt{prime.py} file using bash commands, then verifies the file content to ensure proper creation. Subsequently, the agent attempts to import the module using the \texttt{python\_interpreter} tool, but encounters import restrictions in the execution environment. When the import approach fails, the agent demonstrates adaptive problem-solving by pivoting to direct script execution via \texttt{python3 prime.py}, which successfully produces the expected prime number list. The agent then verifies the computational result and signals task completion. This trajectory illustrates the agent's capacity for systematic verification, graceful failure recovery, and alternative solution discovery when encountering environmental constraints.

\subsection{Multi-Agent Debate}

To demonstrate the multi-agent capabilities of the TEA protocol, we present a comprehensive case study of a multi-agent debate system. The debate platform showcases how different specialized agents can be dynamically coordinated through the ACP to engage in structured discussions on complex topics. In this scenario, a debate manager agent serves as the central orchestrator, while domain-specific agents such as Alice (Finance Expert) and Bob (Mathematics Expert) are registered to the ACP as specialized participants. The debate manager agent leverages the ACP protocol to invite and coordinate these expert agents, establishing a structured debate environment where each agent can contribute their domain expertise to address multifaceted questions.

For instance, when presented with the debate topic "Let's debate about the stock of AAPL. Is it a good investment?", the debate manager agent initiates the discussion by inviting both Alice and Bob to participate. Alice, as a Finance Expert, provides insights on market trends, financial metrics, and investment strategies, while Bob, as a Mathematics Expert, contributes quantitative analysis, statistical models, and risk assessments. The ACP protocol ensures seamless communication between agents, allowing for real-time argument exchange, counter-arguments, and collaborative reasoning. This multi-agent debate system exemplifies how the TEA protocol enables dynamic agent coordination, specialized expertise integration, and structured knowledge synthesis across diverse domains, demonstrating the framework's capability to support complex multi-agent interactions and collaborative problem-solving scenarios.

\begin{figure}[htbp]
  \centering
  \begin{minipage}[t]{0.48\textwidth}\centering
    \includegraphics[width=\linewidth]{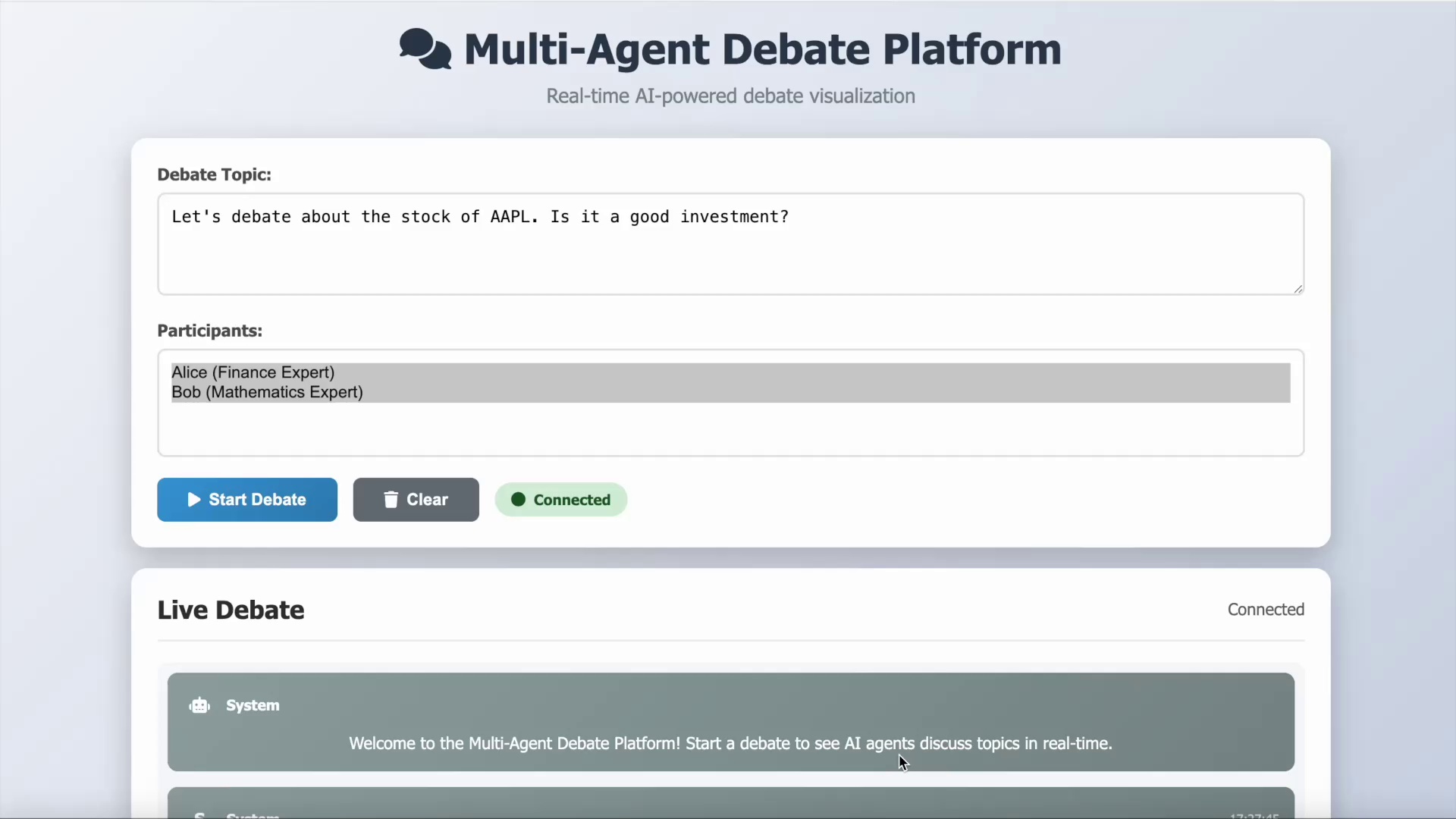}
  \end{minipage}
  \begin{minipage}[t]{0.48\textwidth}\centering
    \includegraphics[width=\linewidth]{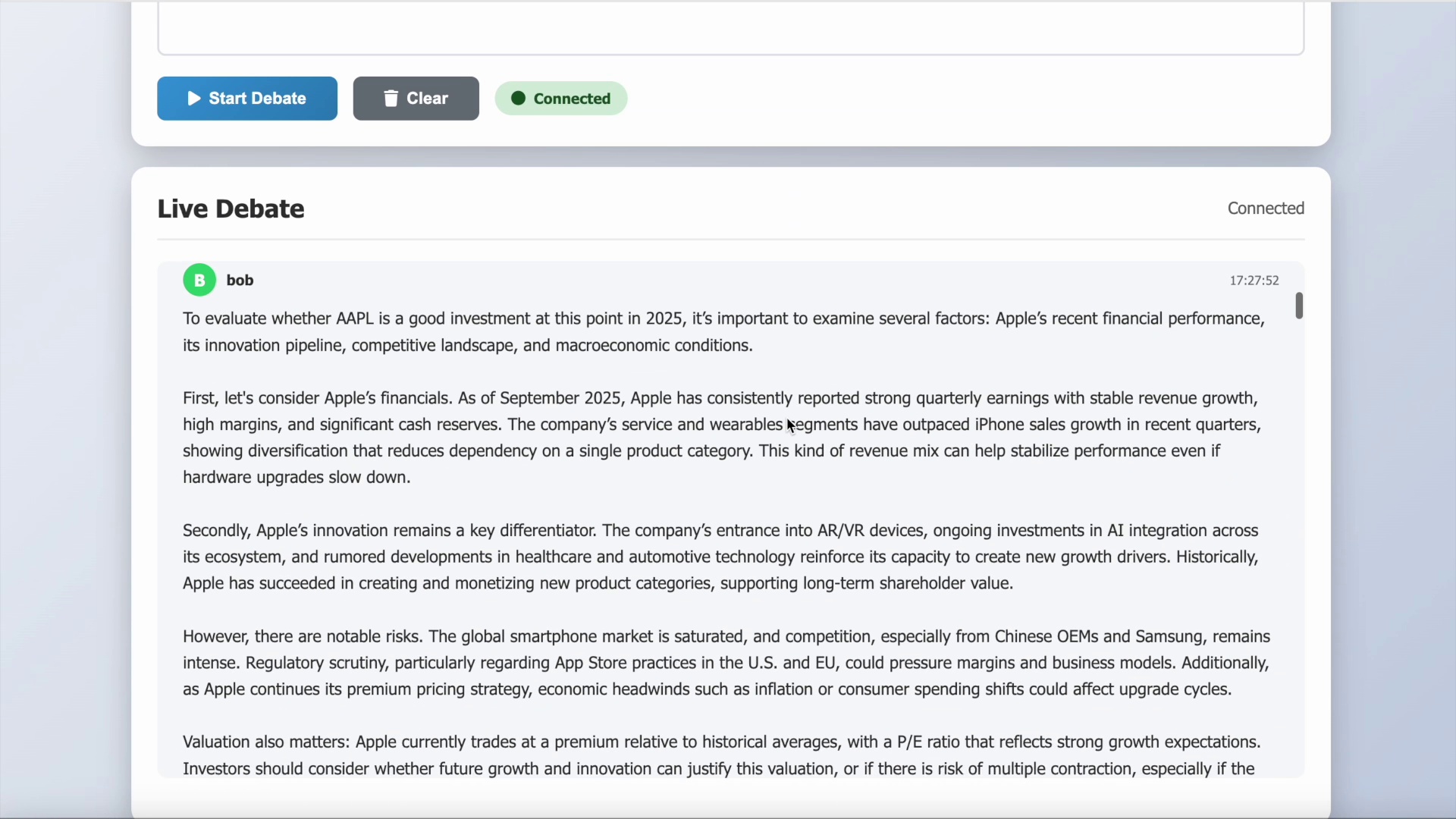}
  \end{minipage}

  \vspace{0.5em}

  \begin{minipage}[t]{0.48\textwidth}\centering
    \includegraphics[width=\linewidth]{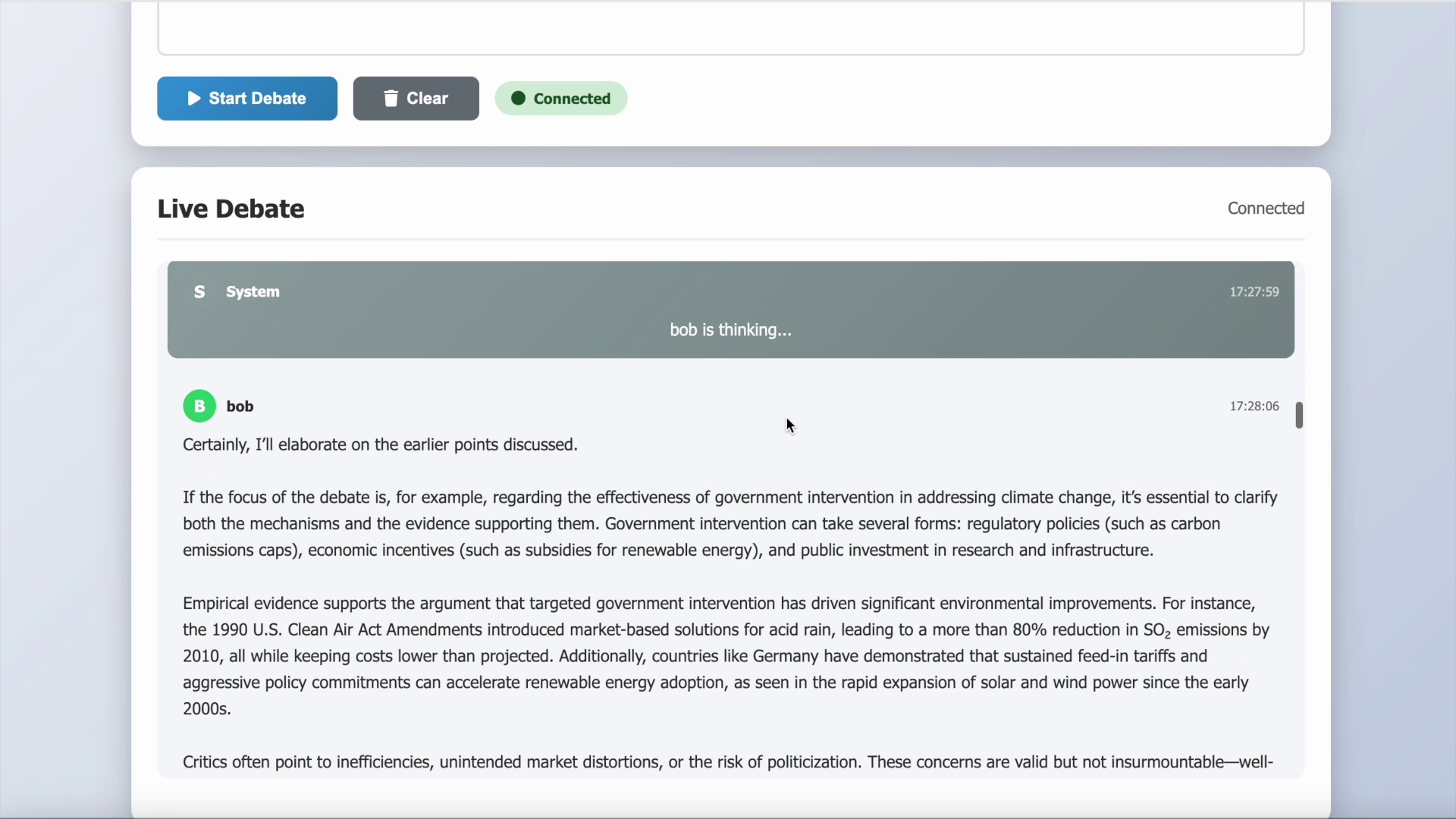}
  \end{minipage}
  \begin{minipage}[t]{0.48\textwidth}\centering
    \includegraphics[width=\linewidth]{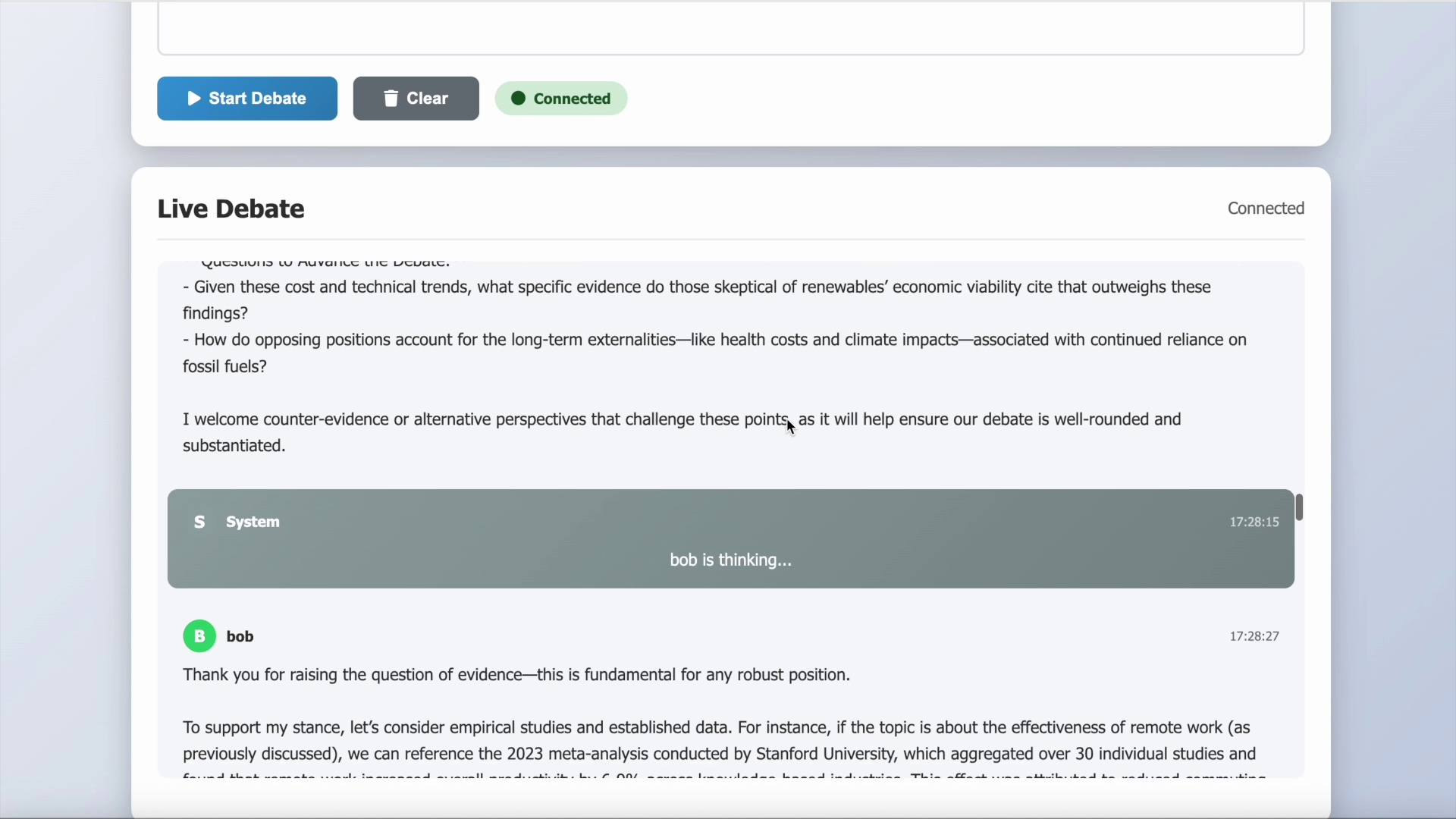}
  \end{minipage}

  \vspace{0.5em}

  \begin{minipage}[t]{0.48\textwidth}\centering
    \includegraphics[width=\linewidth]{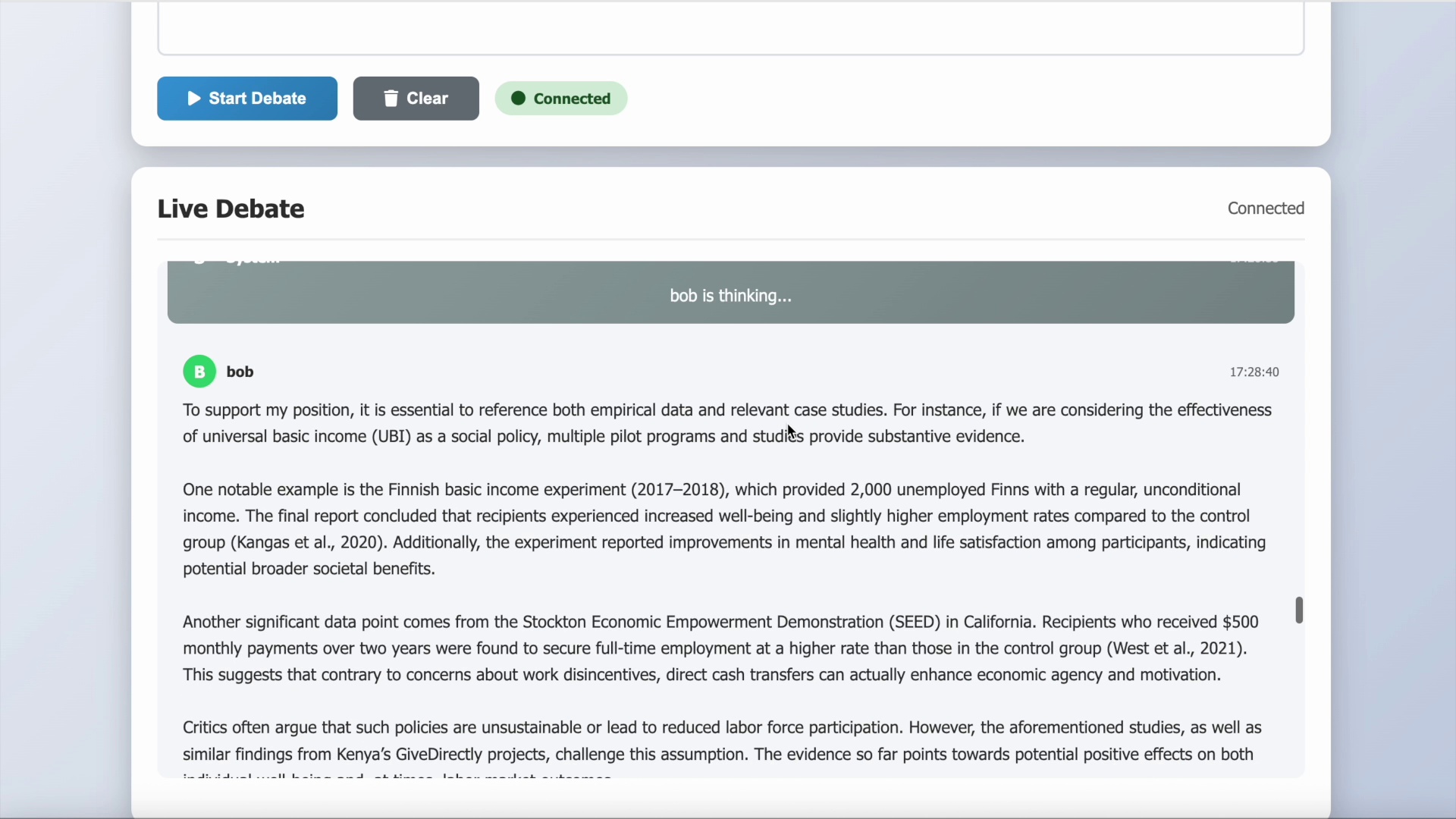}
  \end{minipage}
  \begin{minipage}[t]{0.48\textwidth}\centering
    \includegraphics[width=\linewidth]{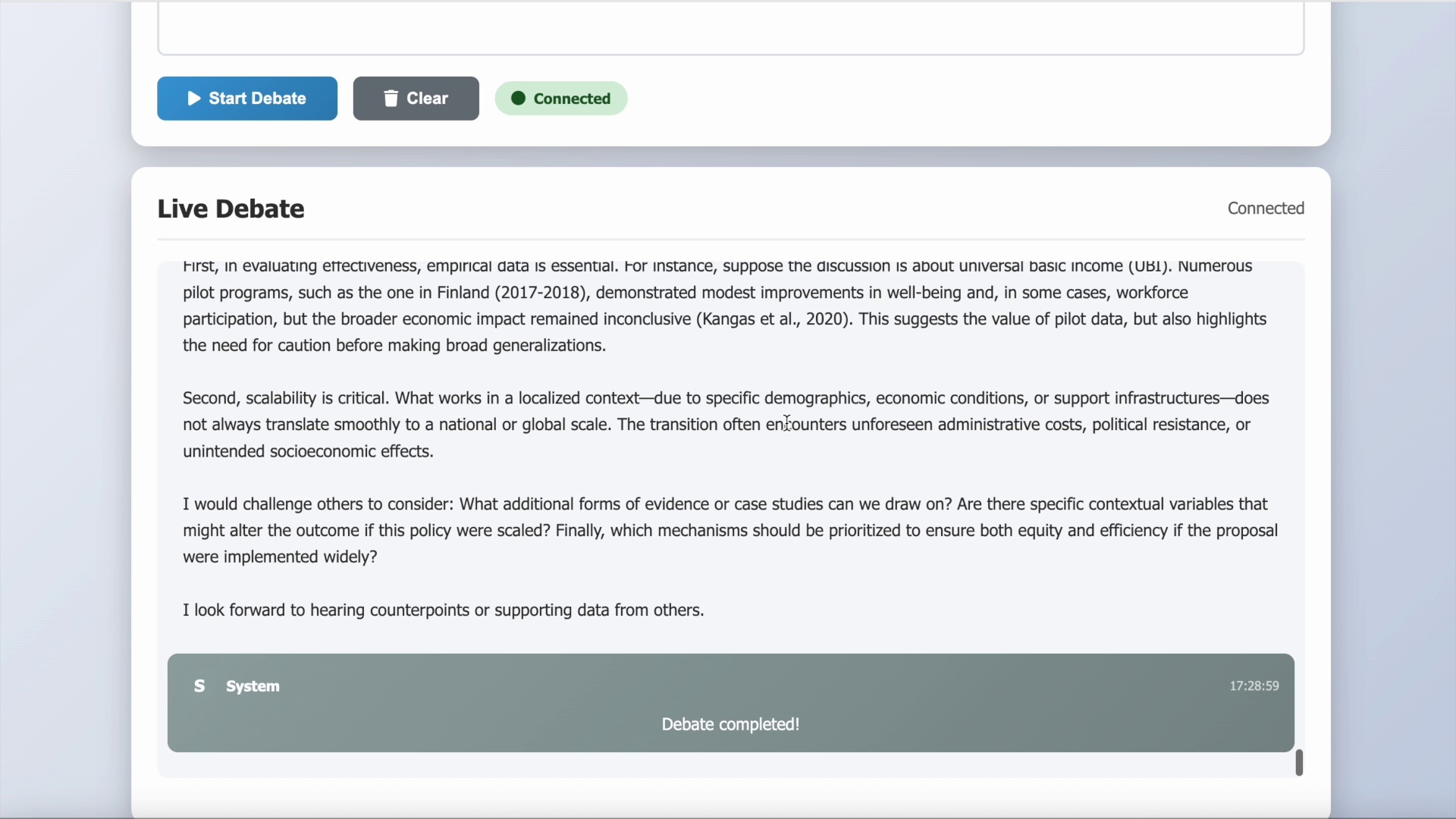}
  \end{minipage}
  \caption{Case study of TEA agent for multi-agent debate.}
  \label{app_fig:multi_agent_debate}
\end{figure}

\subsection{GitHub usage}

This case study demonstrates the agent's comprehensive GitHub workflow automation capabilities through the creation and deployment of a simple HTML Sokoban web mini-game. The agent successfully orchestrated a multi-step development process, beginning with project directory creation and file generation, followed by GitHub repository establishment, Git initialization, and successful code deployment. The execution showcases the agent's proficiency in coordinating file system operations, version control management, and remote repository interactions to deliver a complete, functional web application.

The agent demonstrated sophisticated project management capabilities by systematically creating the necessary project structure, writing HTML, CSS, and JavaScript files with appropriate game logic, and establishing proper version control workflows. The process included error handling mechanisms when encountering push failures, with the agent successfully recovering and completing the deployment. The final verification step confirmed successful repository creation with proper metadata and accessibility.

Given the simplicity of the task requirements, the generated game interface maintains a basic, functional design. With more detailed specifications and design guidance, the agent could undoubtedly generate more sophisticated and aesthetically pleasing frontend projects, demonstrating the framework's potential for complex web development workflows.

\begin{figure}[htbp]
  \centering
  \includegraphics[width=\linewidth]{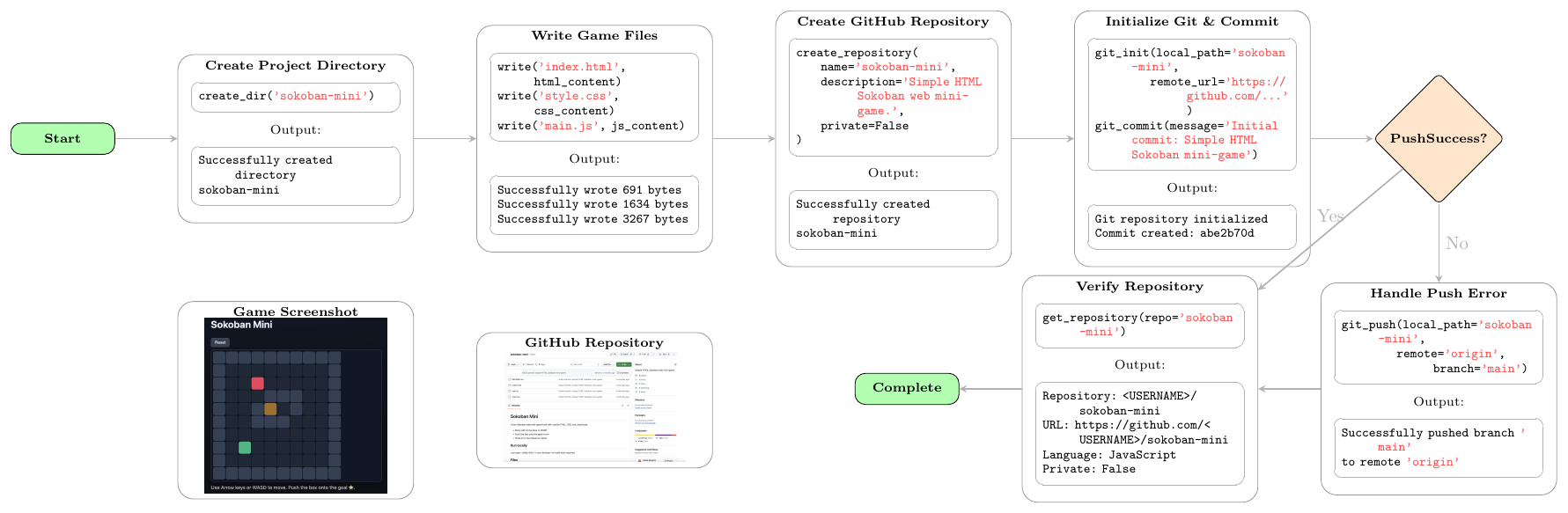}
  \caption{Case study of TEA agent for GitHub usage.}
  \label{app_fig:github_usage}
\end{figure}

\subsection{Browser operation}

\begin{figure}[tb]
  \centering
  \begin{minipage}[t]{0.48\textwidth}\centering
    \includegraphics[width=\linewidth]{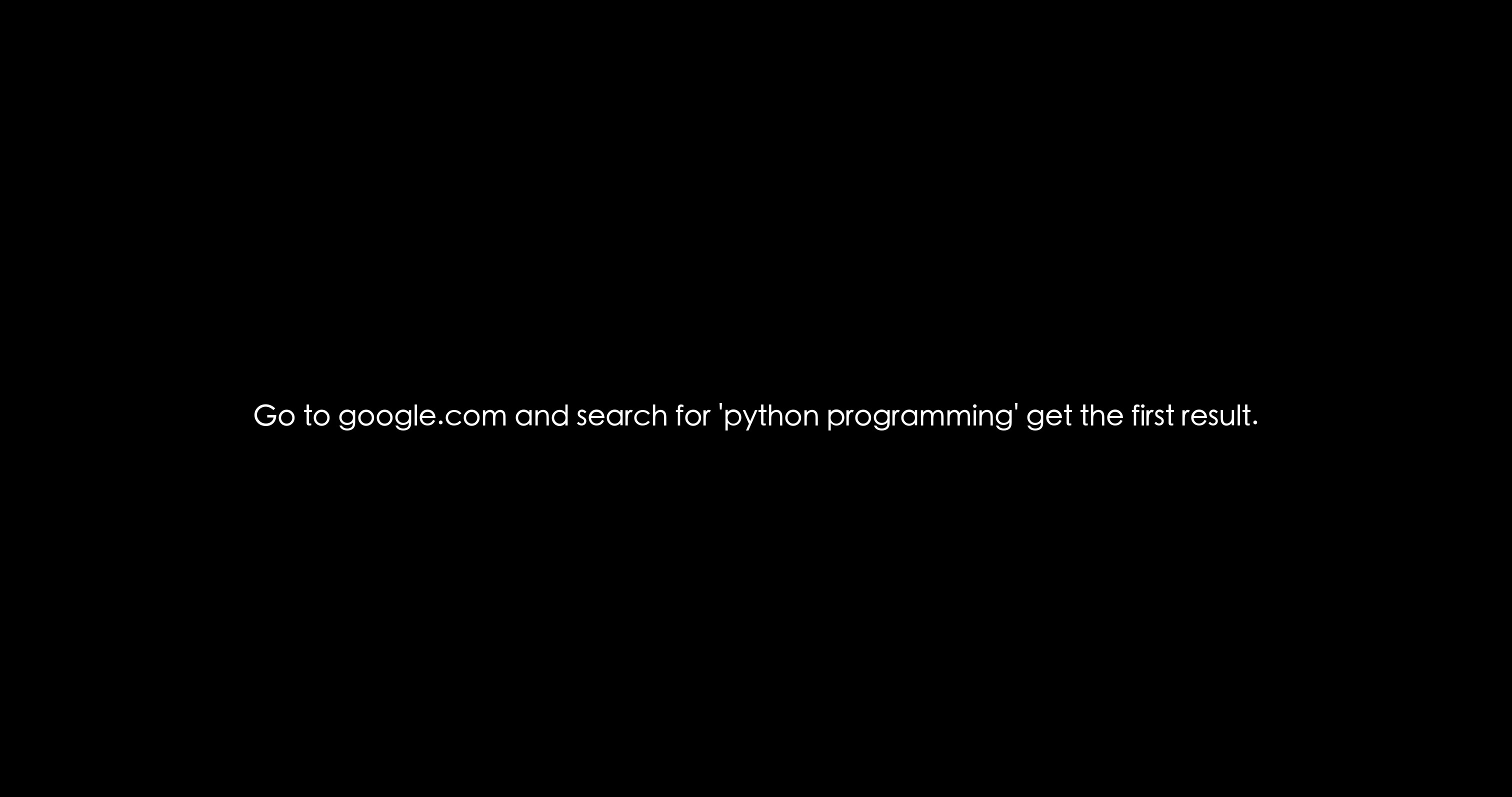}
  \end{minipage}\hfill
  \begin{minipage}[t]{0.48\textwidth}\centering
    \includegraphics[width=\linewidth]{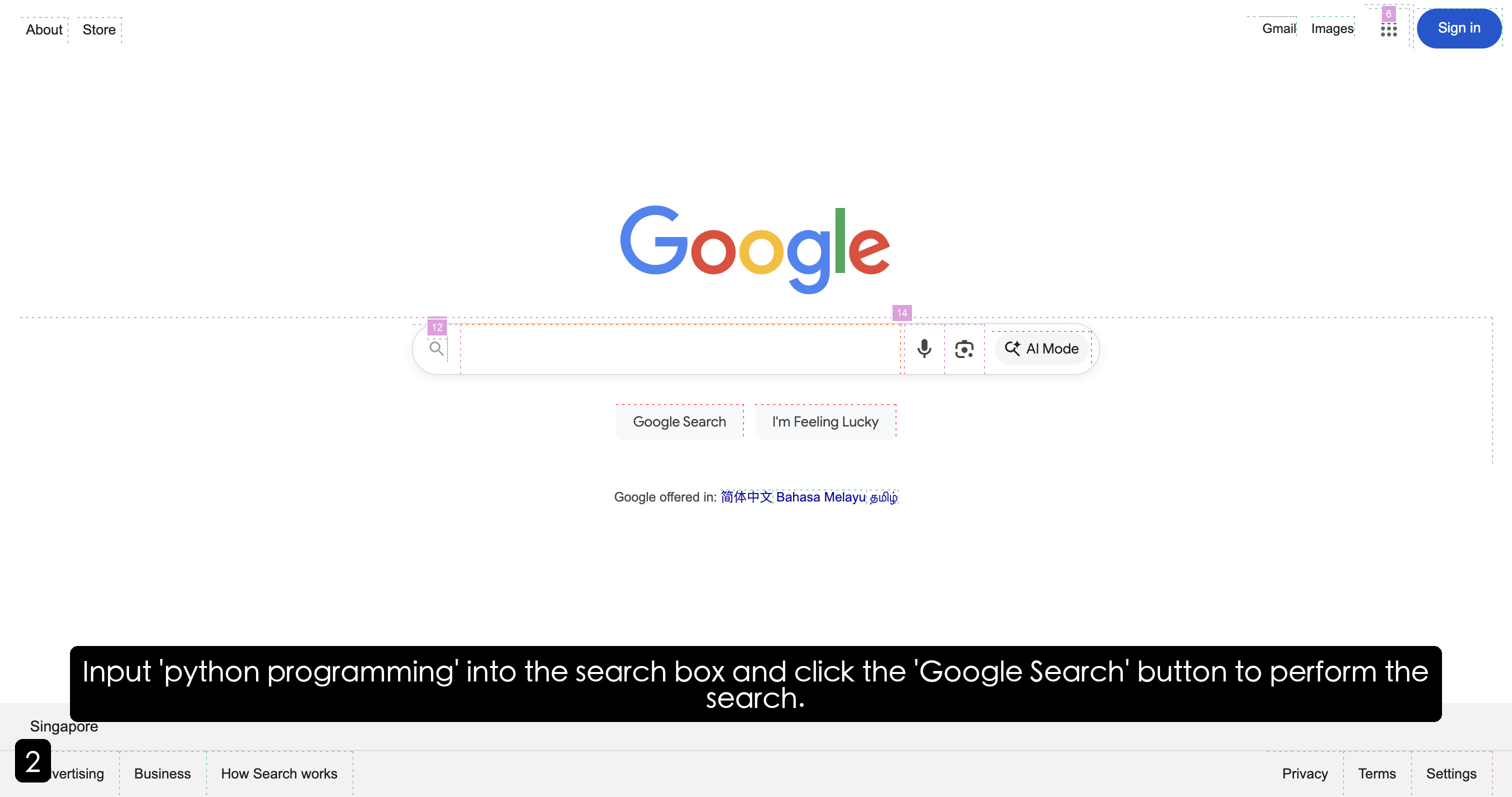}
  \end{minipage}
  
  \vspace{0.5em}
  
  \begin{minipage}[t]{0.48\textwidth}\centering
    \includegraphics[width=\linewidth]{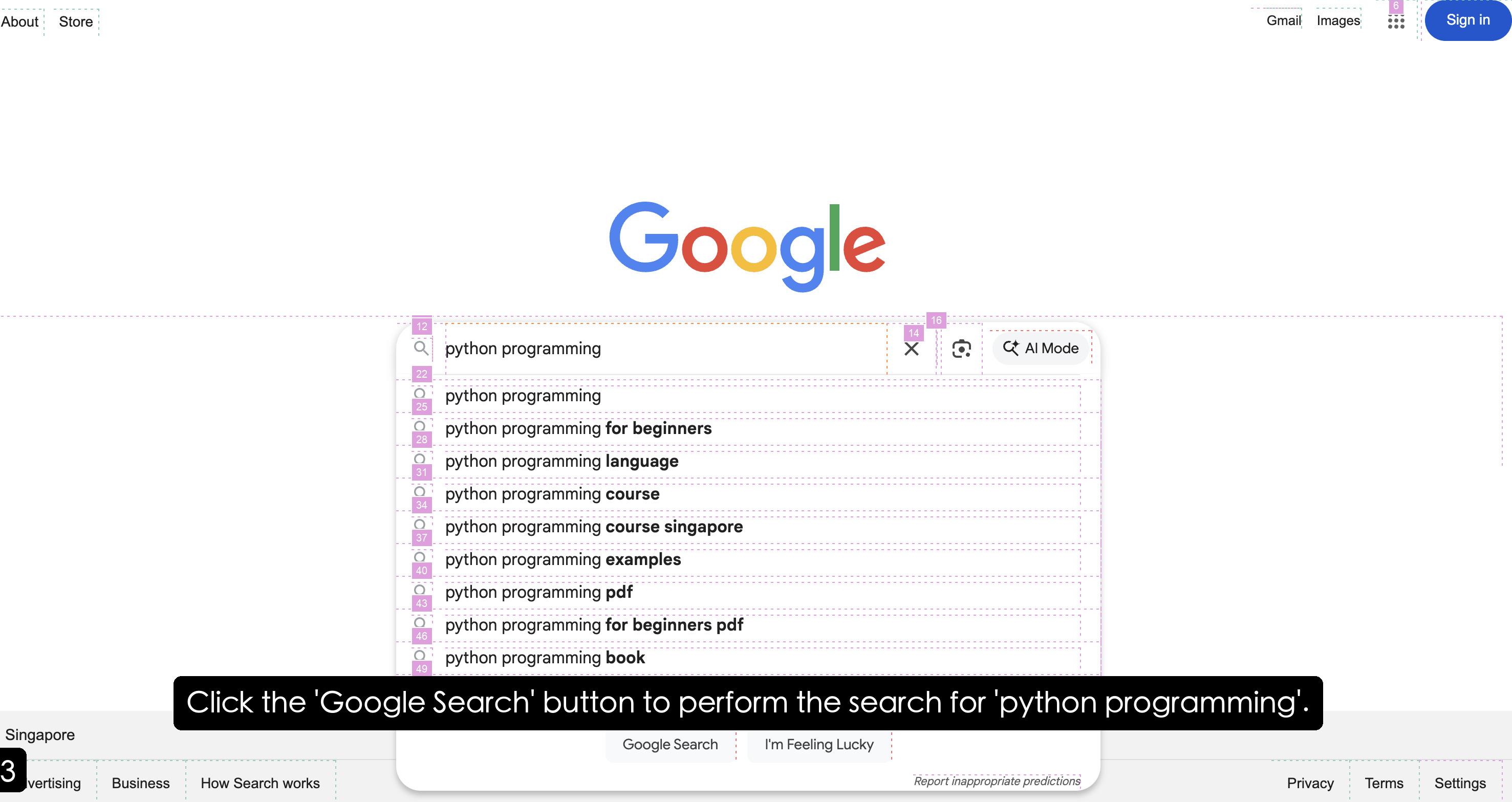}
  \end{minipage}\hfill
  \begin{minipage}[t]{0.48\textwidth}\centering
    \includegraphics[width=\linewidth]{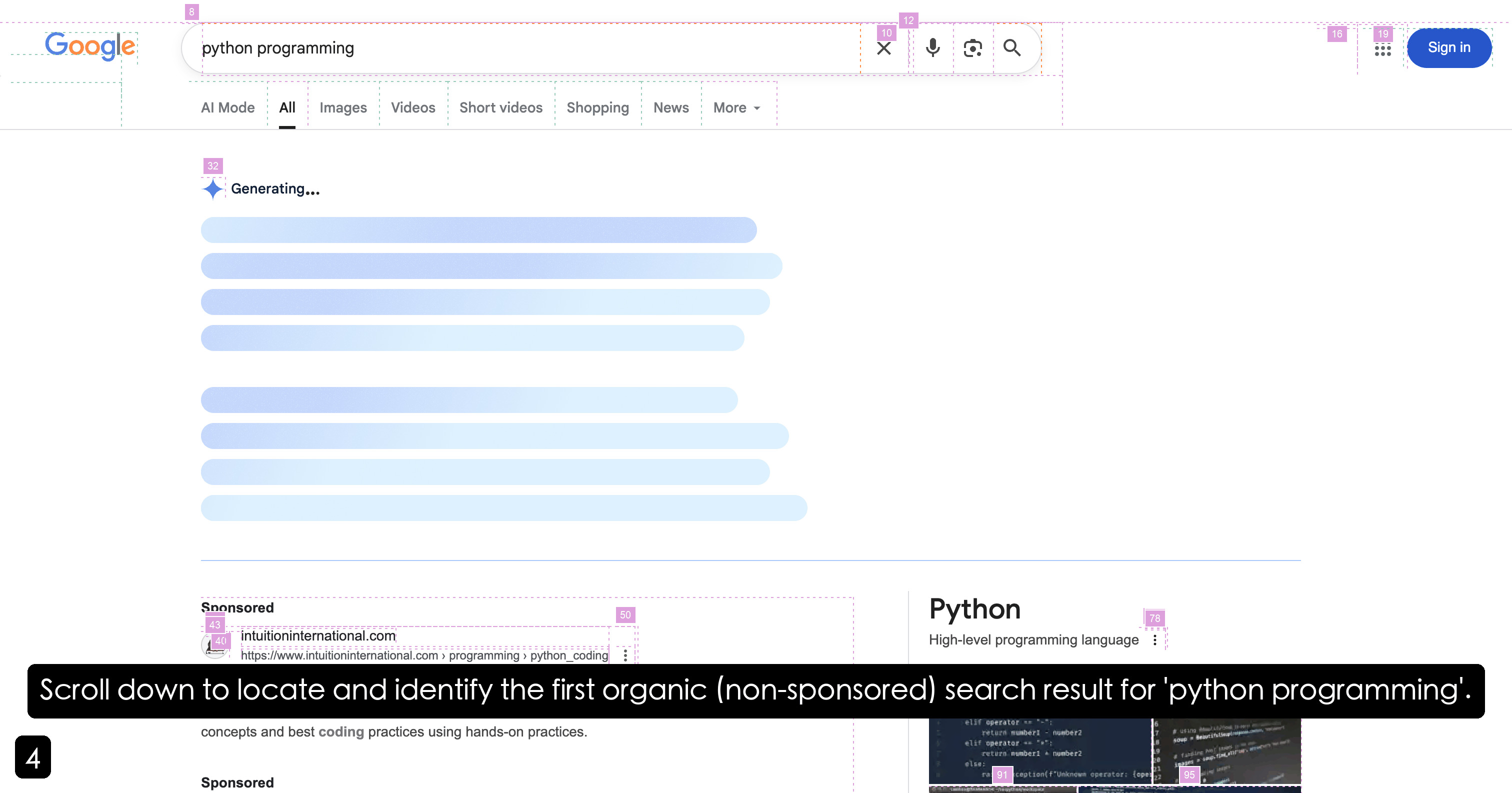}
  \end{minipage}
  
  \vspace{0.5em}
  
  \begin{minipage}[t]{0.48\textwidth}\centering
    \includegraphics[width=\linewidth]{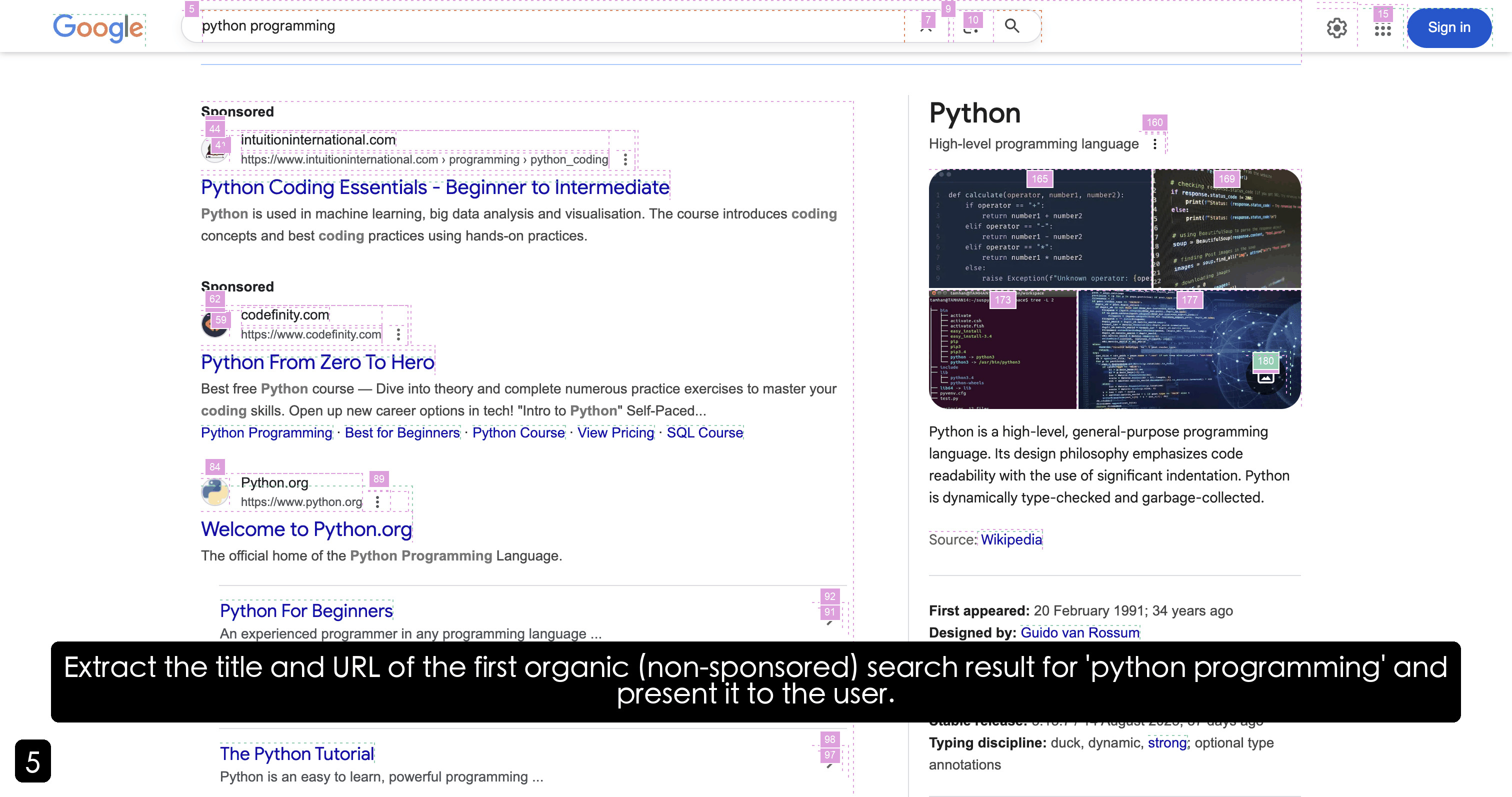}
  \end{minipage}
  \caption{Case study of TEA agent for browser operation.}
  \label{app_fig:browser_operation}
\end{figure}

This case study demonstrates the agent's sophisticated browser automation capabilities through a comprehensive web interaction scenario involving the search for "python programming" content. The agent exhibits advanced multi-modal reasoning by simultaneously processing both DOM (Document Object Model) structures and visual elements to understand webpage layout and functionality. Through systematic analysis of page elements, the agent can identify interactive components, assess their relevance to the search objective, and make informed decisions about subsequent navigation actions. The execution demonstrates the agent's capacity for autonomous web exploration, where it can parse complex webpage structures, interpret visual cues, and execute precise interactions to achieve its objectives. This capability extends beyond simple element clicking to encompass sophisticated understanding of webpage semantics and user interface patterns, with remarkable proficiency in handling dynamic content, managing asynchronous operations, and adapting to varying webpage architectures across different domains and platforms.

The browser automation framework incorporates several advanced technical components that enable robust web interaction. The agent leverages hierarchical DOM parsing algorithms to construct semantic representations of webpage structure, enabling precise element localization and interaction planning. Visual processing capabilities allow for the interpretation of complex layouts, including responsive design elements, dynamic content loading, and multi-modal interface components. The system demonstrates particular strength in handling modern web applications that rely heavily on JavaScript-driven interactions and asynchronous content loading. Furthermore, the agent exhibits sophisticated error recovery mechanisms when encountering unexpected webpage behaviors, such as dynamic content changes, popup interventions, or navigation redirects. This resilience is achieved through continuous monitoring of page state changes and adaptive strategy modification based on real-time feedback from the browser environment.

Our browser environment supports not only conventional multi-modal models combined with DOM manipulation (limited to clicking and controlling page elements without pixel-level operations), but also integrates computer-use-preview functionality that enables operator-like pixel-level precision operations, significantly expanding the scope of environmental exploration capabilities. This dual-mode architecture provides unprecedented flexibility in web automation, allowing for both high-level semantic interactions and low-level pixel-accurate operations when necessary.

\newpage
\section{Prompts}
\label{app_sec:prompts}

Our foundational agent framework is built upon a ReAct-based tool-calling agent architecture, which follows a systematic "thinking-then-action" paradigm. During execution, the agent records its decision-making process and execution trajectory, continuously summarizing experiences and extracting insights through its memory mechanism. The agent employs a \texttt{done} tool to determine task completion, ensuring reliable termination of complex workflows. Notably, the planning agent is built upon this comprehensive tool-calling foundation to coordinate multifaceted resources, while specialized agents such as the deep researcher, deep analyzer, browser operator, and tool manager utilize optimized custom workflows to achieve an optimal balance between high task completion rates and reduced resource consumption. We do not provide the detailed prompts for other specialized agents and the self-evolution module here; for further details, please refer to the source code in the supplementary materials. 

The agent's prompt structure consists of two primary components: a static \textbf{system prompt} that establishes the agent's role, capabilities, and behavioral guidelines, and a dynamic \textbf{agent message prompt} that provides the task instructions, environmental state, and execution history. These components work together to guide the agent's reasoning process and action selection. The template of the tool-calling prompt is shown as follows:

\textbf{Tool Calling Prompt Template:}
\tcbinputlisting{
    colback=gray!10,
    colframe=gray!50,
    boxrule=1pt,
    arc=3pt,
    left=5pt,
    right=5pt,
    top=5pt,
    bottom=5pt,
    breakable,
    listing file={assets/tool_calling_prompt.txt},
    listing only,
    listing options={
        basicstyle=\small\ttfamily,
        breaklines=true,
        frame=none
    }
}

The system prompt is structured to support the TEA (Tool-Environment-Agent) protocol through comprehensive context management and rule enforcement across three core components. The prompt explicitly manages \textbf{Agent Context} through role definition (\texttt{agent\_profile}), core capabilities (\texttt{agent\_introduction}), and behavioral guidelines (\texttt{language\_settings}). It further incorporates rigorous task management (\texttt{task\_rules}), working directory constraints (\texttt{workdir\_rules}), and an iterative execution history framework (\texttt{agent\_history\_rules}) coupled with memory accumulation (\texttt{memory\_rules}) to ensure continuous progress monitoring and context maintenance. \textbf{Environment Context} management is implemented through environment rules (\texttt{environment\_context\_rules}) that define interaction patterns, state transitions, and multimodal feedback mechanisms, providing structured access to environmental status and observations. \textbf{Tool Context} management is achieved through strict tool-use rules and efficiency guidelines (\texttt{tool\_use\_rules}), alongside a strategic \texttt{todo} mechanism (\texttt{todo\_rules}) for systematic planning of multi-step tasks. The entire process is underpinned by systematic reasoning rules (\texttt{reasoning\_rules}) and a rigid JSON output protocol (\texttt{output}), enabling seamless coordination between agent reasoning, environmental awareness, and tool utilization within the TEA distributed architecture.

\end{document}